\documentclass{article}

\PassOptionsToPackage{numbers}{natbib}
 \usepackage[main, final]{neurips_2026}

\usepackage[utf8]{inputenc} 
\usepackage[T1]{fontenc}    
\usepackage{hyperref}       
\usepackage{url}            
\usepackage{multirow}
\usepackage{booktabs}       
\usepackage{amsfonts}       
\usepackage{nicefrac}       
\usepackage[ruled,vlined,linesnumbered]{algorithm2e}
\usepackage{microtype}      
\usepackage[dvipsnames]{xcolor}         
\usepackage{tikz}
\usetikzlibrary{calc}
\usepackage{tikzscale}
\usepackage{graphicx}
\usepackage{xspace}
\usepackage{pifont}
\usepackage{amsmath}
\usepackage{gensymb}
\usepackage{hhline}
\usepackage{todonotes}
\usepackage{caption}
\usepackage{colortbl}
\usepackage{subcaption}
\usepackage{ctable} 
\usepackage{cleveref}
\usepackage{wrapfig}
\usepackage{float}

\newcommand{\eg}{e.g.\xspace}
\newcommand{\name}{PaGeR}
\title{Unified Panoramic Geometry Estimation via Multi-View Foundation Models}

\author{%
  Vukasin Bozic \\
  ETH Zürich \\
  \And
  Isidora Slavkovic \\
  Google \\
  \And
  Dominik Narnhofer \\
  ETH Zürich \\
  \And
  Nando Metzger \\
  ETH Zürich \\ 
  \And
  Denis Rozumny \\
  ETH Zürich\thanks{Now at Meta.} \\  
  \And
  Konrad Schindler \\
  ETH Zürich \\
  \And
  Nikolai Kalischek \\
  Google \\
}

\begin{document}

\maketitle

\vspace{-0.4cm}
\begin{center}
    \begin{tikzpicture}[
    x=1cm, y=1cm,
    inlay/.style={
        fill=black!60,         
        fill opacity=0.6,      
        text opacity=1,        
        text=white,            
        font=\sffamily\footnotesize,
        inner sep=2pt,         
        anchor=north,
        rounded corners=2pt
    }
]

    \begin{scope}[shift={(0, 0)}]
        \clip [rounded corners=0.15cm] (-2, -6.0) rectangle (10, 2);
        
        \begin{scope}[shift={(0, 0)}]
            \node[inner sep=0] at (0, 1) {
                \includegraphics[width=4cm]{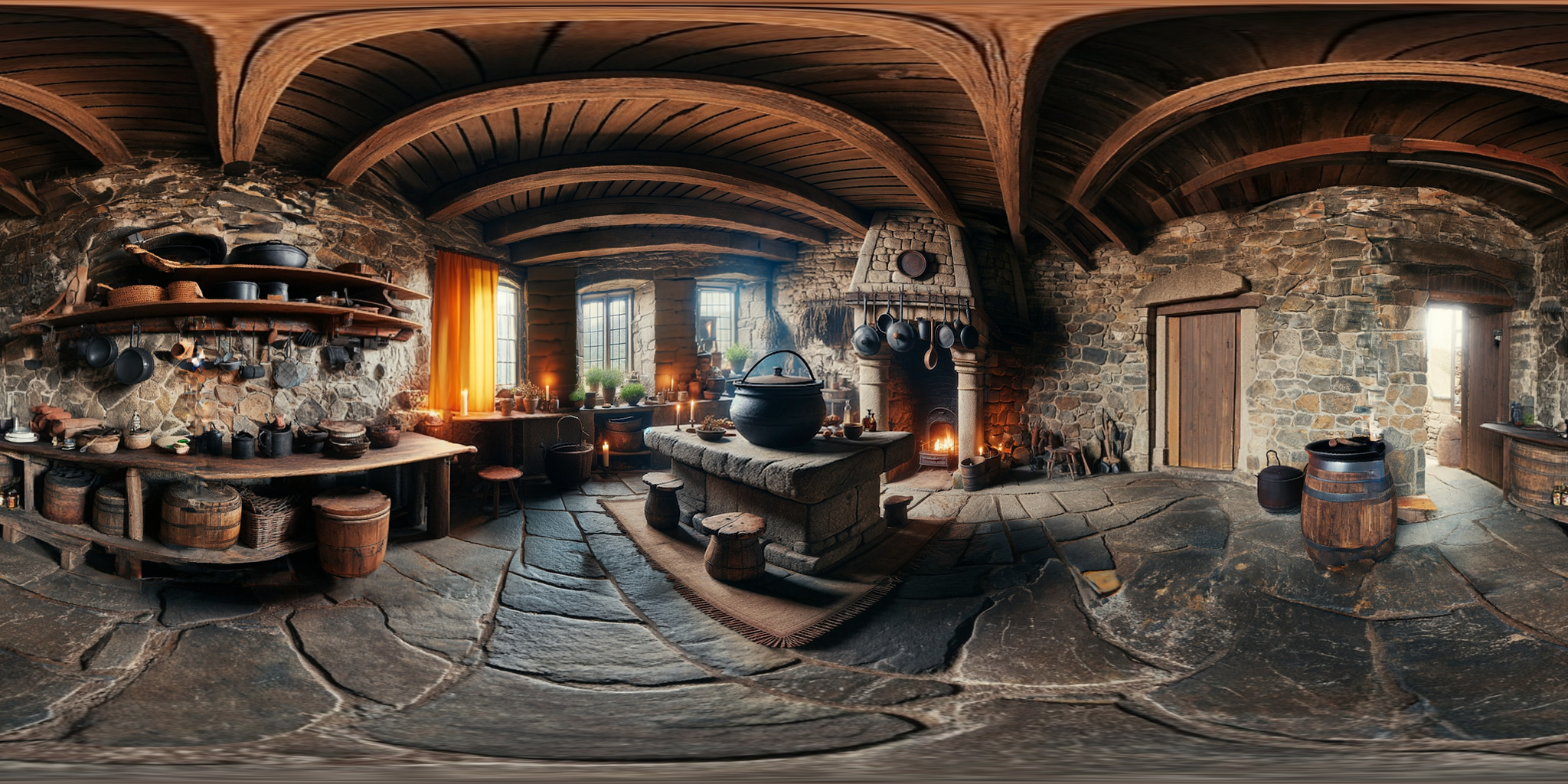}
            };
            
            \node[inner sep=0, fill=black] at (0, -1) { 
                \includegraphics[width=4cm, height=2cm]{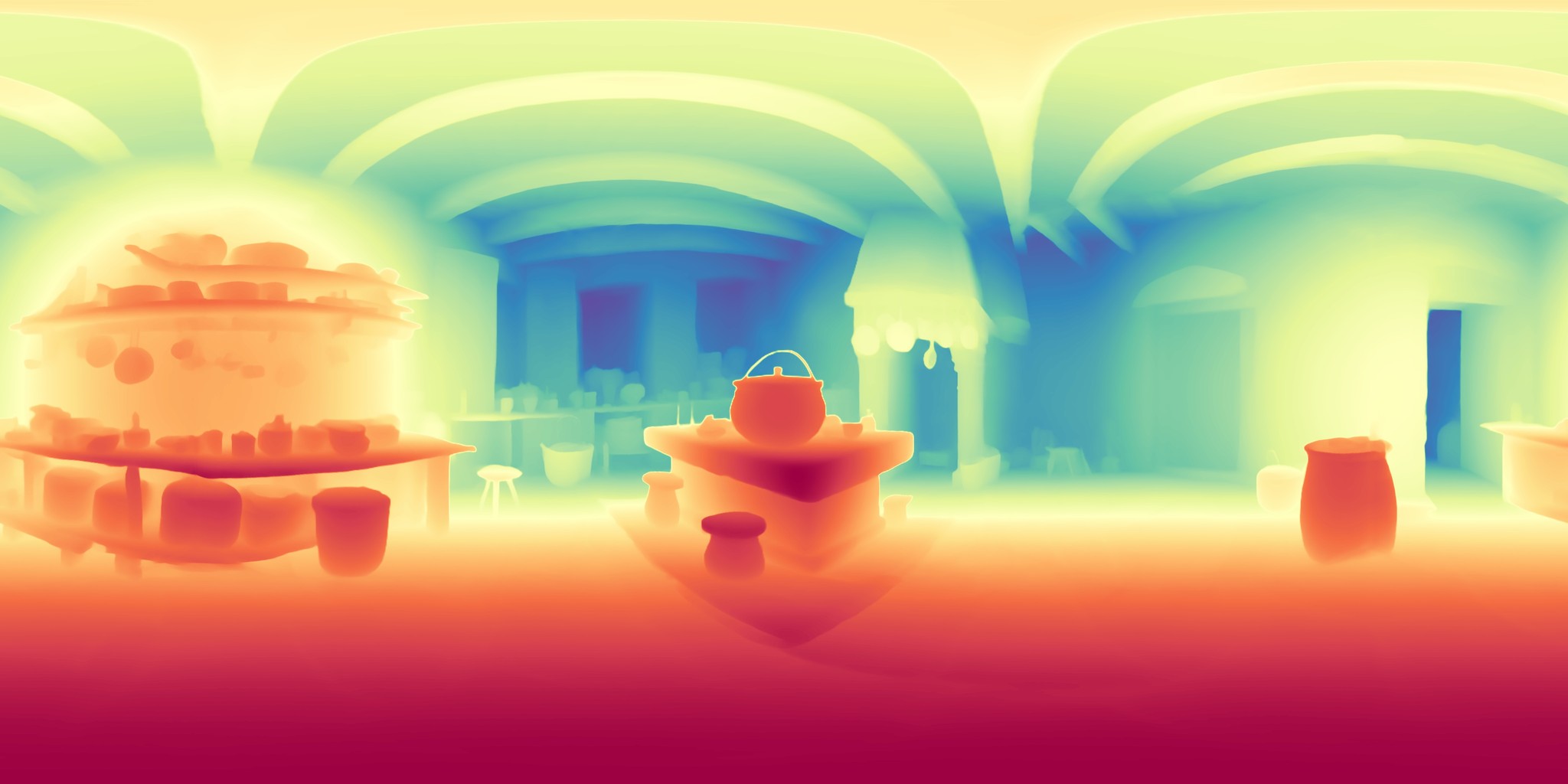}
            };

            \node[inner sep=0] at (6, 0) { 
                \includegraphics[width=8cm, height=4cm]{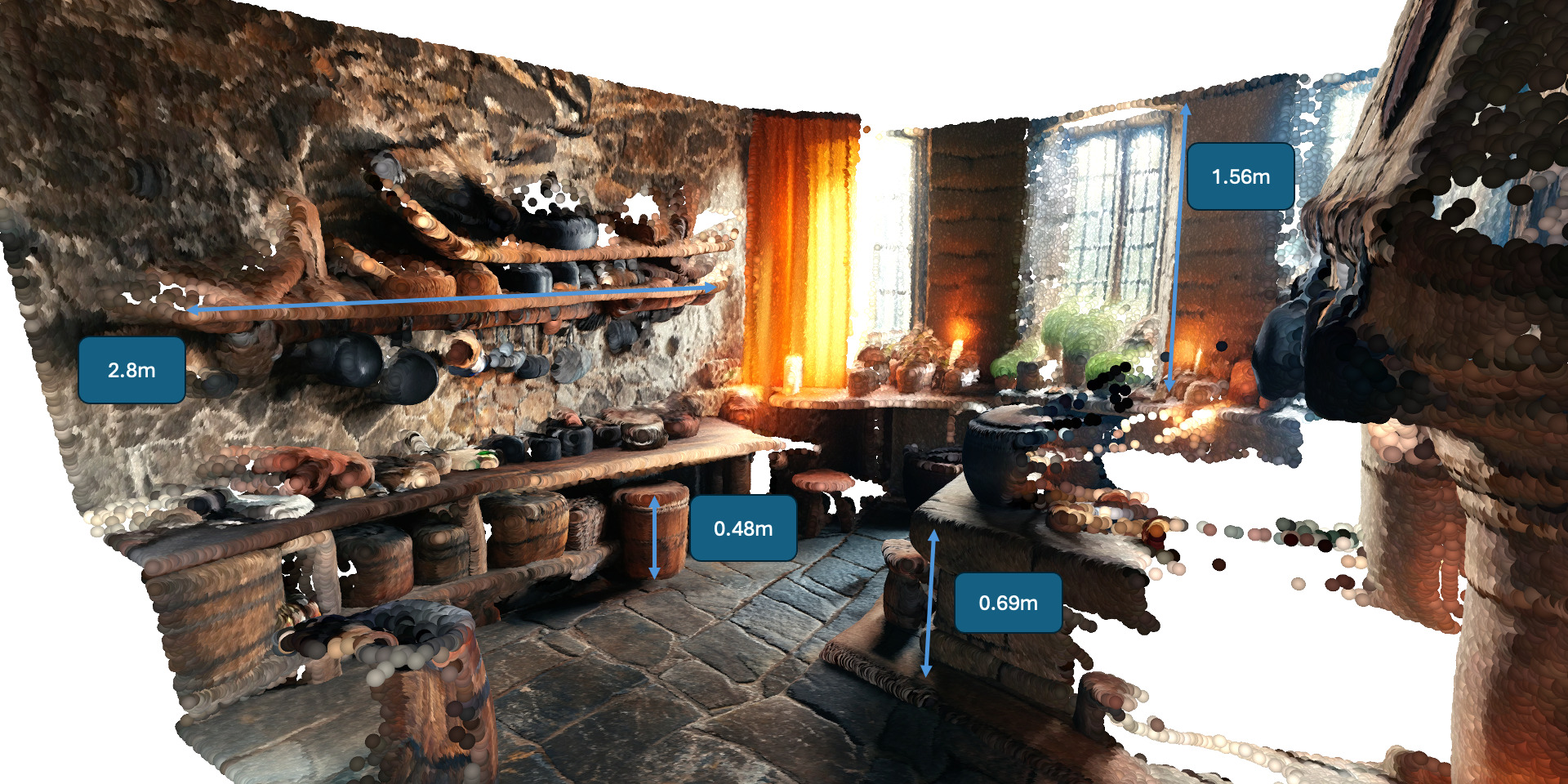}
            };
            
            \node[inlay] at (0, 1.95) {Input};
            \node[inlay] at (6, 1.95) {Point Cloud};
            \node[inlay] at (0, -0.05) {Depth};
        \end{scope}
    
        \begin{scope}[shift={(0, -3.05)}]
            \node[inner sep=0] at (0, 0) {
                \includegraphics[width=4cm]{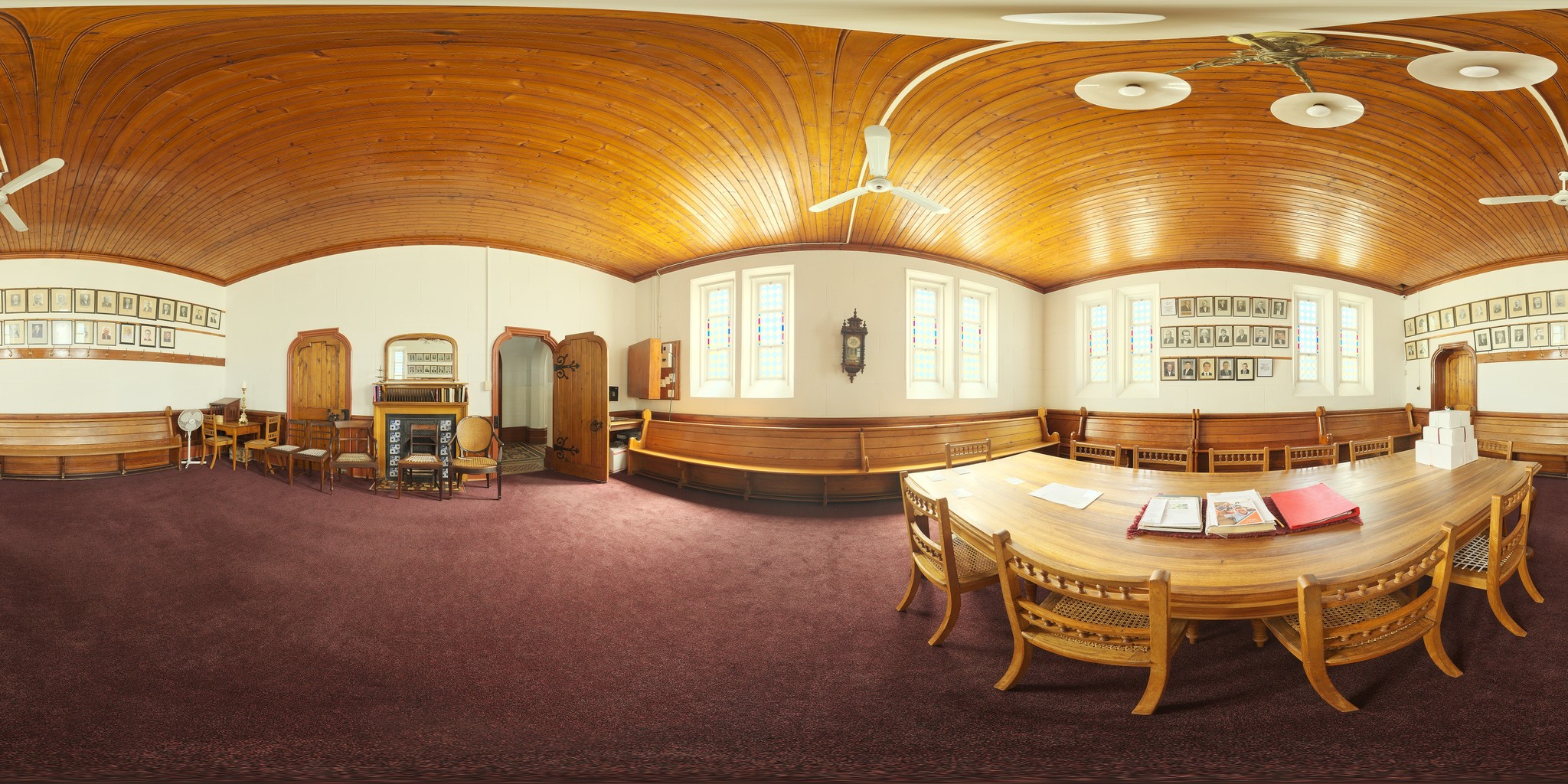}
            };
            
            \node[inner sep=0] (dp2) at (4, 0) { 
                \includegraphics[width=4cm]{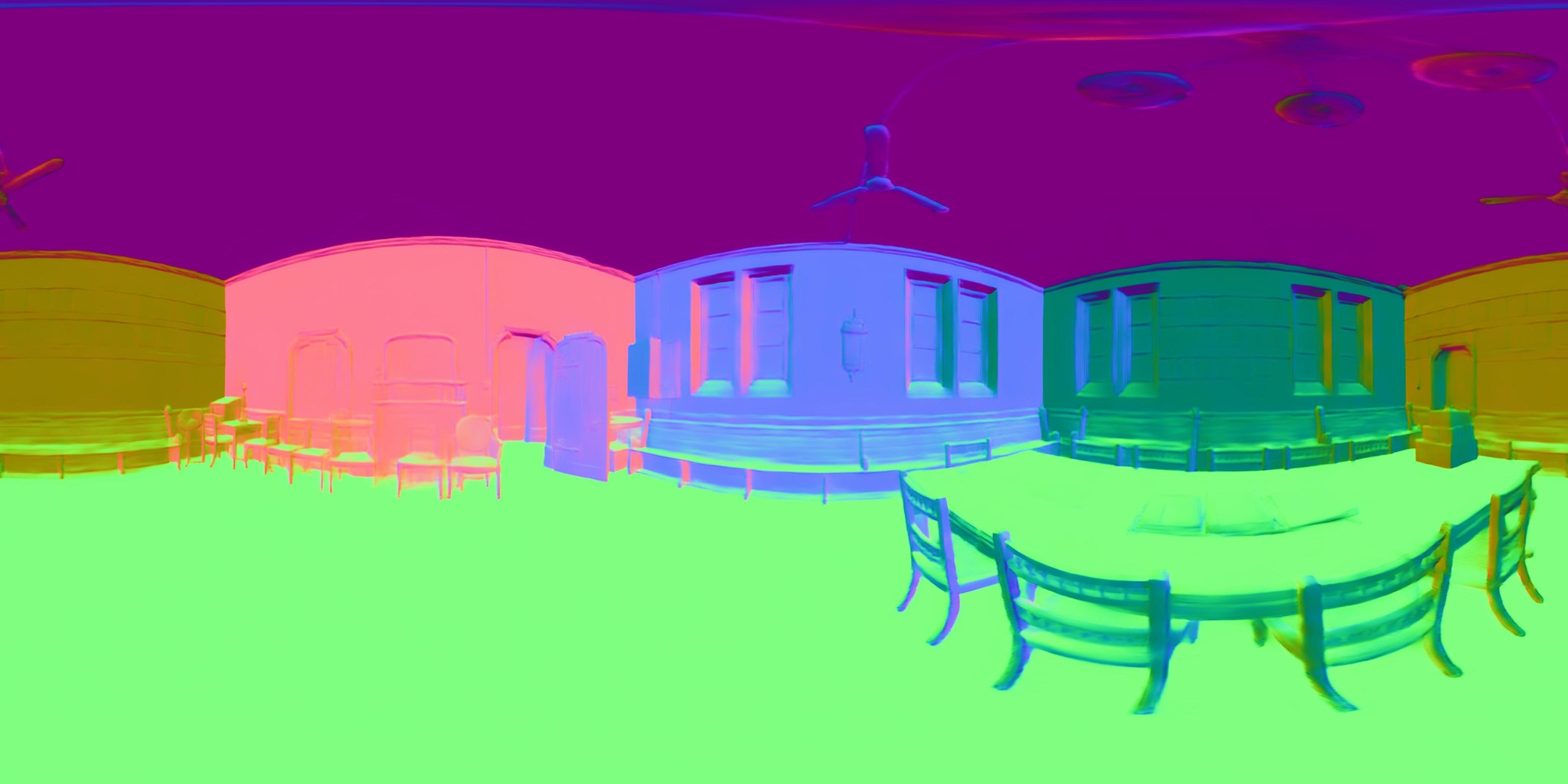}
            };
            \begin{scope}
                \clip ($(dp2.south west) + (1,0)$) -- (dp2.south west) -- (dp2.north west) -- (dp2.north east) -- ($(dp2.north east) + (1,0)$) -- cycle;
                \node[inner sep=0] at (4, 0) {
                    \includegraphics[width=4cm]{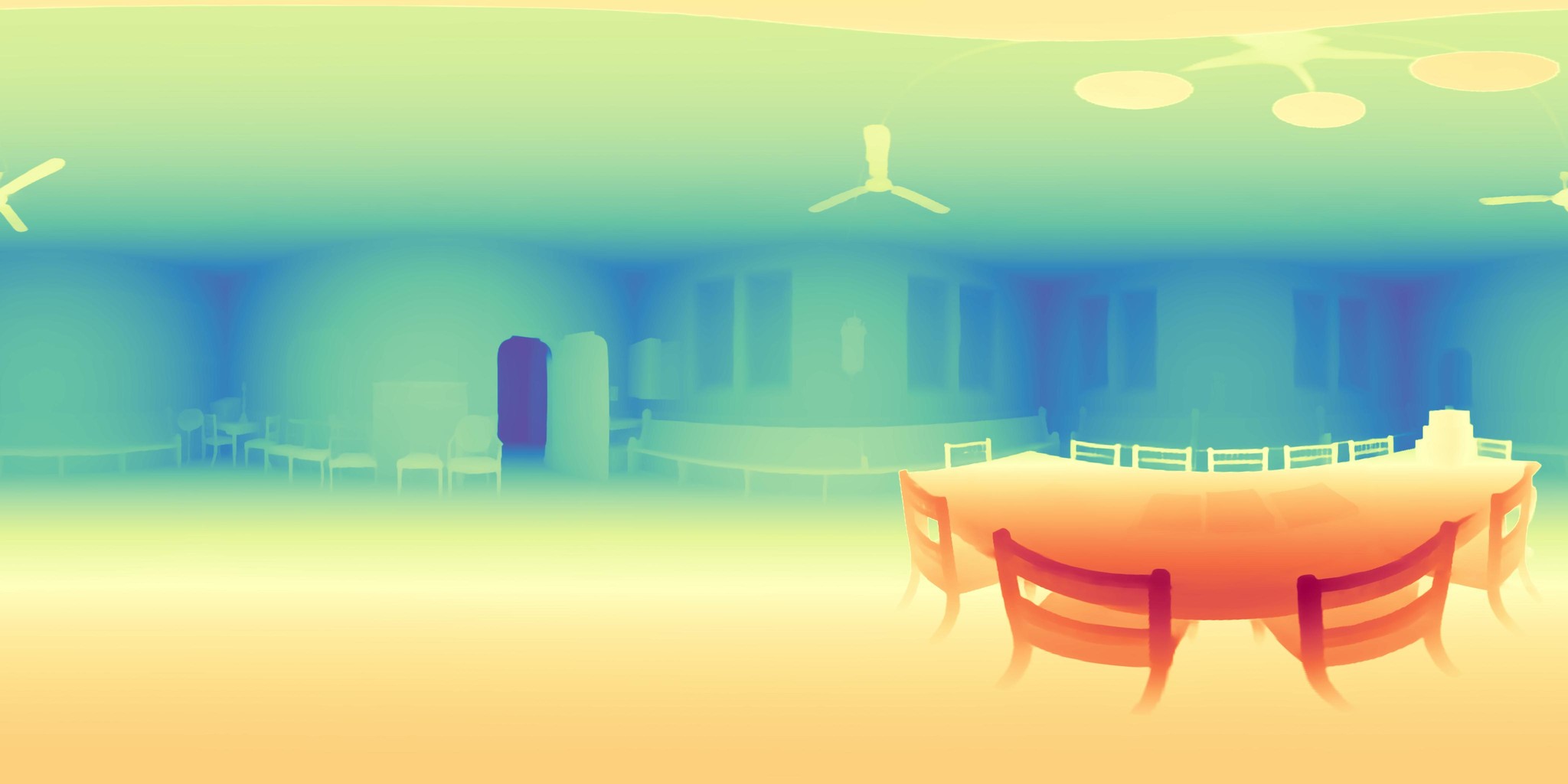}
                };
            \end{scope}
            \begin{scope}
                \clip (dp2.south west) rectangle (dp2.north east);
                \draw[white, ultra thin] ($(dp2.south west) + (1,0)$) -- ($(dp2.north east) + (1,0)$);
            \end{scope}
            
            \node[inner sep=0] at (8, 0) { 
                \includegraphics[width=4cm, height=2cm]{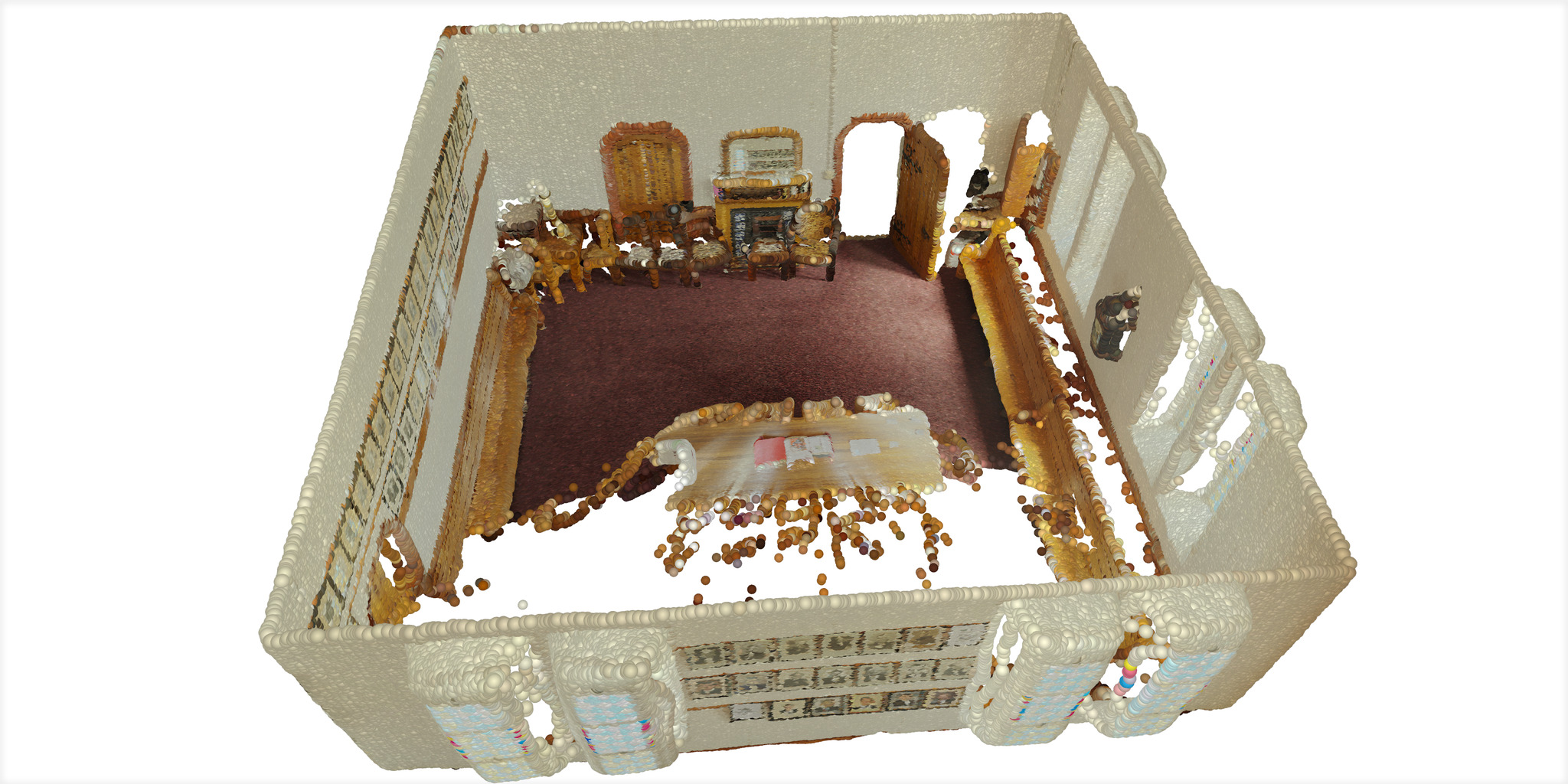}
            };
            
            \node[inlay] at (0, 0.95) {Input};
            \node[inlay] at (4, 0.95) {Depth / Normal};
        \end{scope}
    
        \begin{scope}[shift={(0, -5.10)}]
            \node[inner sep=0] at (0, 0) {
                \includegraphics[width=4cm]{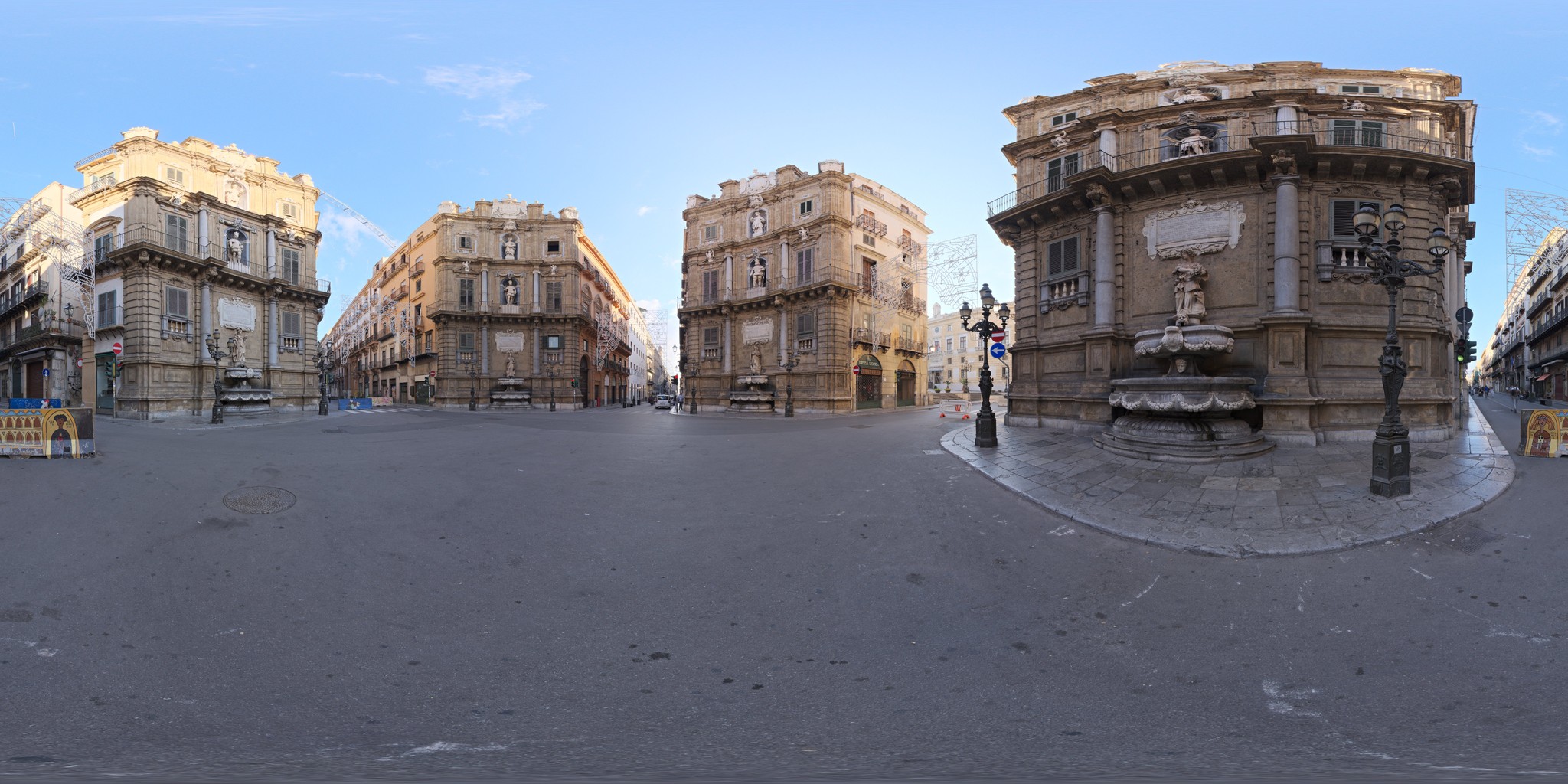}
            };
            
            \node[inner sep=0] (dp3) at (4, 0) { 
                \includegraphics[width=4cm]{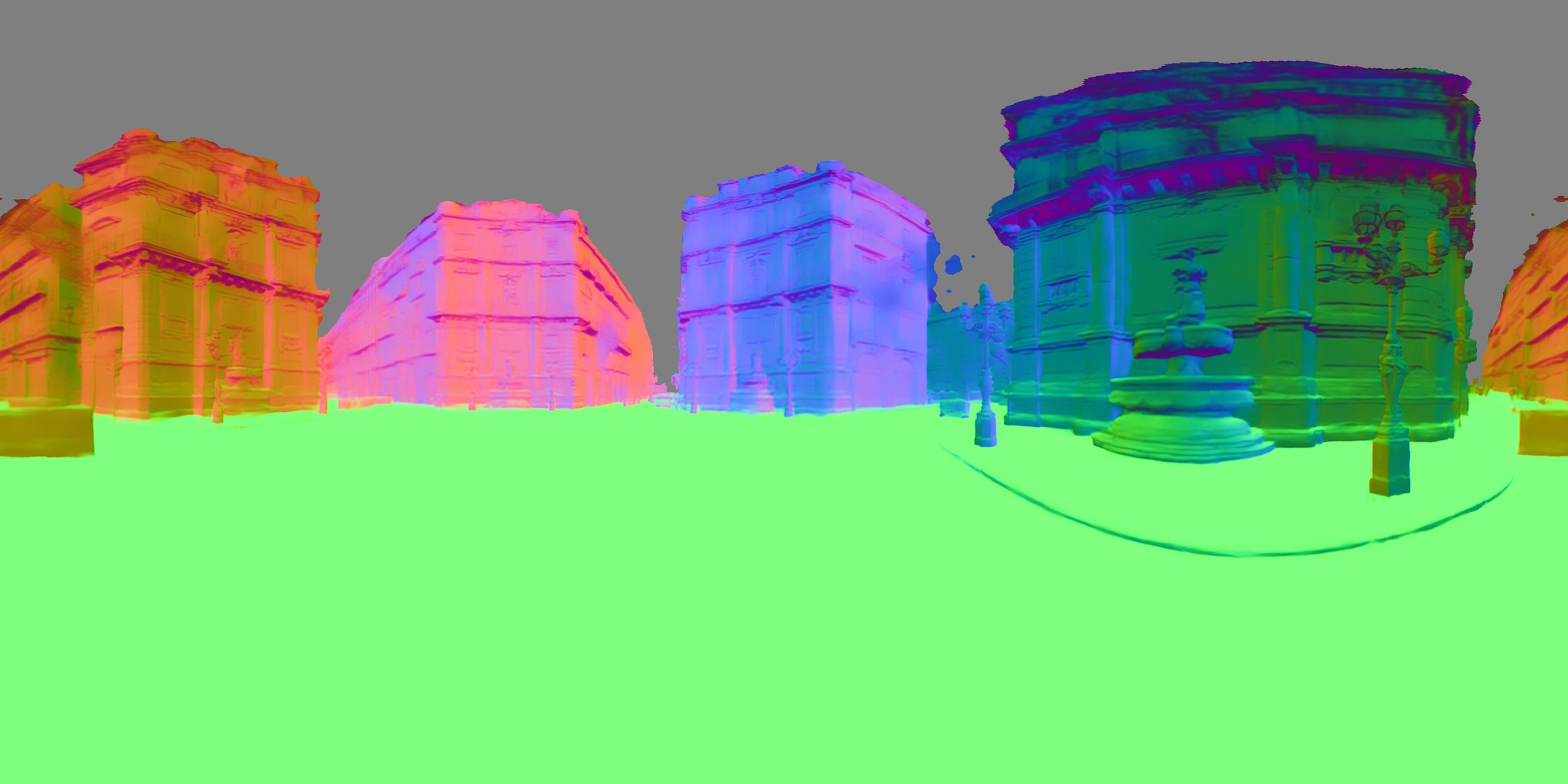}
            };
            \begin{scope}
                \clip ($(dp3.south west) + (1,0)$) -- (dp3.south west) -- (dp3.north west) -- (dp3.north east) -- ($(dp3.north east) + (1,0)$) -- cycle;
                \node[inner sep=0] at (4, 0) {
                    \includegraphics[width=4cm]{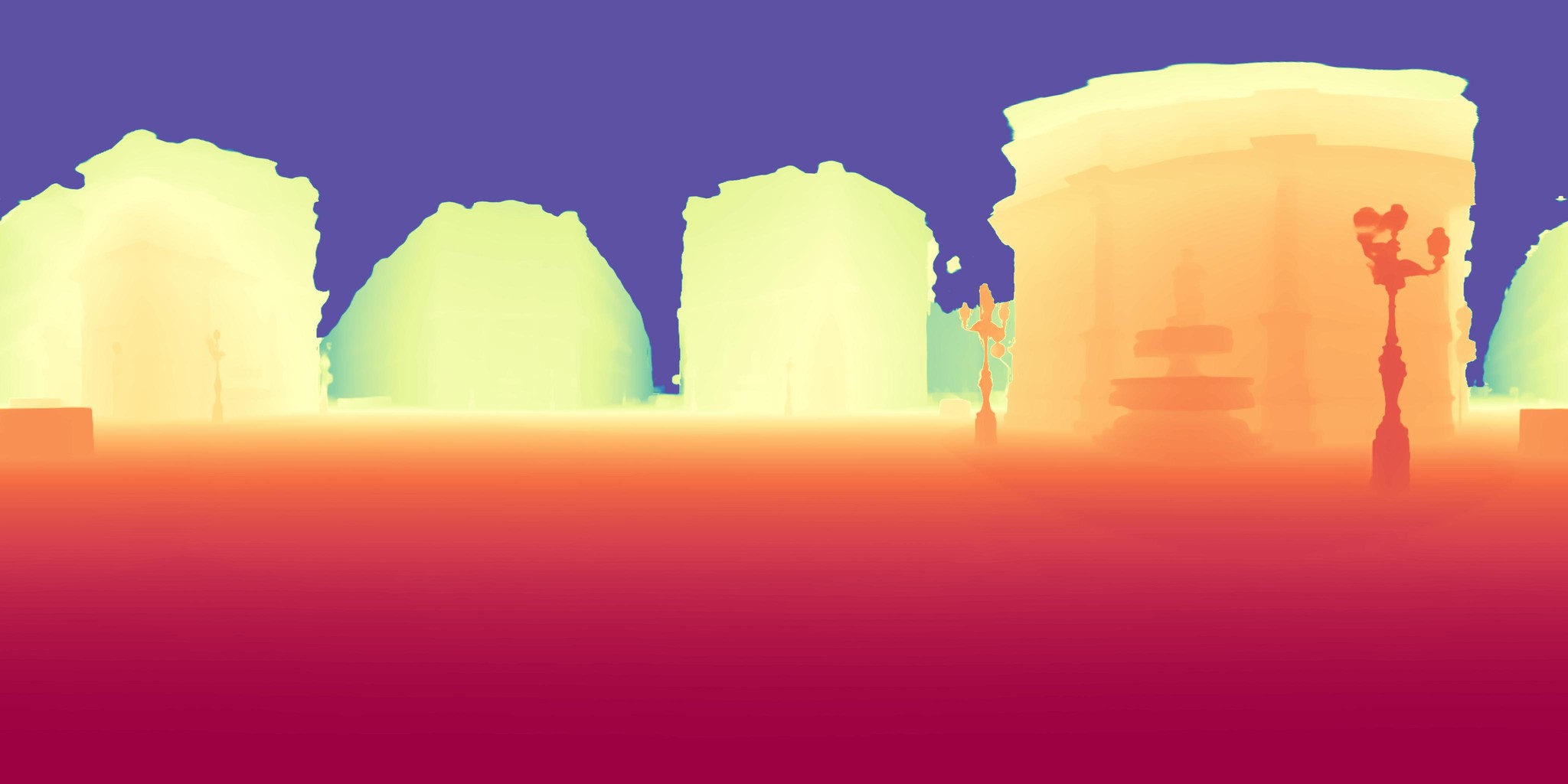}
                };
            \end{scope}
            \begin{scope}
                \clip (dp3.south west) rectangle (dp3.north east);
                \draw[white, ultra thin] ($(dp3.south west) + (1,0)$) -- ($(dp3.north east) + (1,0)$);
            \end{scope}
        
            \node[inner sep=0] at (8, 0) { 
                \includegraphics[width=4cm, height=2cm]{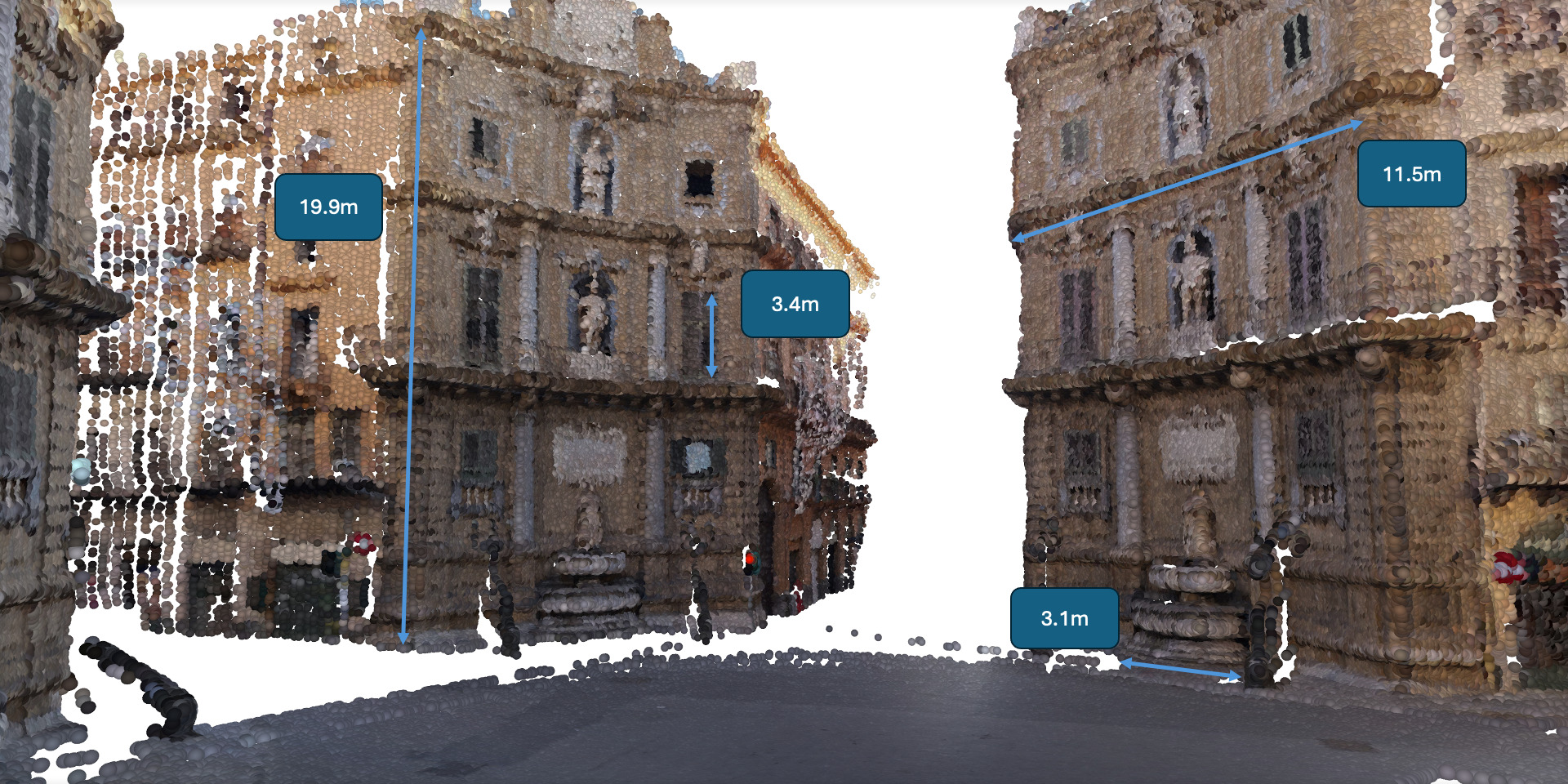}
            };
        \end{scope}
        
    \end{scope}

\end{tikzpicture}
    \captionof{figure}{\textbf{Different geometric modalities predicted by \name{}.} Given only a single monocular panoramic image, our framework simultaneously reconstructs highly detailed scale-invariant depth, absolute metric depth, surface normals, and sky segmentation masks across both indoor and outdoor environments. Spatial measurements are presented in meters.}
    \label{fig:teaser}
\end{center}

\begin{abstract}
Geometry estimation from perspective images has greatly advanced, maturing to the point where off-the-shelf foundation models are able to reconstruct 3D scene structure not only from multi-view imagery, but even from a single view.
A natural extension is 3D reconstruction from panoramas, with the exciting prospect of recovering a full $360^\circ$ scene from a single panoramic image. 
In this work, we introduce \name{} (\underline{Pa}noramic \underline{Ge}ometry \underline{R}econstruction), a framework to lift powerful 3D foundation models designed for perspective imagery to the panorama domain. Our strategy is to start from a pre-trained transformer for 3D reconstruction and turn it into a unified high-performance model that predicts scale-invariant depth, metric depth, surface normals, and sky masks from both perspective and omnidirectional images, in a single forward pass. By keeping architectural changes to a minimum and mixing perspective and panoramic images during training, \name{} retains the rich 3D prior of the underlying foundation model while learning to also estimate geometrically consistent $360^\circ$ scenes from single panoramas. We extensively test our method in both indoor and outdoor environments and find that it delivers state-of-the-art performance and excellent zero-shot performance across a wide range of scenes.
\end{abstract}

\section{Introduction}
\label{intro}

Sensing and understanding the 3D structure of the surrounding world is important in many applications, ranging from virtual and augmented reality to autonomous driving and robotics.
Scene depth and surface normals are two central geometric properties in that context: together, they describe the position and the local surface orientation at any point of the scene, providing a complete representation that supports graphics tasks like rendering and relighting as well as high-level perception tasks like spatial reasoning and path planning.

A particularly attractive, but also heavily ill-posed variant is to recover depth or surface normals from a single RGB image~\cite{hu2024metric3d, ke2024repurposing}, obviating the need for multi-view capture and camera pose estimation. Early attempts relied on limited datasets and convolutional backbones~\cite{eigen2014depthmappredictionsingle, monodepth17}, but large-scale data collection~\cite{ranftl2020towards} and advances in neural architectures, most notably vision transformers and denoising diffusion models~\cite{yang2024depth,ke2024repurposing}, have greatly advanced monocular geometry estimation. Most recently, this trend has converged with learning-based multi-view reconstruction, leading to foundational feed-forward models~\cite{wang2025vggtvisualgeometrygrounded, lin2025depth3recoveringvisual} capable of zero-shot, dense 3D reconstruction. 
From their massive training datasets, captured under diverse imaging conditions, these models acquire not only an understanding of multi-view geometry but also an elaborate prior of the world's 3D surface structure, which supports detailed, dense depth estimation from single views.

Yet, these models are designed for perspective images such that every image covers only a limited field of view, and many viewpoints must be aggregated and fused to build up spatial context and perceive the complete environment.
Panoramic images, by construction, provide a full 360° view around the camera location, offering rich global context for holistic 3D understanding. However, high-quality panoramic datasets with metrically accurate depth and surface normals, needed to train panoramic reconstruction models, are laborious to collect and remain scarce. As a result, existing models tend to overfit to comparatively small datasets and struggle to generalize to unseen scenes. 
Another limitation is that existing models commonly represent panoramic images in equirectangular projection, which introduces serious geometric distortions. On the one hand, this means an extremely uneven sampling of the ray space (and, after unwarping, of the 3D environment). On the other hand, and perhaps more importantly, it means that one cannot easily employ transfer learning from models trained with perspective images.

We take a different route and explore cubemaps as our panorama representation. Rather than designing custom architectures applicable specifically to panoramas, we use this parametrization to repurpose state-of-the-art perspective foundation models for the panoramic domain.
The cubemap representation has been used to adapt pre-trained, diffusion-based image generators to 360° imagery \cite{kalischek2025cubediff}. For our purposes, we prefer to build on top of deterministic, feed-forward foundation models. Besides avoiding the computational overhead of diffusion and the practical limitations of operating in a compressed latent space, models like DA3~\cite{lin2025depth3recoveringvisual} are already designed for (perspective) multi-view input. They offer a natural synergy with the cubemap format, as their geometric prior includes the integration of multiple viewing directions and is fundamentally stronger than previous, purely monocular schemes.
To properly anchor the spatial context, we explicitly condition the architecture on camera parameters and introduce targeted modifications of the decoder to ensure distortion-free and seamless reconstruction across face boundaries.
Furthermore, we propose to use a mixed training regime with both synthetic panoramas and real perspective imagery. This strategy allows the network to adapt to the 360° setting while remaining firmly grounded in real-world image statistics, thus preventing overfitting to peculiarities of synthetic data and preserving the prior of the pre-trained foundation model.

Taken together, we introduce \textit{\name{}}, a unified geometry estimation framework for panoramic (and perspective) images. Its latent representation, inherited from the geometry foundation model, is holistic and allows for simultaneous decoding of multiple scene properties. We exploit this property and equip the backbone with multiple, coupled task heads: Scale-Invariant (SI) depth, metric scale estimation, surface normals, and sky segmentation. That design allows the network to jointly reason about related geometric properties and to efficiently extract a comprehensive 3D representation in a single forward pass (see Fig.~\ref{fig:teaser}). By reparameterizing panoramic geometry as a structured multi-view problem, we achieve high-resolution, metrically accurate predictions that set a new state of the art for several benchmarks.
Furthermore, to address the lack of benchmark data for rigorous evaluation in long-range, outdoor scenarios, we curate and introduce ZüriPano, the first real-world outdoor panoramic dataset, acquired through high-accuracy and high-resolution LiDAR scans in an urban environment. In summary, our contributions are:

\begin{itemize}
\item \textbf{A novel strategy to adapt geometry foundation models} to panorama geometry. Our scheme is built around the cubemap representation consisting of six perspective images, and combines it with a hybrid training strategy to seamlessly transfer 3D scene priors to the 360° panorama setting, sidestepping degradations caused by equirectangular distortion.
\item \textbf{\name{}, a unified panoramic geometry estimation model}, featuring a shared transformer backbone (adopted from DA3) and specialized task heads to enable holistic reconstruction in a single forward pass.
\item \textbf{Zero-shot generalization} to unseen indoor and outdoor scenes, outperforming methods limited to a specific domain; and the new \textbf{ZüriPano} benchmark for zero-shot evaluation in urban setting.

\end{itemize}
\section{Method}
\label{sec:method}

This section outlines the geometric preliminaries of panoramic representations and our perspective backbone architecture. We then introduce the panoramic adaptation layers and the hybrid training strategy designed to bridge the perspective and spherical domains. Finally, we detail the unified multi-task architecture and formalize the loss objectives for each geometric modality.

\subsection{Preliminaries}

\noindent\textbf{Panoramic Image Representations.}
Panoramas capture a holistic $360^\circ\times180^\circ$ environment on the unit sphere $\mathbb{S}^2$. The standard Equirectangular Projection (ERP) maps spherical coordinates, i.e., longitude $\theta \in [-\pi, \pi]$ and latitude $\phi \in [-\pi/2, \pi/2]$, to a 2D planar grid $(u,v) \in [0,1]^2$ via:
\begin{equation}
    u = \frac{\theta}{2\pi} + 0.5, \quad v = \frac{\phi}{\pi} + 0.5
    \label{eq:erp}
\end{equation}
While structurally simple, ERP introduces severe nonlinear distortions. The horizontal sampling density scales by $\sec(\phi)$, causing extreme stretching near the poles ($\phi \to \pm \pi/2$). This domain shift degrades the efficacy of translation-invariant architectures optimized for perspective imagery.

To mitigate polar distortions, the cubemap projection maps $\mathbb{S}^2$ onto the six faces of a circumscribed unit cube $\mathcal{C}$. Each cube face constitutes a standard $90^\circ$ FoV perspective image. A 3D ray $\mathbf{p} = (x,y,z) \in \mathbb{S}^2$ is mapped to local face coordinates via gnomonic projection (e.g., $u_c = x/z, v_c = y/z$ for the front face where $z=1$). This piecewise perspective formulation offers uniform sampling and directly aligns with the inductive priors of models trained on perspective data. However, partitioning the continuous sphere introduces geometric and photometric discontinuities at face boundaries, requiring custom adaptations of the architecture to maintain global consistency.

\noindent\textbf{Geometry Transformer Backbone.}
Our framework is compatible with any multi-view transformer architecture. We instantiate our model using Depth Anything 3 (DA3)~\cite{lin2025depth3recoveringvisual}, which couples a vision transformer encoder~\cite{oquab2024dinov2learningrobustvisual} with a dense prediction transformer decoder~\cite{ranftl2021visiontransformersdenseprediction}. Given a set of $S$ perspective views $\mathcal{I} = \{I_i\}_{i=1}^S$, the encoder tokenizes the inputs and routes them through interleaved intra-image and cross-image attention layers. This global attention mechanism can optionally be conditioned on explicit camera parameters, namely intrinsic matrices $\mathbf{K}_i \in \mathbb{R}^{3 \times 3}$ and extrinsic poses $\mathbf{E}_i \in SE(3)$, to guide spatial cross-view reasoning. The encoder yields hierarchical feature maps $\mathcal{F} = \{F^{(\ell)}\}$ across transformer layers $\ell$, which the decoder progressively upsamples and fuses to output dense spatial predictions.

\subsection{Panoramic Adaptation and Joint Training}

To adapt the multi-view architecture for holistic $360^\circ$ estimation, we format the panoramic input as a six-face cubemap and supply fixed camera matrices $\mathbf{K}$ alongside axis-aligned extrinsics $\mathbf{E}_i$, $i=1,\dots,6$. While these geometric parameters explicitly define the spatial configuration, naively assembling independent face predictions into an equirectangular projection yields pronounced discontinuities at the boundaries. Furthermore, training exclusively on synthetic panoramas can cause the model to quickly diverge from its pre-training weights. We resolve these challenges through structural adaptations that favor global feature extraction and local decoding, complemented by a regularized joint training regime.

\noindent\textbf{Implicit Encoder Synchronization.} We fine-tune the ViT encoder on panoramic data without any structural modifications. Guided by the fixed camera tokens, face positional embeddings, and cross-view attention layers, the network naturally learns to route context and synchronize features across adjacent cubemap faces. The fine-tuning allows the backbone to adapt to the spherical topology while preserving the rich perspective priors learned during pre-training.

\noindent\textbf{Spherically Aware Decoder Padding.} Although global synchronization occurs in the encoder, local boundary artifacts can still emerge during dense upsampling in the decoder. To ensure continuous spherical sampling, we integrate cross-face valid padding into all convolutional and interpolation operations within the decoder architecture~\cite{huang2025dreamcube}. Instead of standard zero padding, this layer dynamically extracts features from geometrically adjacent cubemap faces, enforcing seamless geometric and photometric transitions across all boundaries.

\noindent\textbf{Mixed Panoramic / Perspective Co-Training.} To preserve the rich priors inherited from the perspective backbone and mitigate the sim-to-real domain gap, we employ a training strategy that alternates between two data streams. For panoramic batches, the network processes the full six-face configuration ($S=6$) with active cross-face padding. For perspective batches, we isolate a single real-world image ($S=1$), warp it to a $90^\circ$ field of view to match the imaging geometry of cubemap faces, and assign it the extrinsics of a random equatorial face. Cross-face padding is dynamically disabled for these perspective samples, with the layer reverting to standard zero padding. The dual-stream training protects the model from catastrophic forgetting while teaching it to handle continuous spherical observations.
\begin{figure}[!t]
    \centering
    \includegraphics[width=0.95\linewidth]{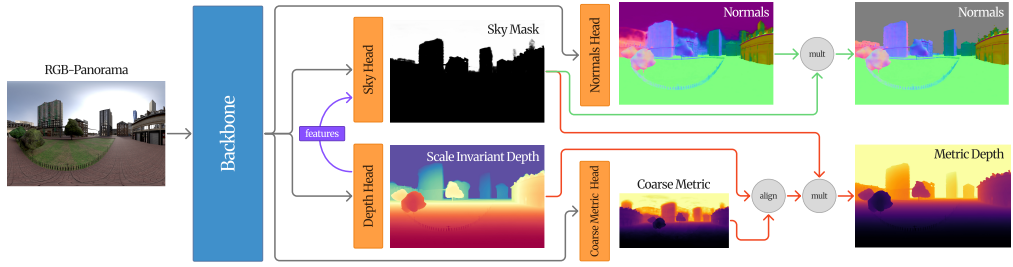}
    \caption{\textbf{\name{} Architecture.} An input RGB panorama in cubemap format is processed by a shared geometry transformer backbone to predict a sky mask, scale-invariant (SI) depth, surface normals, and coarse metric depth. In the \textcolor{BrickRed}{metric branch}, the final absolute depth is obtained by aligning the SI depth with coarse metric predictions and masking the sky. In the \textcolor{Green}{normal branch}, predicted orientations are masked by the sky segmentation to produce the final surface normal map.}
    \label{fig:method}
    \vspace{-1.4em}
\end{figure}%

\subsection{Multi-Task Geometric Decoding}

Existing geometric foundation models are typically confined to a single modality, such as scale-invariant depth estimation. For a comprehensive 3D understanding of $360^\circ$ environments, we add multi-task decoding to the unified backbone. Specialized prediction heads simultaneously decode depth, surface orientation, and sky masks in a single forward pass (see Fig.~\ref{fig:method}), always operating on the planar cubemap faces to benefit from the underlying perspective prior. 

\noindent\textbf{Scale-Invariant Depth.}
The model is supervised with the local, orthogonal per-face log-planar depth $z^* = \log(Z^*_{\text{planar}})$ to compress metric variance and avoid optimization bias from distant background objects. We remove the final exponential activation of the decoder to work directly in its native log space. The head outputs both the predicted scale-invariant log depth $\hat{z}_{\mathrm{SI}}$ and an aleatoric confidence map $c_p$. To isolate relative shape from metric size, we dynamically compute an optimal log-space shift $\beta^* = \arg\min_{\beta} \| (\hat{z}_{\mathrm{SI}} + \beta) - z^* \|_2^2$ and optimize the aligned predictions $\hat{z}_{\text{aligned}} = \hat{z}_{\mathrm{SI}} + \beta^*$. We supervise the scale-invariant depth branch using a composite loss function $\mathcal{L}_{\text{depth}}$ that balances per-pixel precision, local smoothness, and surface alignment:
\begin{equation}
\begin{aligned}
    \mathcal{L}_{\text{depth}} &= \lambda_{L_1} \mathcal{L}_1 + \lambda_{\text{grad}} \mathcal{L}_{\text{grad}} + \lambda_{\text{norm}} \mathcal{L}_{\text{norm}}
\end{aligned}
\label{eq:depth_losses}
\end{equation}
where $\mathcal{L}_1 = \frac{1}{N} \sum_{p} \left( c_p \left| \hat{z}_{\text{aligned}, p} - z^{*}_p \right| - \lambda_c \log c_p \right)$, 
$\mathcal{L}_{\text{grad}} = \frac{1}{N} \sum_{p} \sum_{i \in \{x, y\}} \left| \partial_i \hat{z}_{\text{aligned}, p} - \partial_i z^{*}_p \right|$ and 
$\mathcal{L}_{\text{norm}} = \frac{1}{N} \sum_{p} \left( 1 - \hat{\mathbf{n}}_{p} \cdot \mathbf{n}^{*}_{p} \right).$
Here, $N$ denotes the total number of valid pixels. The primary loss $\mathcal{L}_1$ measures the absolute discrepancy scaled by the predicted aleatoric confidence $c_p$. We complement this with an edge-aware gradient penalty $\mathcal{L}_{\text{grad}}$ to preserve discontinuities at object boundaries, and a normal consistency loss $\mathcal{L}_{\text{norm}}$ that enforces geometric alignment via the cosine similarity between ground-truth orientations $\mathbf{n}^*$ and surface normals $\hat{\mathbf{n}}$, derived analytically from the predicted depth maps.

\noindent\textbf{Surface Normals.}
We instantiate a dedicated, parallel decoding branch for normals. It is initialized with the pre-trained depth weights to benefit from the close connection between depth and normals. The final layer is modified to output three-dimensional unit vectors $\hat{\mathbf{n}}$. Training utilizes a joint objective $\mathcal{L}_{\text{normal}} = \lambda_{\text{cos}} \mathcal{L}_{\text{cos}} + \lambda_{\text{perc}} \mathcal{L}_{\text{perc}}$, which combines a pixel-wise cosine similarity loss with a VGG-based perceptual loss~\cite{simonyan2015deepconvolutionalnetworkslargescale}. The latter serves to prevent over-smoothing and promote sharp edges.

\noindent\textbf{Metric Scale.}
To reconstruct an absolute scale without disrupting the reconstruction of relative local geometry, we decouple metric estimation from the high-resolution, scale-invariant branch. A parallel, coarse decoder predicts a low-resolution metric log-depth map $\hat{z}_{\mathrm{m}}$ alongside an aleatoric confidence map $\hat{c}_{\mathrm{m}}$. From that map, we infer a global scale factor $\hat{\beta}$ as the median difference between the coarse metric log-depth and an average-pooled version of its scale-invariant counterpart, computed over a lower-resolution grid of spatial anchors $\mathbf{a}$:
\begin{equation}
    \hat{\beta} = \mathrm{median}_{\mathbf{a}}\Bigl( \hat{z}_{\mathrm{m}}(\mathbf{a}) - \text{pool}[\hat{z}_{\mathrm{SI}}](\mathbf{a}) \Bigr)
\end{equation}
The median filters out localized geometric discrepancies. The final, absolute metric depth is recovered as $\hat{Z}_{\mathrm{m}} = \exp(\hat{\beta})\hat{Z}_{\mathrm{SI}}$. The metric head is trained with a coverage-weighted version of the confidence-aware $\mathcal{L}_1$ loss against appropriately downsampled ground-truth targets, ensuring that invalid regions do not corrupt the scale estimation.

\noindent\textbf{Sky Segmentation.}
Modeling infinite depth directly destabilizes metric regression. We explicitly decouple unbounded regions by introducing a lightweight sky segmentation branch, such that the primary depth heads can focus on structures with finite depth. The branch reads out geometric cues from intermediate decoder features and fuses them with semantic tokens extracted from the deep encoder layers and passed through a small, fully connected network. The concatenated feature maps are mapped to binary sky probabilities $\hat{Y}$ with a shallow convolutional decoder. This head is trained with a combination of binary cross-entropy, focal~\cite{Lin_2017_ICCV}, and dice losses~\cite{Sudre_2017} w.r.t. the ground-truth mask. Its outputs serve to mask sky regions with undefined geometry in the depth and normal outputs.

\section{Experiments}
\label{sec:experiments}

We evaluate \name{} across diverse quantitative and qualitative experiments on both indoor and outdoor environments. We compare against existing state-of-the-art panoramic geometry estimators and provide detailed ablation studies to isolate and validate the individual structural adaptations and joint training choices of our approach.

\subsection{Training Details}

We initialize our framework from pre-trained DA3 weights~\cite{lin2025depth3recoveringvisual} featuring a DINOv2 backbone~\cite{oquab2024dinov2learningrobustvisual}. Optimization proceeds in two sequential stages. First, we jointly train the scale-invariant depth and surface normal decoders, adapting the backbone features to support both geometric modalities. Second, we freeze these components and independently train the metric scale and sky segmentation heads using the frozen feature representations. We optimize using AdamW~\cite{loshchilov2019decoupledweightdecayregularization} with an exponentially decaying learning rate schedule initialized at $3 \cdot 10^{-4}$ and an Exponential Moving Average decay of 0.999. The first stage requires 12 hours of training on 8 NVIDIA H200 GPUs, while the second stage completes in an additional 8 hours.

\usetikzlibrary{spy,decorations.pathreplacing,angles,quotes,calc,arrows,positioning,}
\definecolor{spycolor}{RGB}{150,150,200}
\tikzset{
    rectspy/.default={lens={scale=3}, size=3cm},
    rectspy on/.style={#1,},
    rectspy/.style={
        draw=spycolor,
        connect spies,
        spy scope={
        every spy on node/.style={
            draw=spycolor,
            very thick,
            rectangle, 
            rectspy on,
        },
        every spy in node/.style={
            draw=spycolor,
            very thick,
            rectangle,
        },
        #1
        },
        spy connection path={\draw[spycolor, very thick] (tikzspyonnode) -- (tikzspyinnode);}
    }
}
\tikzset{
    sepbar/.style={
        very thick,
        black!30!white,
    }
}

\begin{figure*}[!t]
\resizebox{\textwidth}{!}{%
\begin{tikzpicture}[inner sep=0,rectspy={lens={scale=4.25}, width=3.9cm, height=2cm}]

\node[label={[inner sep=0pt,anchor=mid,rounded corners,yshift=12.1em]below:{\large Input}}] at (4*0,0) {\includegraphics[width=3.9cm]{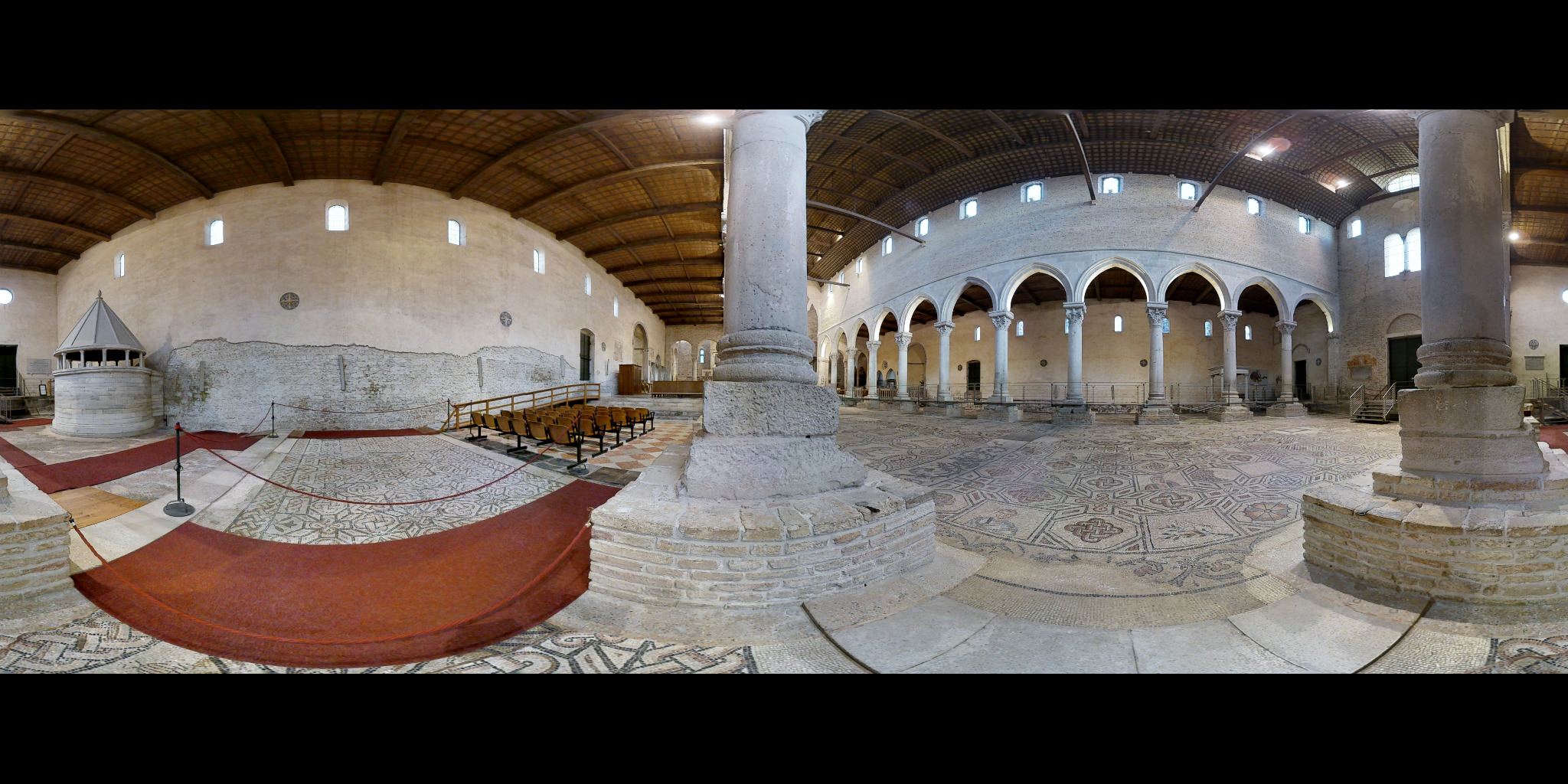}};
\node[label={[inner sep=0pt,anchor=mid,rounded corners,yshift=12.1em]below:{\large DAP}}] at (4*1,0) {\includegraphics[width=3.9cm]{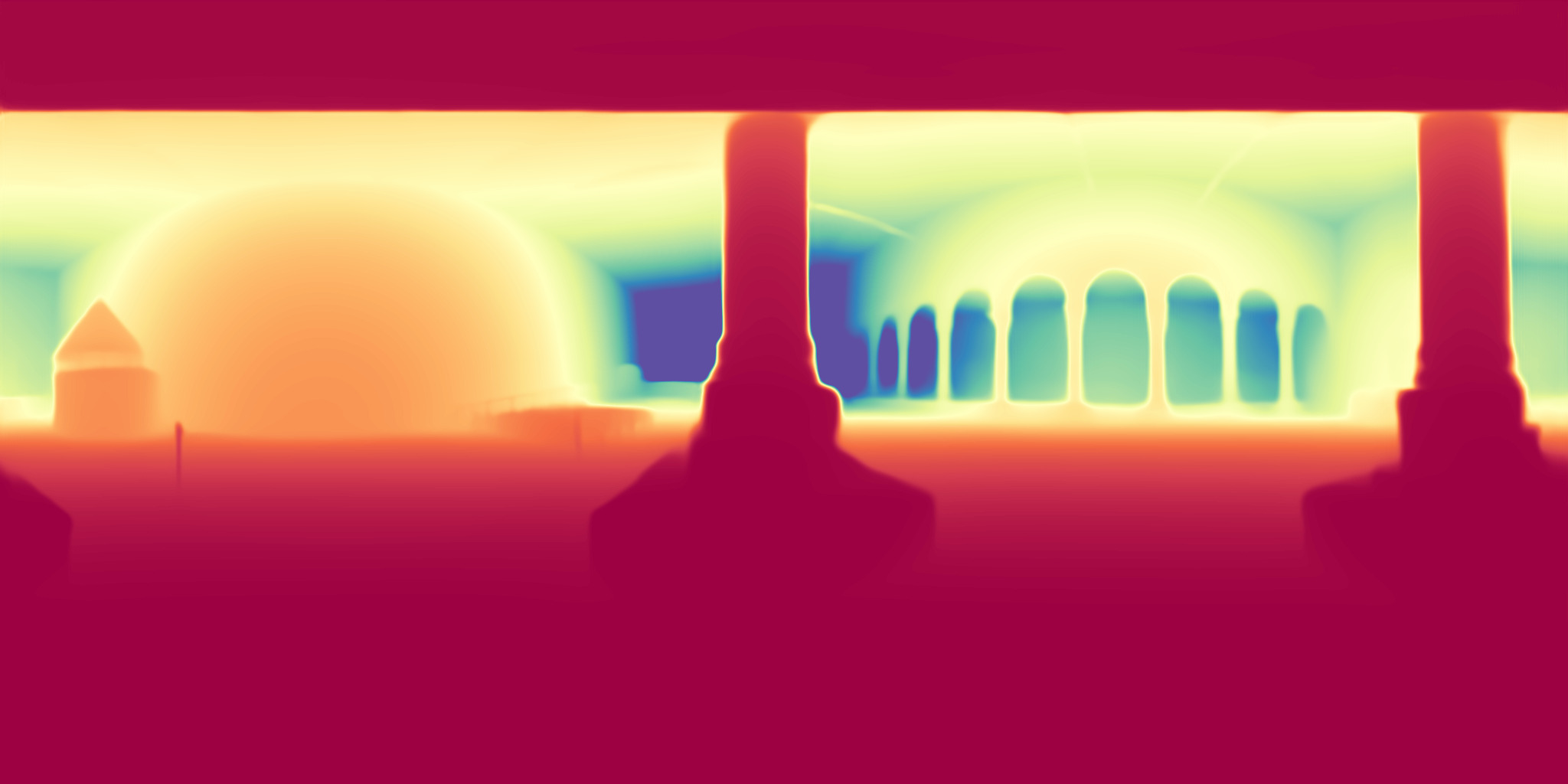}};
\node[label={[inner sep=0pt,anchor=mid,rounded corners,yshift=12.1em]below:{\large $\mathrm{DA}^2$}}] at (4*2,0) {\includegraphics[width=3.9cm]{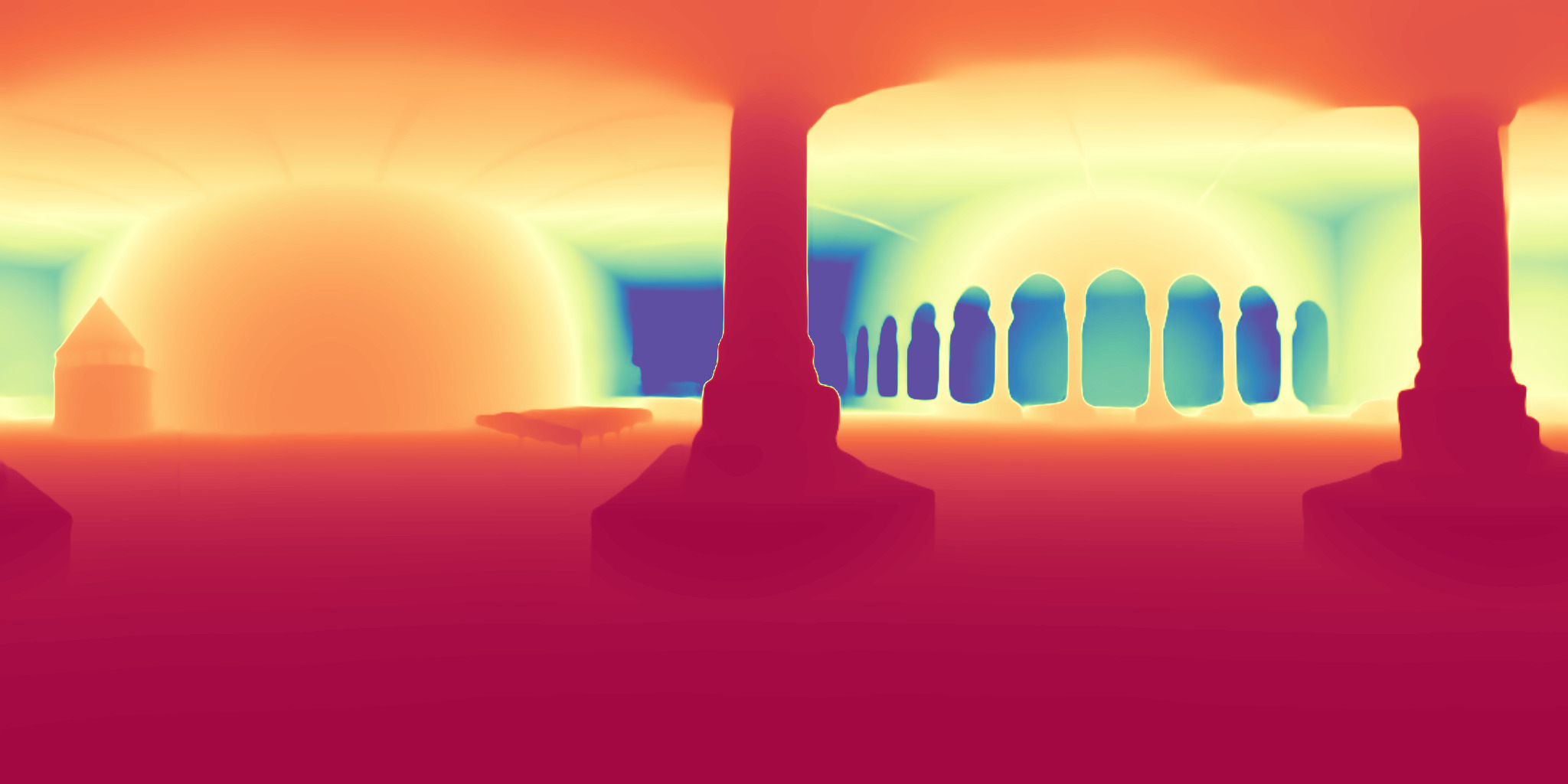}};
\node[label={[inner sep=0pt,anchor=mid,rounded corners,yshift=12.1em]below:{\large Ours}}] at (4*3,0) {\includegraphics[width=3.9cm]{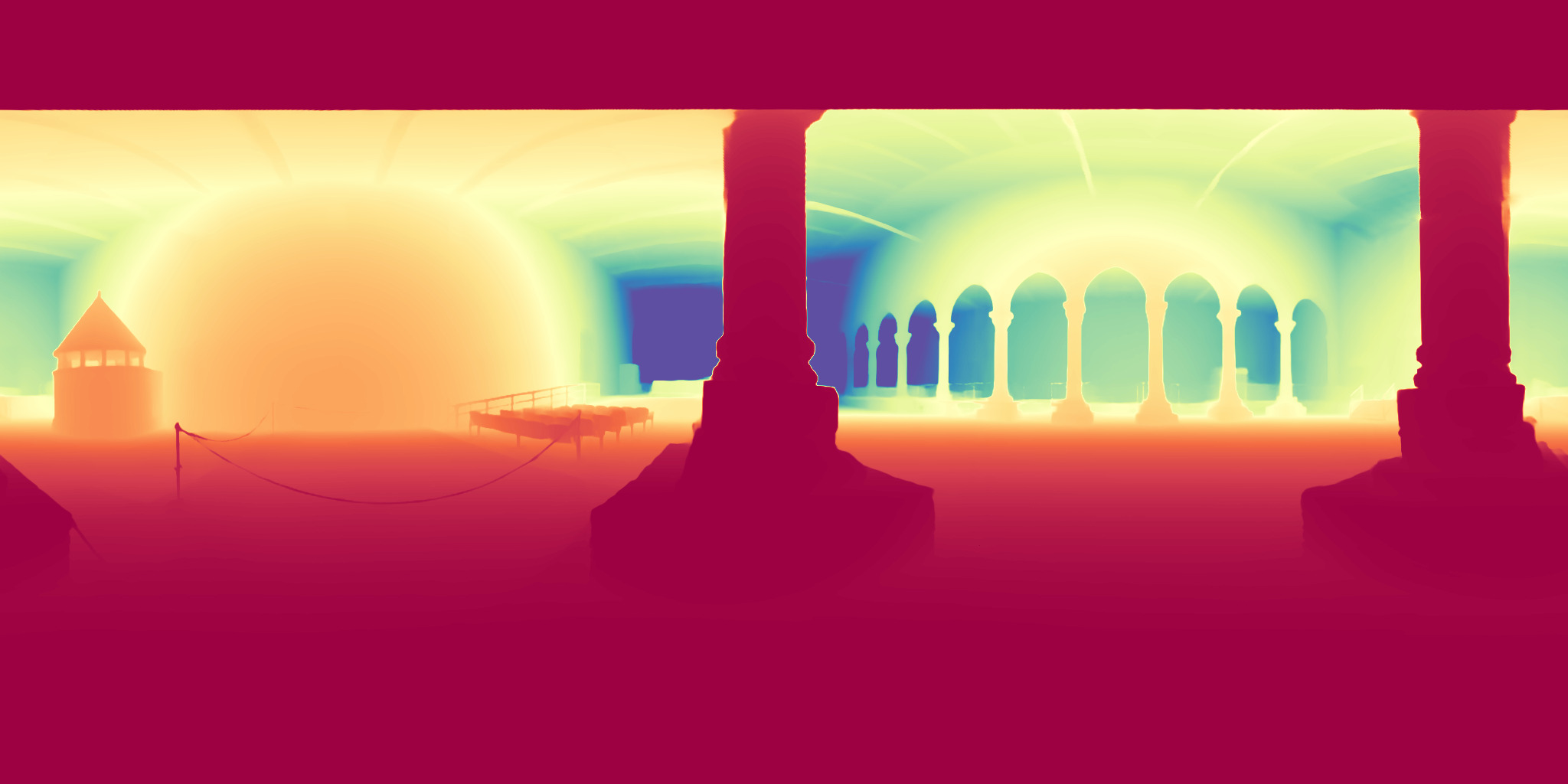}};
\node[label={[inner sep=0pt,anchor=mid,rounded corners,yshift=12.1em]below:{\large Ground Truth}}] at (4*4,0) {\includegraphics[width=3.9cm]{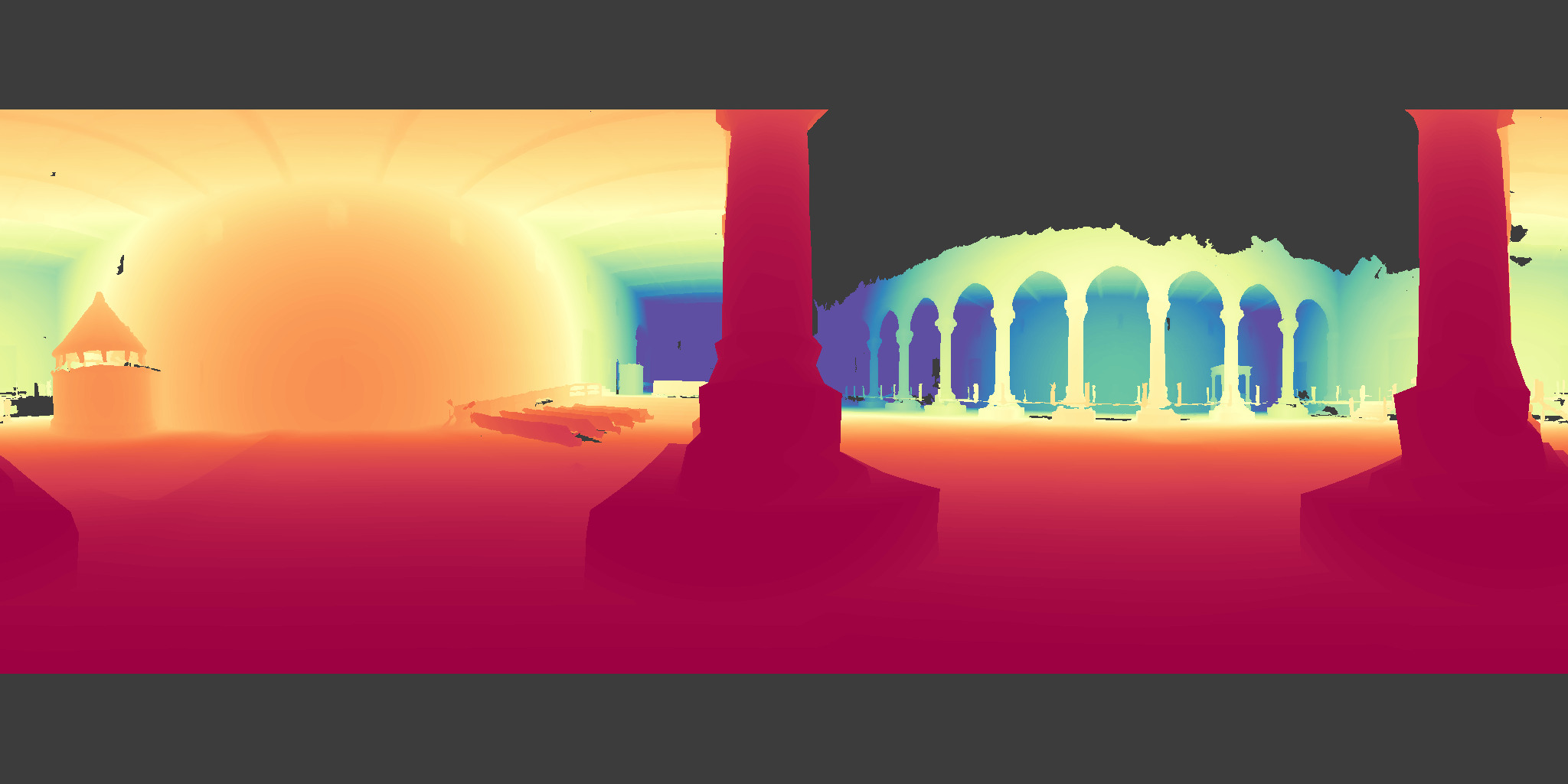}};
\node[rotate=90] at (-2.3,1) {\large Matterport3D360};
\node at (4*0,-4.25) {\includegraphics[width=3.9cm]{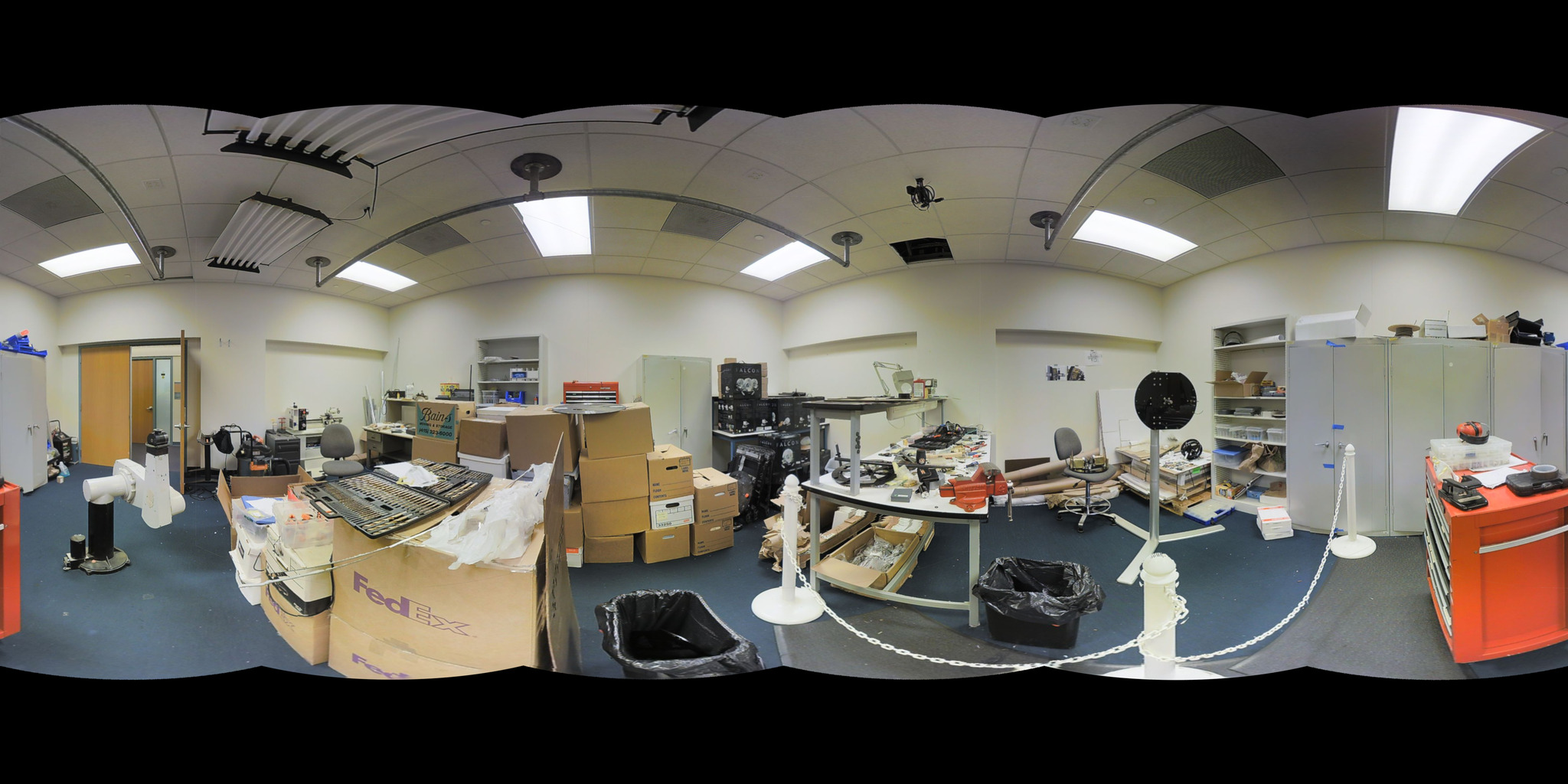}};
\node at (4*1,-4.25) {\includegraphics[width=3.9cm]{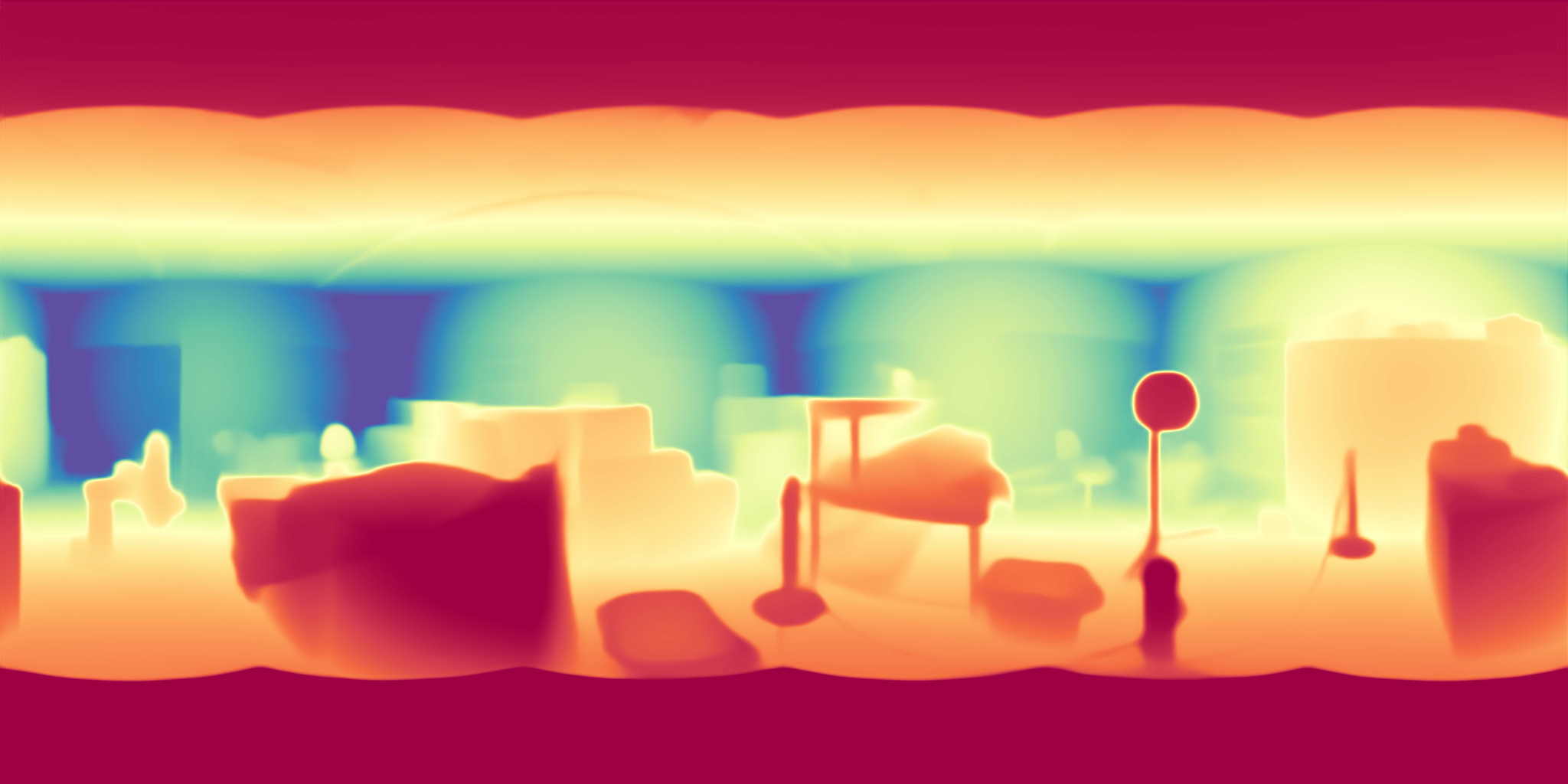}};
\node at (4*2,-4.25) {\includegraphics[width=3.9cm]{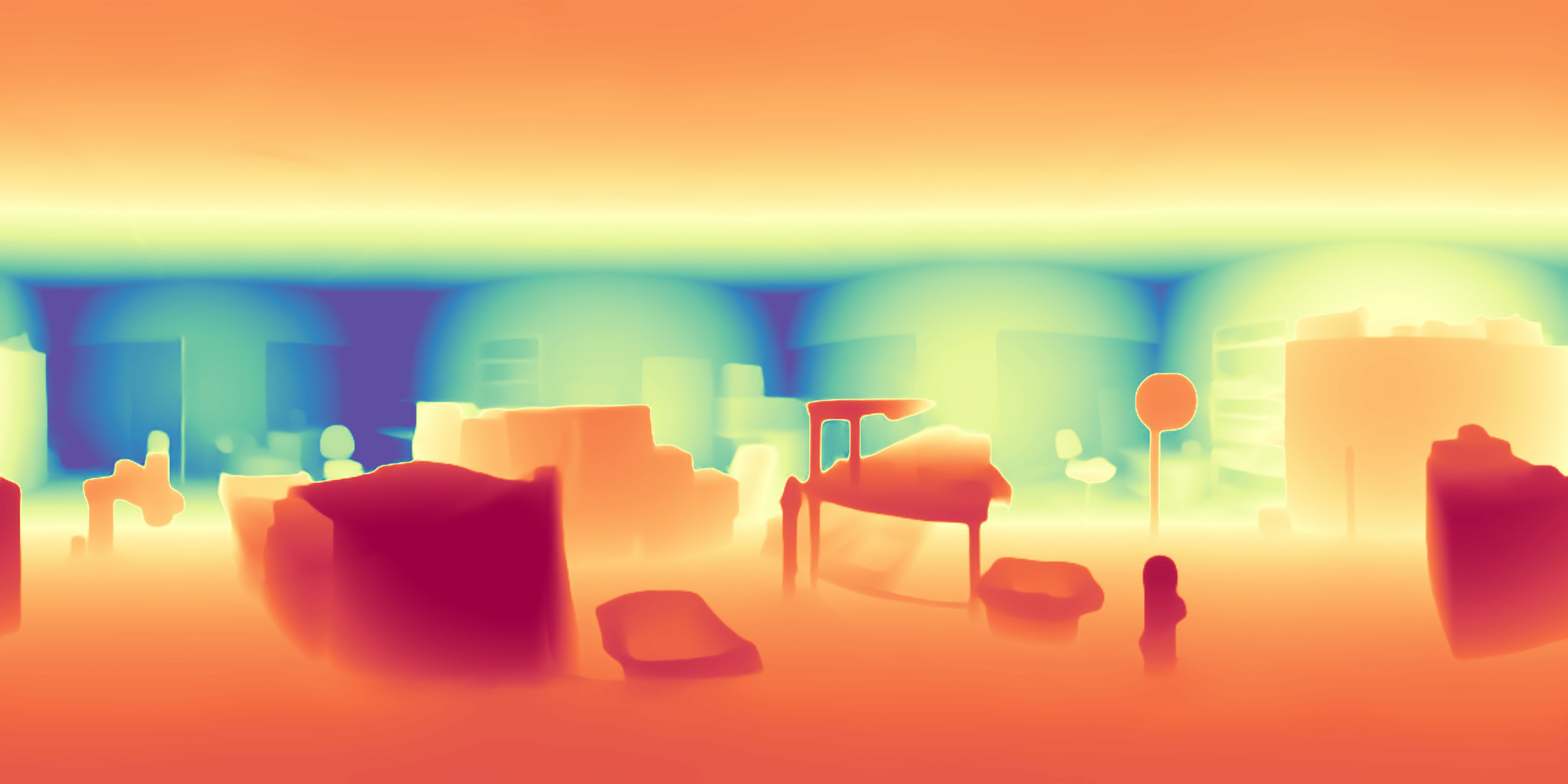}};
\node at (4*3,-4.25) {\includegraphics[width=3.9cm]{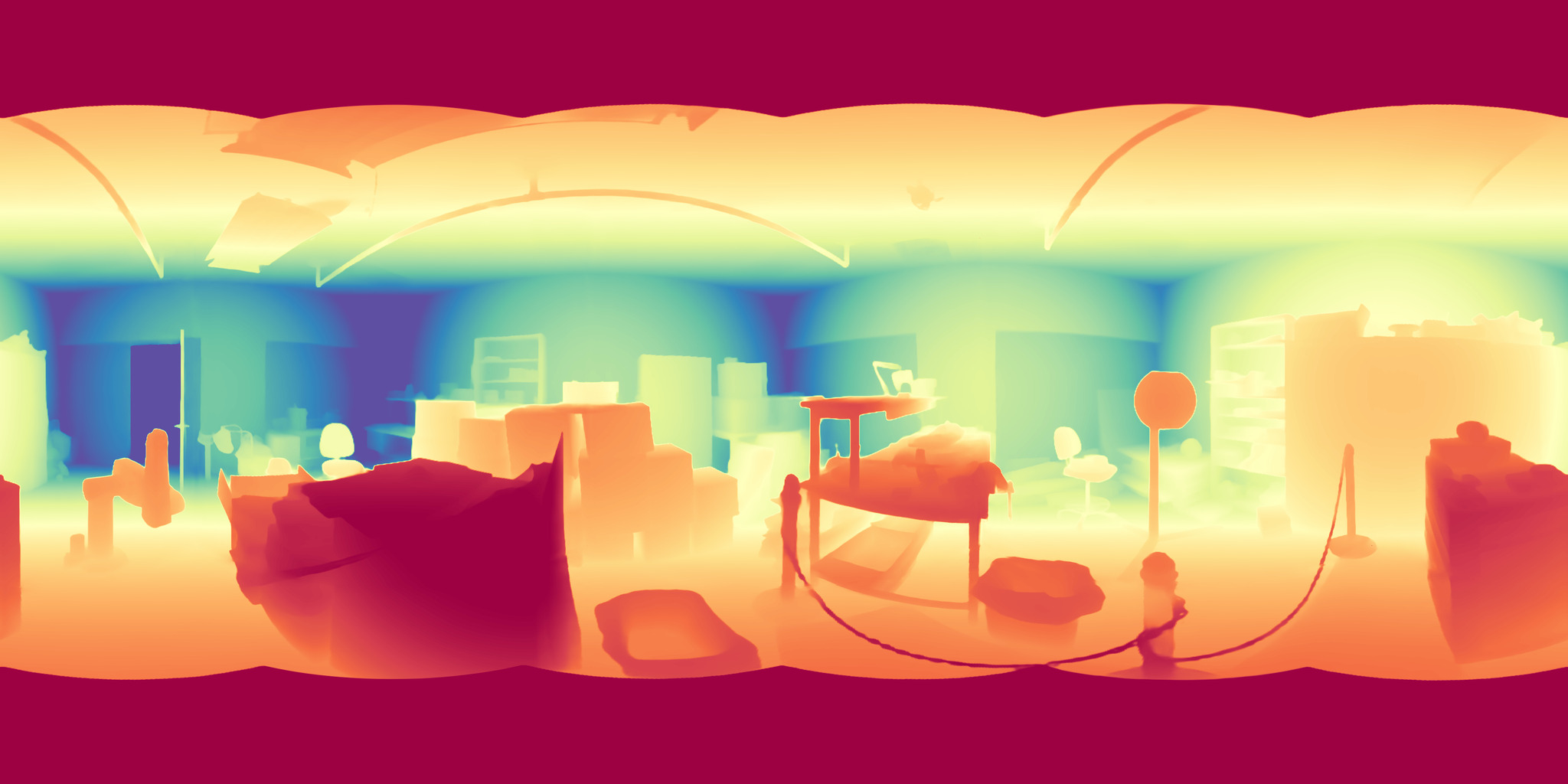}};
\node at (4*4,-4.25) {\includegraphics[width=3.9cm]{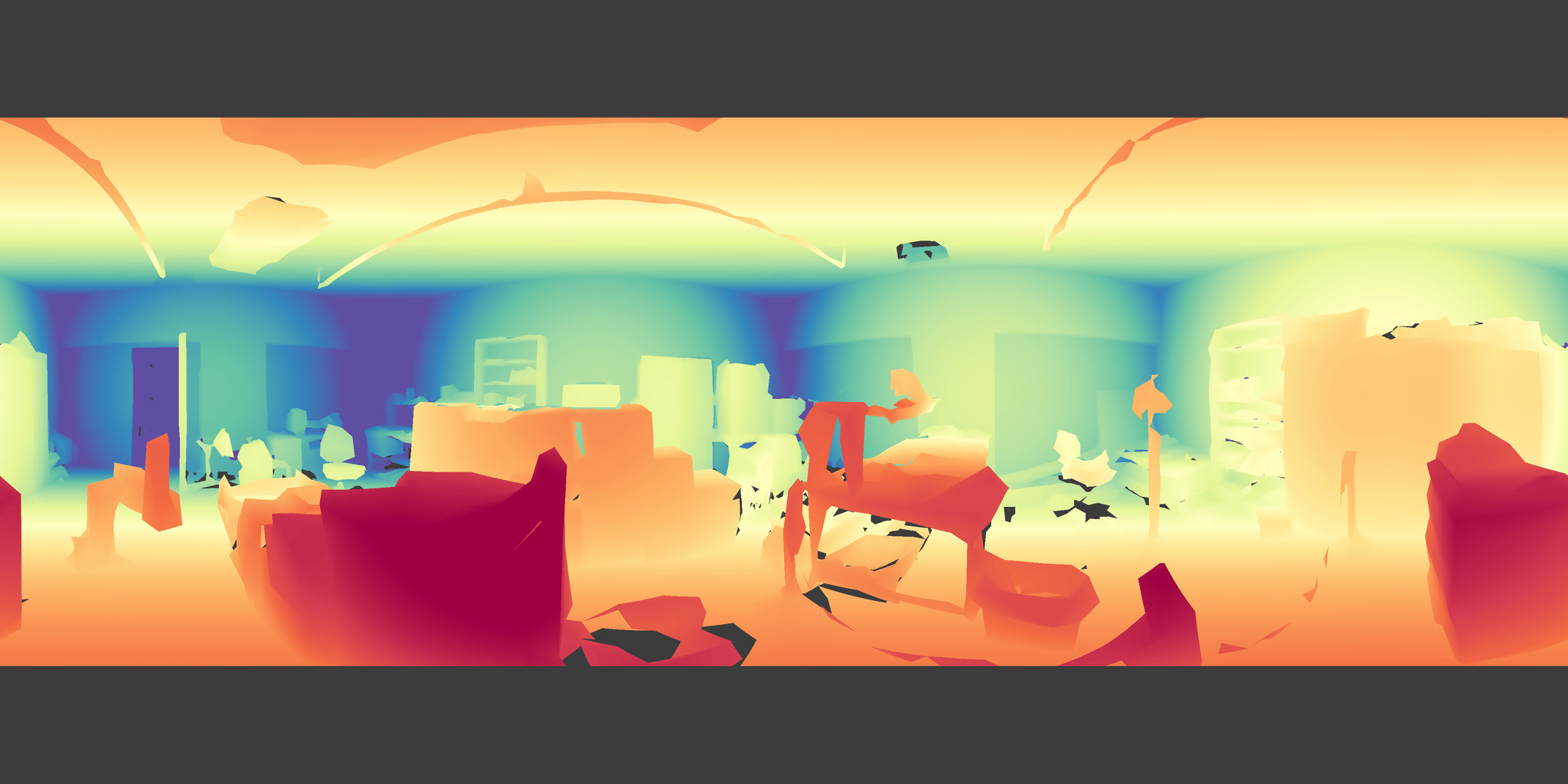}};
\node[rotate=90] at (-2.3,-3.25) {\large Stanford2D3DS};
    \ifthenelse{\equal{rgb.jpg}{BLANK}}{
        \node at (4*0,-8.5)
            [draw=white!40, very thick,
             minimum width=3.9cm, minimum height=2cm]
            {};
    }{
        \node at (4*0,-8.5)
            {\includegraphics[width=3.9cm]{images/depth_comparisons/Franklinstrasse006/rgb.jpg}};
    }
    \ifthenelse{\equal{DAP100.jpg}{BLANK}}{
        \node at (4*1,-8.5)
            [draw=white!40, very thick,
             minimum width=3.9cm, minimum height=2cm]
            {};
    }{
        \node at (4*1,-8.5)
            {\includegraphics[width=3.9cm]{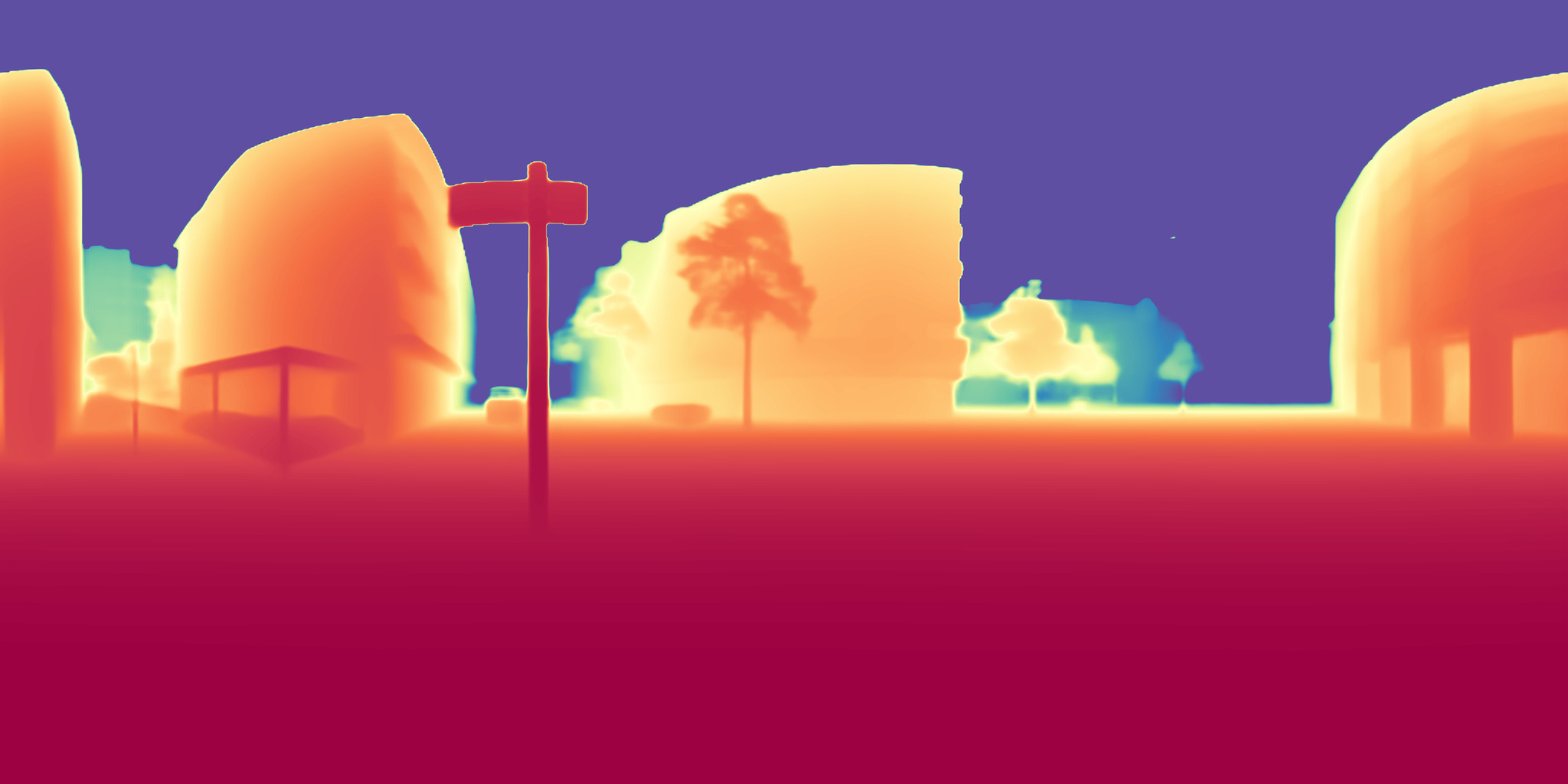}};
    }
    \ifthenelse{\equal{da2.jpg}{BLANK}}{
        \node at (4*2,-8.5)
            [draw=white!40, very thick,
             minimum width=3.9cm, minimum height=2cm]
            {};
    }{
        \node at (4*2,-8.5)
            {\includegraphics[width=3.9cm]{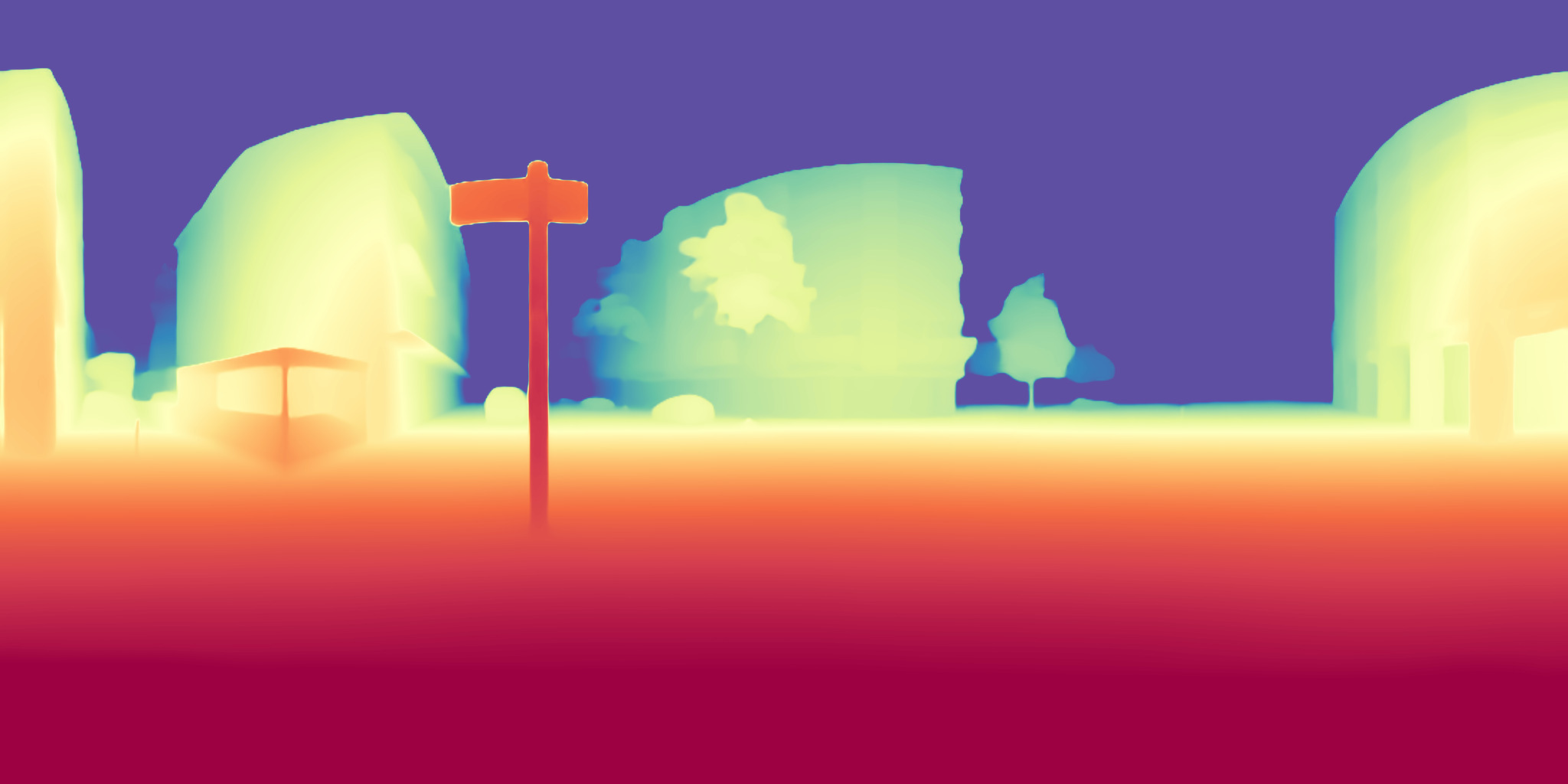}};
    }
    \ifthenelse{\equal{ours.jpg}{BLANK}}{
        \node at (4*3,-8.5)
            [draw=white!40, very thick,
             minimum width=3.9cm, minimum height=2cm]
            {};
    }{
        \node at (4*3,-8.5)
            {\includegraphics[width=3.9cm]{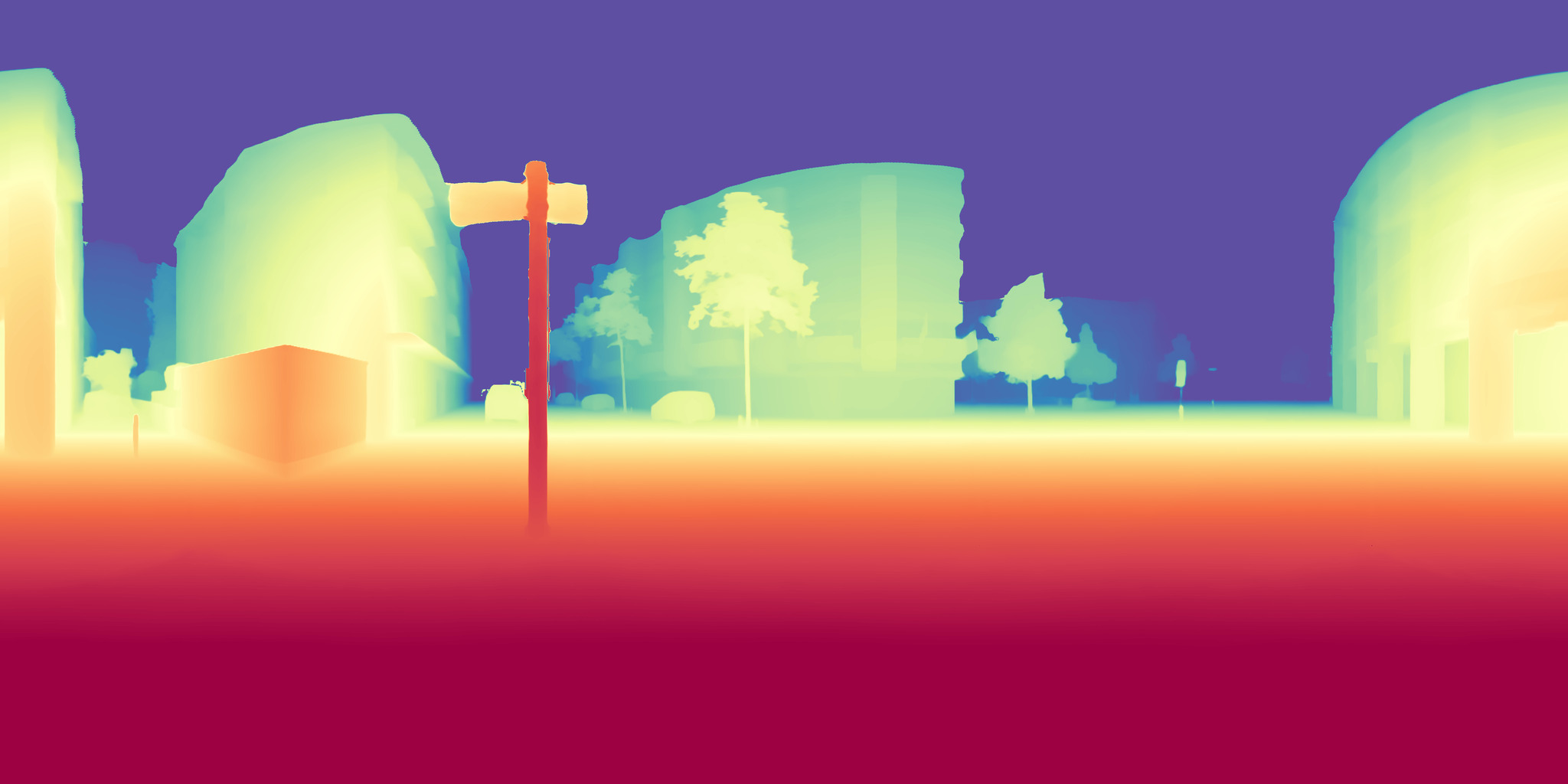}};
    }
    \ifthenelse{\equal{gt.jpg}{BLANK}}{
        \node at (4*4,-8.5)
            [draw=white!40, very thick,
             minimum width=3.9cm, minimum height=2cm]
            {};
    }{
        \node at (4*4,-8.5)
            {\includegraphics[width=3.9cm]{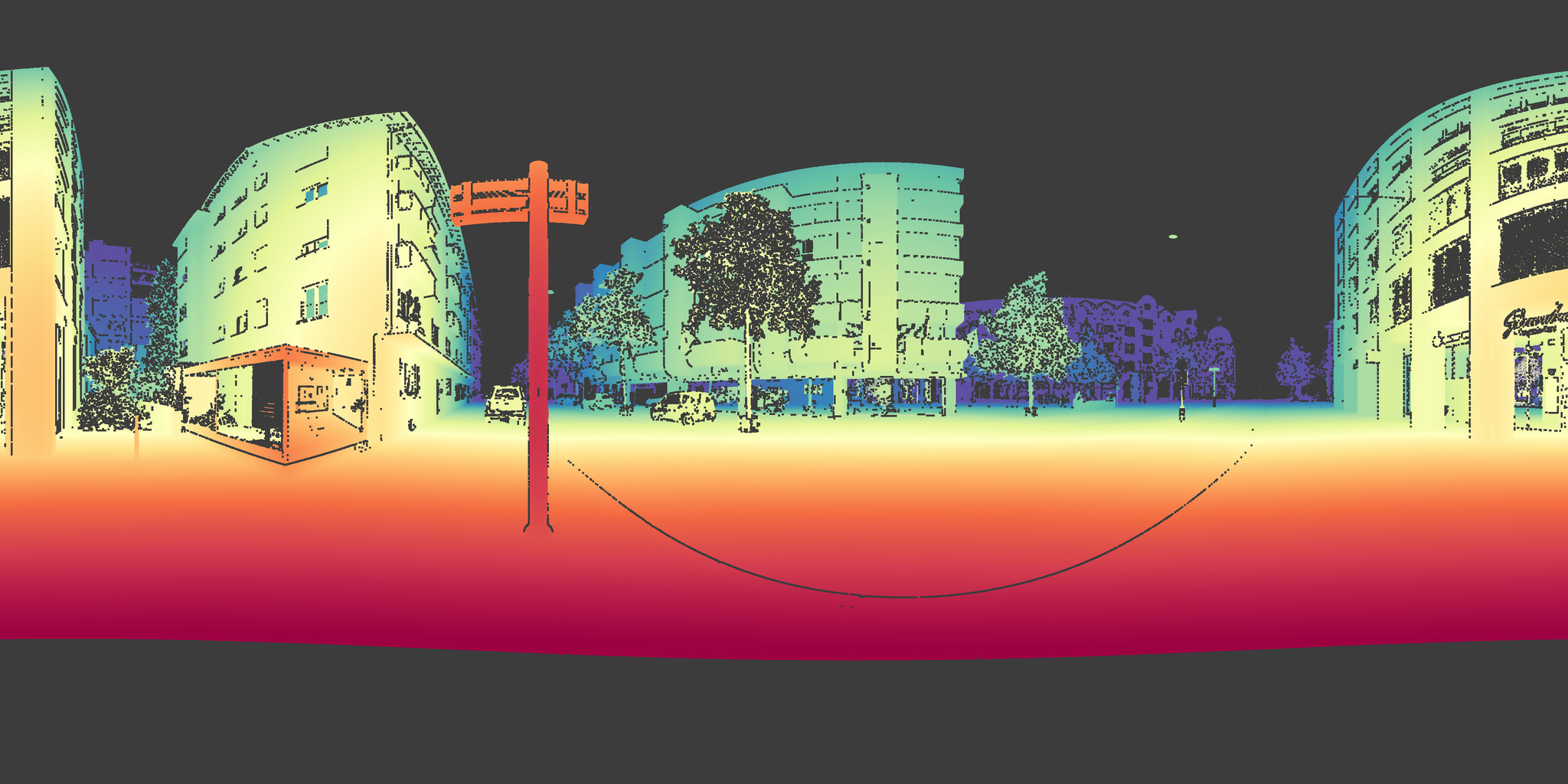}};
    }
\node[rotate=90] at (-2.3,-7.25) {\large ZüriPano};

\spy on (-0.6,-0.1) in node at (0,2);
\spy on (3.4,-0.1) in node at (4,2);
\spy on (7.4,-0.1) in node at (8,2);
\spy on (11.4,-0.1) in node at (12,2);
\spy on (15.4,-0.1) in node at (16,2);

\spy on (0.5,-4.37) in node at (0,-2.25);
\spy on (4.5,-4.37) in node at (4,-2.25);
\spy on (8.5,-4.37) in node at (8,-2.25);
\spy on (12.5,-4.37) in node at (12,-2.25);
\spy on (16.5,-4.37) in node at (16,-2.25);

\spy on (0.0,-8.5) in node at (0,-6.5);
\spy on (4.0,-8.5) in node at (4,-6.5);
\spy on (8.0,-8.5) in node at (8,-6.5);
\spy on (12.0,-8.5) in node at (12,-6.5);
\spy on (16.0,-8.5) in node at (16,-6.5);

\end{tikzpicture}
}
\vspace{-.5cm}
\caption{\textbf{Qualitative comparison of panoramic depth estimation.} Visual results from \name{}, DAP~\cite{lin2025depthpanoramasfoundationmodel}, and $\mathrm{DA}^2$~\cite{li20252} (the strongest metric and scale-invariant baselines) alongside the RGB input and ground-truth depth on Matterport3D360, Stanford2D3DS, and ZüriPano. Our framework recovers sharper boundaries and more accurate global structures than competitors. Additional examples are in the appendix. Best viewed zoomed in.}
\label{fig:comparison}
\end{figure*}
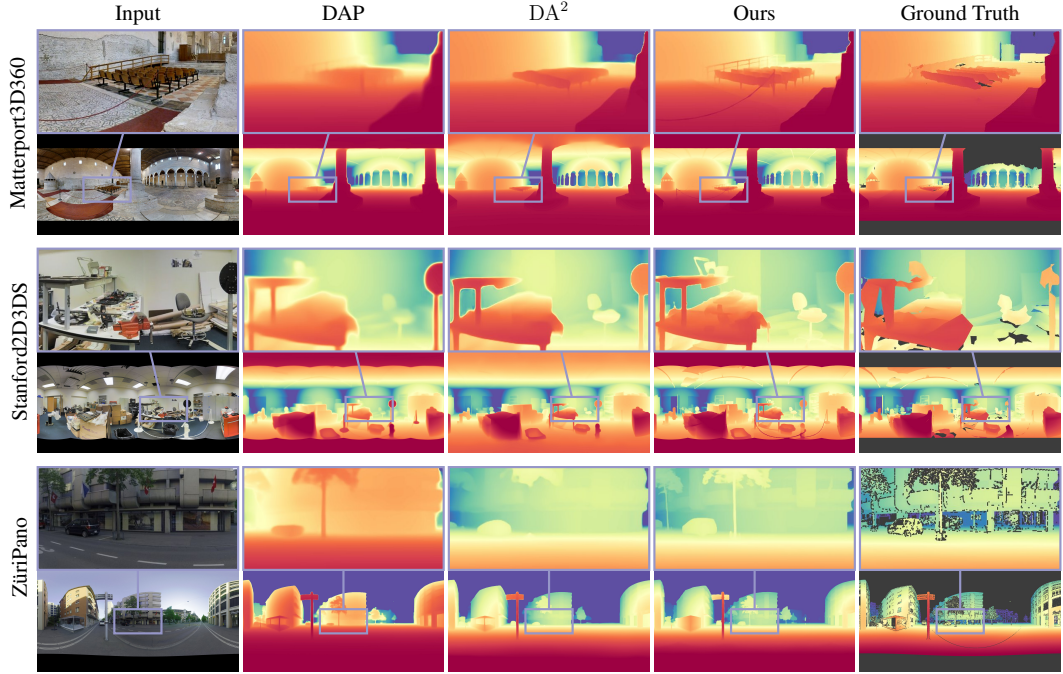
\begin{table*}[th]
\centering
\small
\renewcommand{\arraystretch}{1.1}
\setlength{\tabcolsep}{3pt}

\resizebox{\textwidth}{!}{%
\begin{tabular}{l ccc c ccc c ccc c ccc}
    \specialrule{0.75pt}{0pt}{1pt}
    & \multicolumn{3}{c}{\textbf{Matterport3D360}} & & \multicolumn{3}{c}{\textbf{Stanford2D3DS}} & & \multicolumn{3}{c}{\textbf{ZüriPano}} & & \multicolumn{3}{c}{\textbf{Holo360D}} \\
    \hhline{~---~---~---~---}
    Method & AbsRel $\downarrow$ & RMSE $\downarrow$ & $\delta_1$ $\uparrow$ & & AbsRel $\downarrow$ & RMSE $\downarrow$ & $\delta_1$ $\uparrow$ & & AbsRel $\downarrow$ & RMSE $\downarrow$ & $\delta_1$ $\uparrow$ & & AbsRel $\downarrow$ & RMSE $\downarrow$ & $\delta_1$ $\uparrow$ \\
    \hline
    DreamCube*~\cite{huang2025dreamcube}              & 30.45 & 108.81 & 54.26 & & 28.45 & \phantom{0}75.30 & 56.39 & & 29.57 & 484.91 & 53.07 & & -- & -- & -- \\
    DepthAnyCamera~\cite{guo2025depth}                & 25.62 & \phantom{0}94.06 & 68.11 & & 18.35 & \phantom{0}53.73 & 76.23 & & 21.98 & 487.66 & 65.41 & & -- & -- & -- \\
    MoGe~\cite{wang2025moge}                          & 18.12 & \phantom{0}81.68 & 77.11 & & 18.12 & \phantom{0}81.68 & 77.11 & & 19.25 & 484.18 & 77.12 & & -- & -- & -- \\
    EGformer*~\cite{yun2023egformer}                  & 16.74 & \phantom{0}96.15 & 79.32 & & 13.64 & \phantom{0}58.95 & 84.64 & & 55.49 & 721.83 & 31.21 & & -- & -- & -- \\
    DAP~\cite{lin2025depthpanoramasfoundationmodel}   & 15.84 & \phantom{0}85.18 & 82.44 & & \phantom{0}9.76 & \phantom{0}46.69 & 92.37 & & 19.86 & 583.42 & 72.09 & & 42.70 & 8.44 & 34.30 \\
    RPG360~\cite{jung2025rpg360robust360depth}        & 15.40 & \phantom{0}79.40 & 82.40 & & 11.91 & \phantom{0}46.60 & 87.73 & & \underline{18.27} & \underline{455.28} & \underline{78.41} & & -- & -- & -- \\
    UniK3D$^\dagger$~\cite{piccinelli2025unik3d}      & 14.82 & \phantom{0}69.27 & 83.67 & & \phantom{0}9.93 & \phantom{0}40.69 & 93.34 & & 31.00 & 832.13 & 38.78 & & 53.80 & 8.70 & 23.70 \\
    PanDA*~\cite{cao2025panda}                        & 13.84 & \phantom{0}82.82 & 84.26 & & 10.69 & \phantom{0}55.72 & 90.73 & & 24.64 & 515.48 & 58.31 & & \underline{39.90} & \underline{6.16} & \underline{45.90} \\
    $\mathrm{DA}^2$~\cite{li20252}                    & \underline{11.06} & \phantom{0}\underline{67.72} & \underline{89.18} & & \phantom{0}\underline{7.64} & \phantom{0}\underline{37.29} & \underline{95.63} & & 61.22 & 869.08 & \phantom{0}2.17 & & 51.60 & 7.34 & 11.10 \\
    \rowcolor{gray!15} \textbf{PaGeR}          & \phantom{0}\textbf{9.67} & \phantom{0}\textbf{64.69} & \textbf{90.87} & & \phantom{0}\textbf{5.93} & \phantom{0}\textbf{35.34} & \textbf{96.10} & & \phantom{0}\textbf{9.36} & \textbf{299.61} & \textbf{94.75} & & \textbf{18.70} & \textbf{3.54} & \textbf{83.00} \\
\specialrule{0.75pt}{1pt}{1pt}
\end{tabular}
}
\caption{\textbf{Quantitative comparison between \name{} and state-of-the-art baselines across indoor (Matterport3D360, Stanford2D3DS) and outdoor (ZüriPano, Holo360D)}. Best and second-best results are indicated in \textbf{bold} and \underline{underlined}, respectively. Methods optimized using in-domain training are marked with $\dagger$, and affine-invariant methods are denoted with \textsuperscript{*}.}\label{tab:depth_results}
\end{table*}

Our mixed data regime balances 80k synthetic panoramas from Structured3D~\cite{Structured3D} and our PanoInfinigen dataset with 10k real perspective images from ScanNet++~\cite{yeshwanth2023scannet++} and ARKitScenes~\cite{baruch2021arkitscenes} to mitigate the sim-to-real domain gap. Following standard practice~\cite{lin2025depthpanoramasfoundationmodel, wen2025metricsolverslidinganchoredmetric}, we train independent metric scale heads for indoor and outdoor environments to accommodate distinct spatial layouts. We maintain a training resolution of $504 \times 504$ pixels per cubemap face, which assembles into a 2K equirectangular panorama. At inference time, our unified framework processes a full 2K panorama in 0.5 seconds while consuming 12.8 GB of memory, allowing for deployment on a single consumer-grade GPU. The choices of hyperparameters are given in Tab.~\ref{tab:hyperparam}.

\subsection{Evaluation Protocol}

\noindent\textbf{Datasets and Evaluation Ranges.} Consistent with panoramic benchmarks~\cite{cao2025panda, wang2024depth, li20252}, we evaluate scale-invariant and metric depth on the real-world indoor datasets Matterport3D360~\cite{bata1126} and Stanford2D3DS~\cite{https://doi.org/10.57761/gmhc-wx10}. To address the indoor bias of existing literature, we also introduce ZüriPano, a custom outdoor urban LiDAR dataset tailored for long-range geometric evaluation (see Sec.~\ref {sec:zuripano}). Additionally, we evaluate on Holo360D~\cite{ou2026holo360dlargescalerealworlddataset}, a synthetic outdoor dataset. For all depth evaluations, we enforce a broad range constraint of $[0, 75]\,\text{m}$ to ensure a thorough assessment of global structures. This avoids the evaluation bias of prior works~\cite{cao2025panda, li20252} that use a narrow $[0, 5]\,\text{m}$ window, masking far-field errors. For completeness, we report the results for the near-field depth evaluation in the supplementary ~\ref{sec:additional_quantitative_metrics}. For surface orientation, we benchmark on the Structured3D dataset~\cite{Structured3D}, where all compared baselines are trained to guarantee a fair comparison.

\begin{table*}[!t]
\centering
\small
\renewcommand{\arraystretch}{1.1}
\setlength{\tabcolsep}{3pt}

\resizebox{\textwidth}{!}{%
\begin{tabular}{l ccc c ccc c ccc c ccc}
    \specialrule{0.75pt}{0pt}{1pt}
    & \multicolumn{3}{c}{\textbf{Matterport3D360}} & & \multicolumn{3}{c}{\textbf{Stanford2D3DS}} & & \multicolumn{3}{c}{\textbf{ZüriPano}} & & \multicolumn{3}{c}{\textbf{Holo360D}} \\
    \hhline{~---~---~---~---}
    Method & AbsRel $\downarrow$ & RMSE $\downarrow$ & $\delta_1$ $\uparrow$ & & AbsRel $\downarrow$ & RMSE $\downarrow$ & $\delta_1$ $\uparrow$ & & AbsRel $\downarrow$ & RMSE $\downarrow$ & $\delta_1$ $\uparrow$ & & AbsRel $\downarrow$ & RMSE $\downarrow$ & $\delta_1$ $\uparrow$ \\
    \hline
    UniK3D$^\dagger$~\cite{piccinelli2025unik3d}      & 33.43 & 143.24 & 46.48 & & 24.37 & \phantom{0}64.71 & 64.50 & & 35.81 & 1345.48 & 34.86 & & 51.30 & 9.37 & 28.50 \\
    RPG360~\cite{jung2025rpg360robust360depth}        & 29.26 & 136.99 & 35.33 & & 29.35 & \phantom{0}90.73 & 17.36 & & 40.36 & \phantom{0}772.16 & \phantom{0}2.14 & & -- & -- & -- \\
    DepthAnyCamera~\cite{guo2025depth}                & \underline{25.16} & \underline{132.66} & 62.25 & & 17.15 & \phantom{0}59.29 & 77.98 & & \underline{33.23} & \phantom{0}\underline{716.38} & \underline{37.83} & & \underline{27.40} & \underline{6.79} & \underline{61.20} \\
    DAP\textsuperscript{\ddag}~\cite{lin2025depthpanoramasfoundationmodel} & 30.16 & 168.88 & \textbf{80.19} & & \underline{10.97} & \phantom{0}\underline{53.39} & \underline{90.64} & & 47.00 & \phantom{0}839.72 & 24.02 & & 59.30 & 11.05 & 56.90 \\
    \rowcolor{gray!15} \textbf{PaGeR\textsuperscript{\ddag}} & \textbf{21.83} & \textbf{123.48} & \underline{69.50} & & \textbf{10.94} & \textbf{45.43} & \textbf{90.94} & & \textbf{31.97} & \phantom{0}\textbf{530.85} & \textbf{39.30} & & \textbf{23.50} & \textbf{4.04} & \textbf{72.00} \\
\specialrule{0.75pt}{1pt}{1pt}
\end{tabular}
}
\caption{\textbf{Quantitative comparison of panoramic metric depth models}. Methods optimized using in-domain training are marked with $\dagger$, while those using separate indoor or outdoor prediction heads are marked with \textsuperscript{ $\ddagger$}. \textbf{Bold} font marks the best result, \underline{underlined} the second-best.}
\label{tab:metric_depth}
\vspace{-1em}
\end{table*}

\noindent\textbf{Metrics and Processing.} Following established conventions~\cite{yang2024depth}, depth accuracy is measured via Absolute Relative Error (AbsRel), Root Mean Squared Error (RMSE), and the threshold percentage $\delta_1$. Scale-invariant depth maps are adjusted using a standard least-squares alignment prior to scoring, whereas metric depth is evaluated directly without modifications. For surface normals, we report the Mean Angular Error, Mean Squared Error (MSE), and the fraction of pixels with errors below $\delta_{\theta} \in \{5^\circ, 22.5^\circ\}$~\cite{huang2024panonormal}, with all predictions normalized to unit length before evaluation.
All reported numbers for competing methods have been recalculated to evaluate at the native resolution of each dataset; furthermore, to ensure fairness to low-resolution methods, we additionally evaluate in a low-resolution setting of $512 \times 1024$ by downsampling the ground truth, with these additional results provided in the supplementary ~\ref{sec:additional_quantitative_metrics}. The same ground-truth sky masks are used throughout the whole evaluation, for each evaluated method.

\subsection{Quantitative Comparison} 

As demonstrated in \Cref{tab:depth_results}, \name{} consistently outperforms existing methods across all datasets. While it demonstrates notable improvements on indoor-biased benchmarks, its primary advantage lies in its cross-domain generalization. On the challenging outdoor ZüriPano dataset, \name{} reduces the Absolute Relative Error (AbsRel) from the previous best 18.27 for RPG360 to 9.36, nearly cutting it in half. This substantial improvement confirms that our framework enhances structural geometry globally and does not need to trade off indoor vs.\ outdoor accuracy.

This balanced capability extends directly to absolute scale recovery, as detailed in Table~\ref{tab:metric_depth}. 
By decoupling metric scale estimation from the structural backbone, our independent domain heads successfully specialize to indoor or outdoor structures while sharing the same underlying transformer features. On the ZüriPano metric benchmark, \name{} establishes a commanding lead with an RMSE of 530.85 compared to 716.38 for the next best, DepthAnyCamera. At the same time, it maintains high accuracy indoors, outperforming recent baselines such as UniK3D and DAP on both indoor datasets.
\begin{figure}[!b]
\centering
\begin{minipage}[b]{0.48\textwidth}
    \centering
    \small
    \setlength{\tabcolsep}{1.2pt} 
    \renewcommand{\arraystretch}{1.0} 
    \begin{tabular}{lcccc}
        \specialrule{0.75pt}{0pt}{1pt}
        Method & Mean $\downarrow$ & MSE $\downarrow$ & $\delta_{5^\circ}$ $\uparrow$ & $\delta_{22.5^\circ}$ $\uparrow$ \\
        \hline
        UniFuse~\cite{jiang2021unifuse} & \phantom{0}8.25 & \phantom{0}453.1 & 76.24 & 87.55 \\
        PanoFormer~\cite{shen2022panoformer} & 16.92 & 1053.5 & 59.13 & 75.50 \\
        OmniFusion~\cite{li2022omnifusion} & 20.70 & \phantom{0}832.0 & 28.51 & 63.55 \\
        MonoViT~\cite{Zhao_2022} & \phantom{0}5.92 & \phantom{0}277.8 & 78.93 & 90.58 \\
        HyperSphere~\cite{karakottas2019360osurfaceregressionhypersphere} & \phantom{0}5.79 & \phantom{0}253.4 & 78.38 & 90.73 \\
        MTL~\cite{huang2024multi} & \phantom{0}8.98 & \phantom{0}469.0 & 72.51 & 86.02 \\
        PanoNormal~\cite{shen2022panoformer} & \phantom{0}5.56 & \phantom{0}246.6 & 79.18 & 91.01 \\
        \rowcolor{gray!15} \textbf{PaGeR} & \phantom{0}\textbf{5.49} & \phantom{0}\textbf{174.9} & \textbf{79.91} & \textbf{92.83} \\
        \specialrule{0.75pt}{1pt}{1pt}
    \end{tabular}
        
    \captionof{table}{\textbf{Quantitative comparison of surface normals on the Structured3D dataset}. All evaluated methods are in-domain architectures optimized directly on Structured3D. 
    }
    \label{tab:normals}
\end{minipage}
\hfill
\begin{minipage}[b]{0.48\textwidth}
\centering

\begin{minipage}{0.24\linewidth}\centering \textbf{\tiny RGB}\end{minipage}\hfill
\begin{minipage}{0.24\linewidth}\centering \textbf{\tiny MTL}\end{minipage}\hfill
\begin{minipage}{0.24\linewidth}\centering \textbf{\tiny Ours}\end{minipage}\hfill
\begin{minipage}{0.24\linewidth}\centering \textbf{\tiny GT}\end{minipage}

\vspace{0.2em}

\includegraphics[width=0.24\linewidth]{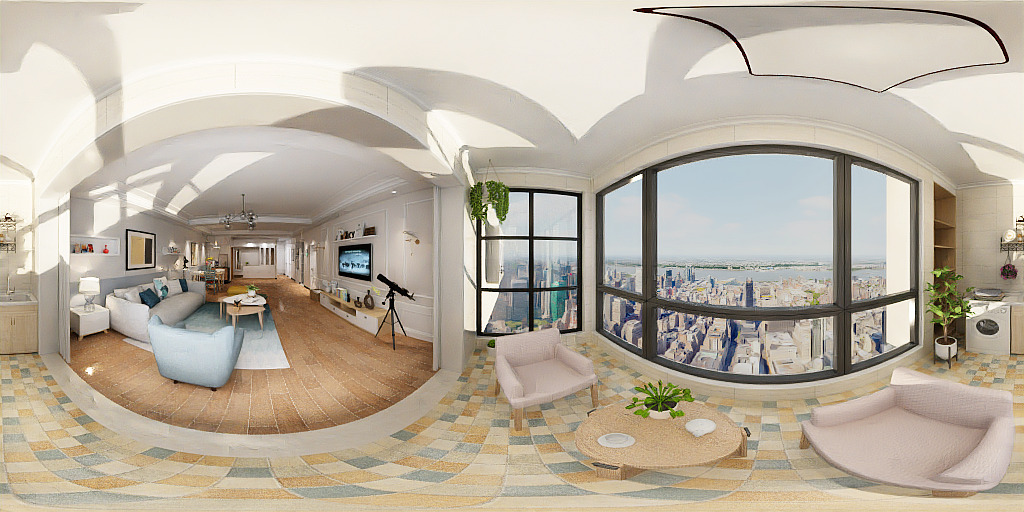}\hfill
\includegraphics[width=0.24\linewidth]{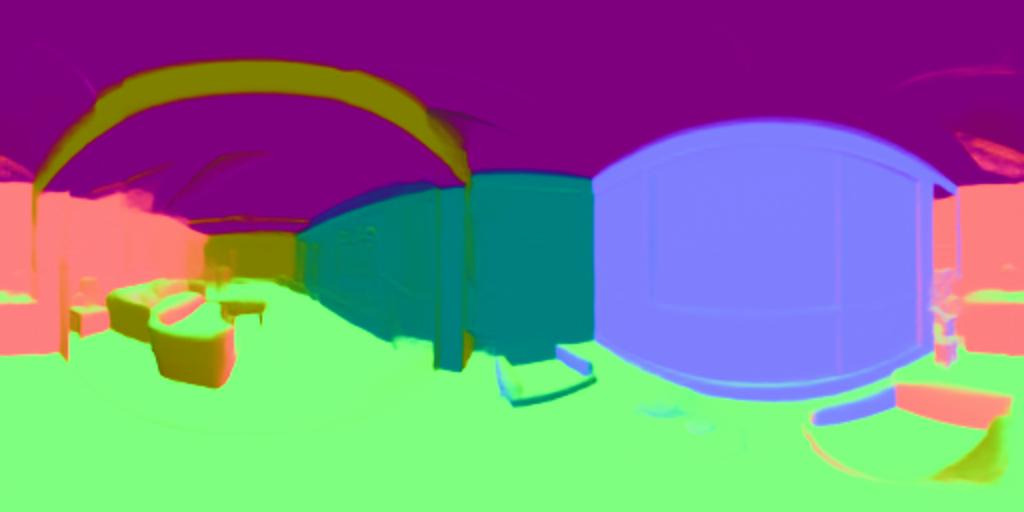}\hfill
\includegraphics[width=0.24\linewidth]{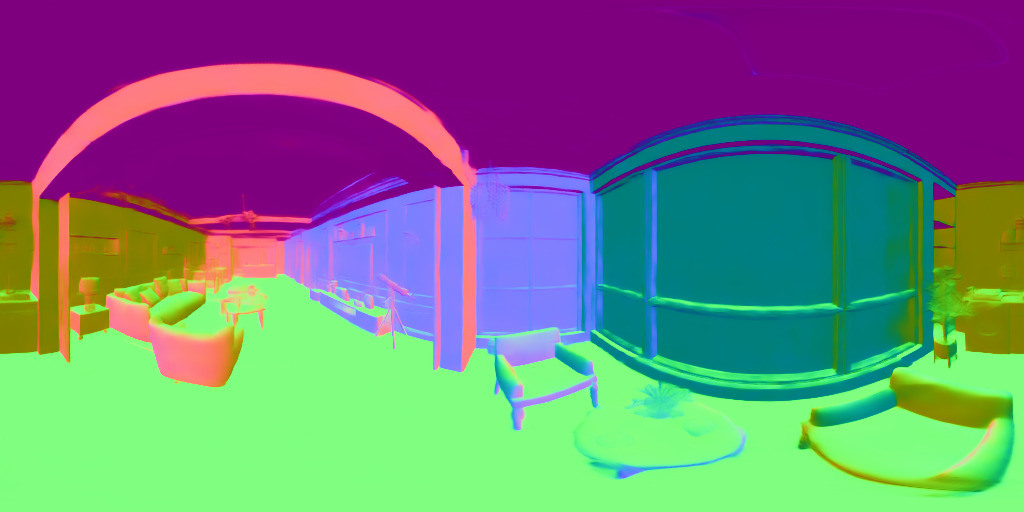}\hfill
\includegraphics[width=0.24\linewidth]{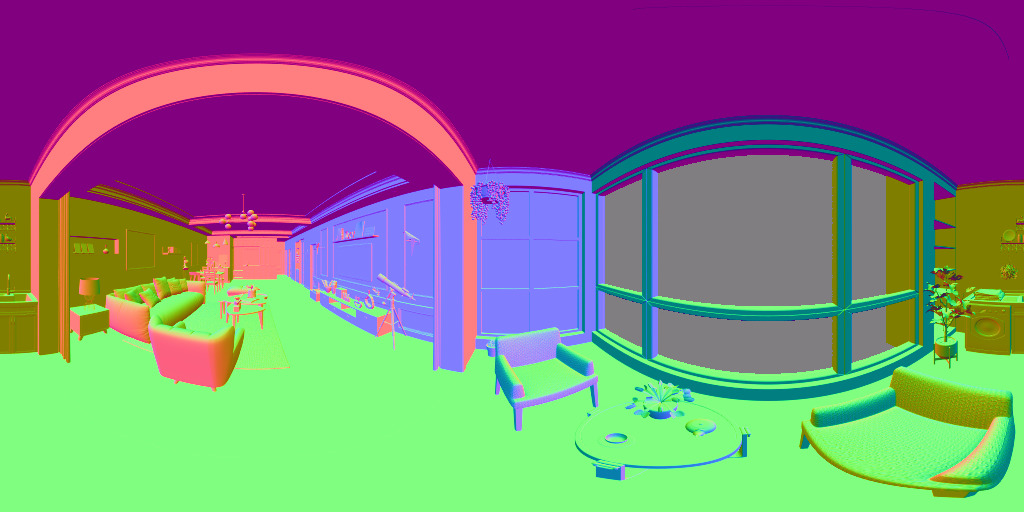}
\vspace{0.1em}

\includegraphics[width=0.24\linewidth]{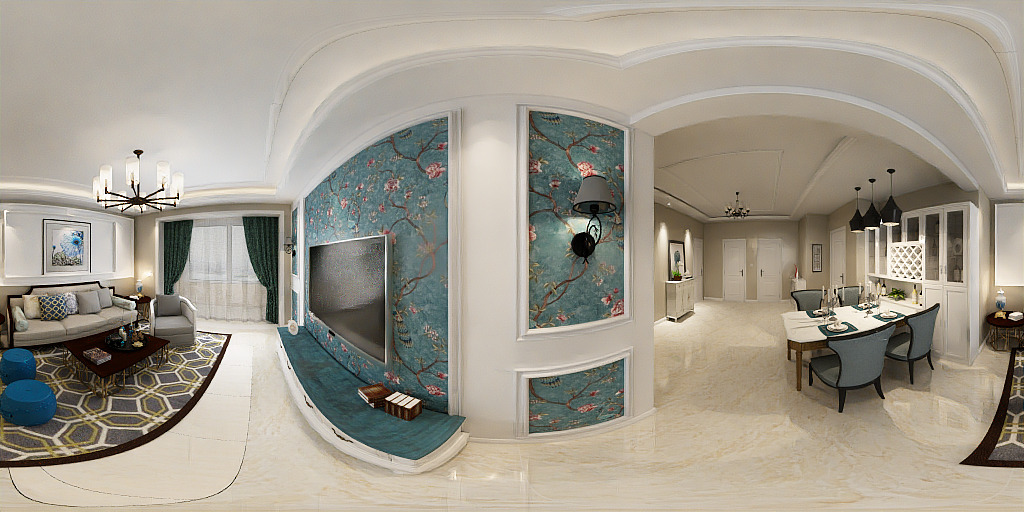}\hfill
\includegraphics[width=0.24\linewidth]{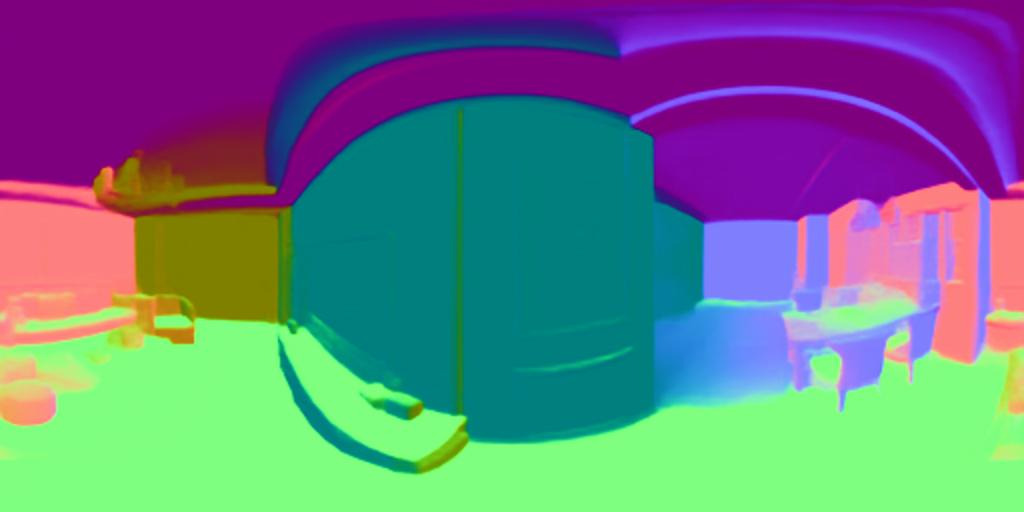}\hfill
\includegraphics[width=0.24\linewidth]{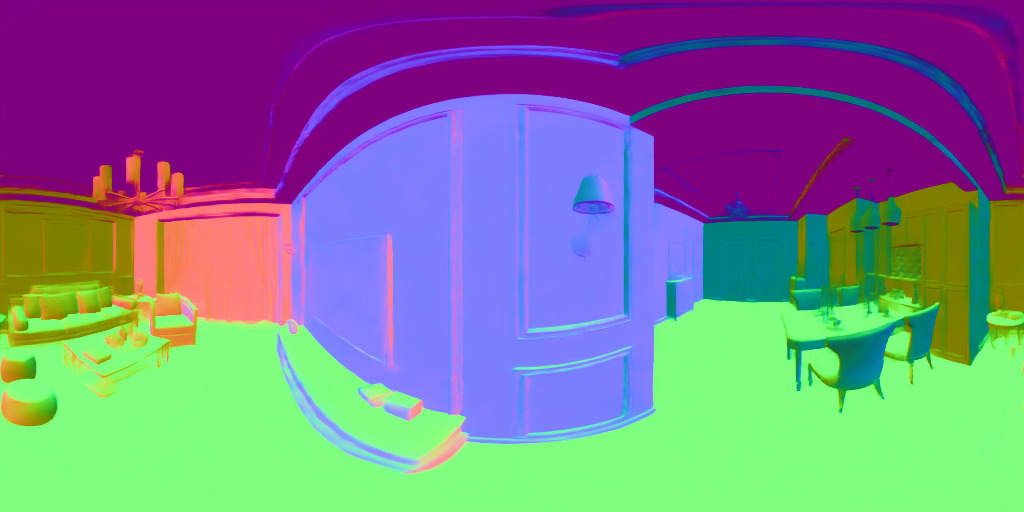}\hfill
\includegraphics[width=0.24\linewidth]{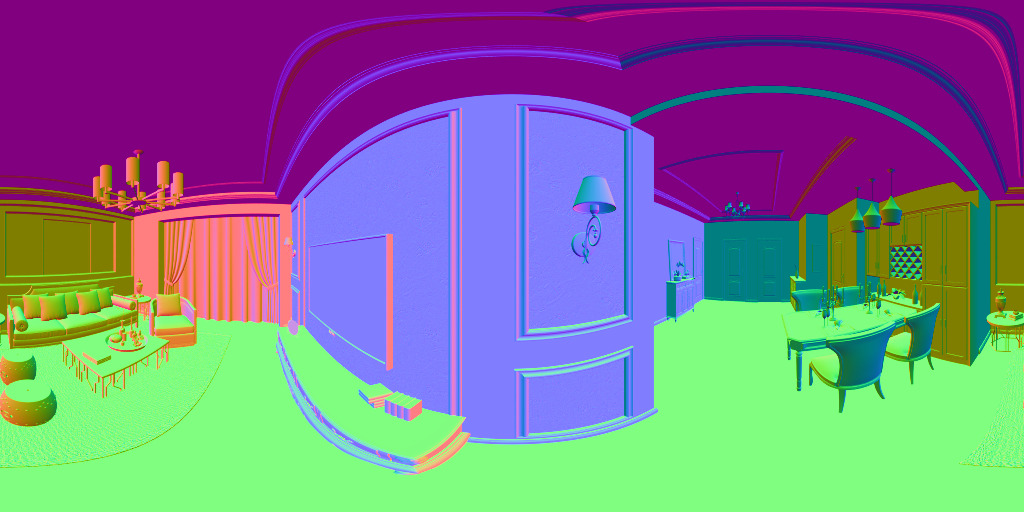}

\vspace{0.1em}

\includegraphics[width=0.24\linewidth]{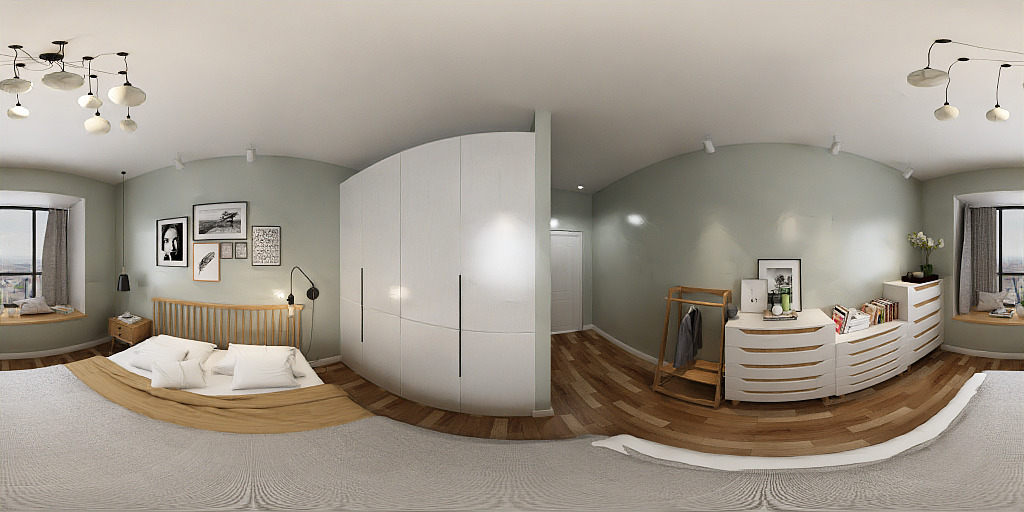}\hfill
\includegraphics[width=0.24\linewidth]{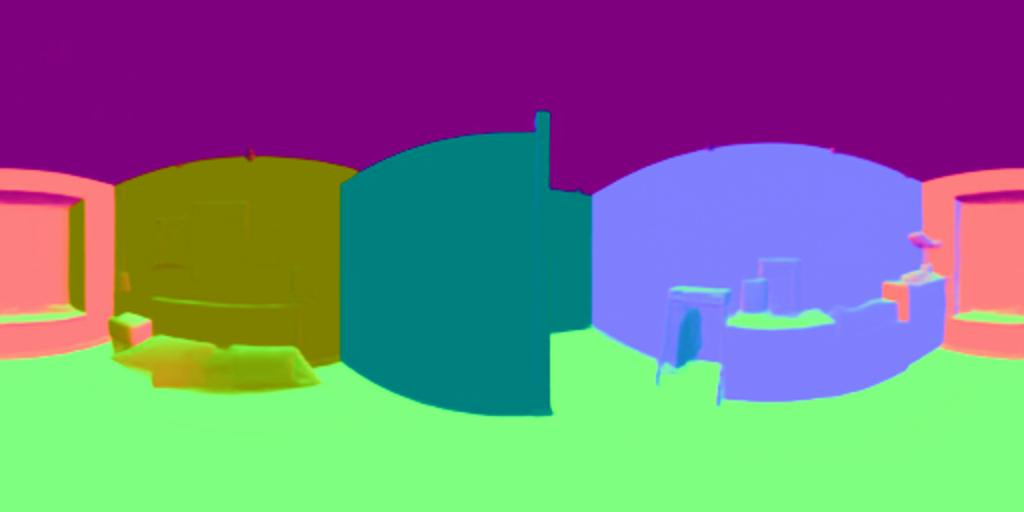}\hfill
\includegraphics[width=0.24\linewidth]{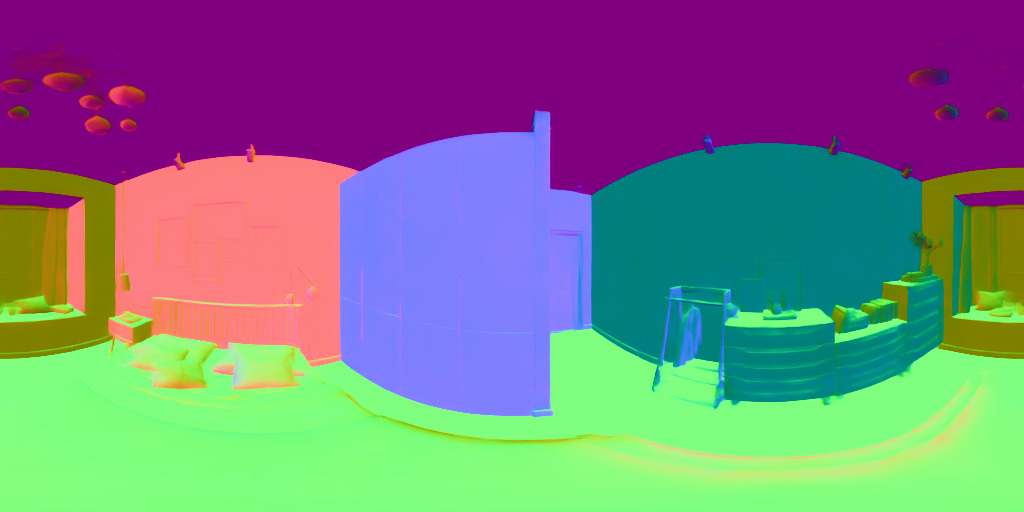}\hfill
\includegraphics[width=0.24\linewidth]{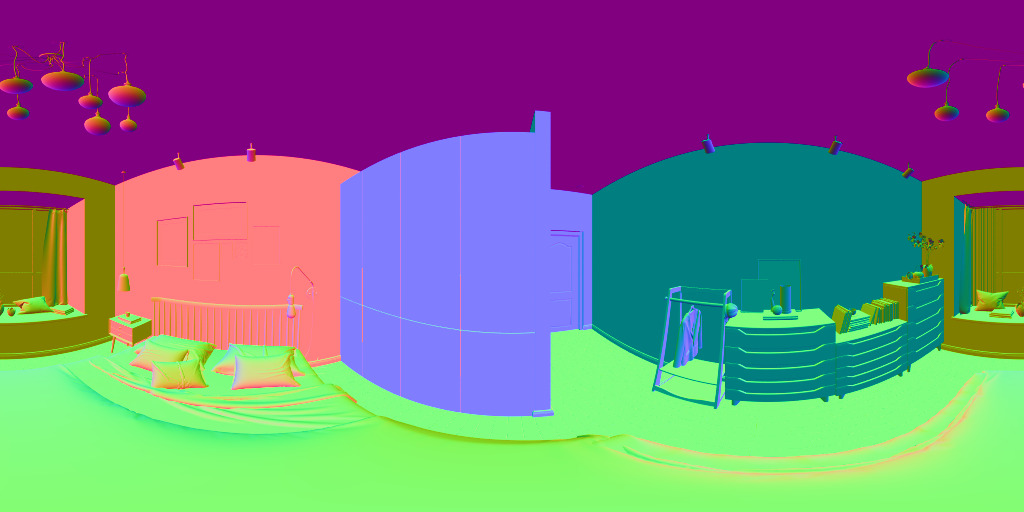}

\vspace{0.4em}

\captionof{figure}{\textbf{Qualitative comparison of panoramic surface normals estimation.} Results from PaGeR and MTL (strongest baseline), shown alongside
the RGB input and ground-truth depth on panoramas from the Structured3D dataset. (Best viewed
zoomed in.)}
\label{fig:normals}

\end{minipage}
\end{figure} 

To evaluate higher-order geometric consistency, we report surface normal estimation on the Structured3D dataset in Table~\ref{tab:normals}. \name{} sets a new state of the art, outperforming specialized architectures including PanoNormal and HyperSphere. Specifically, our framework achieves a Mean Angular Error of $5.49^\circ$ and an MSE of 174.9, which represents a major reduction from the 246.6 MSE of the previous state of the art. This validates our multi-task architecture choice, demonstrating that joint learning across separate heads improves the recovery of fine-grained surface structures.

\subsection{Qualitative Comparison}
Visual comparisons across both scale-invariant and metric depth settings are presented in \Cref{fig:comparison}. While baselines such as DAP and $\mathrm{DA}^2$  suffer from structural over-smoothing and fail to resolve fine details, \name{} delivers sharp geometric boundaries alongside a globally continuous scene layout. 

This structural precision is further highlighted in the point cloud reconstructions illustrated in \Cref{fig:pcl-comparison}. Here, competing approaches frequently introduce warped geometries and surface discontinuities along the boundaries. In contrast, our framework accurately recovers complex outdoor topographies while preserving intricate indoor structural arrangements. Furthermore, \Cref{fig:normals} demonstrates the high-resolution detail of our estimated surface normals; for more qualitative examples, see Appendix~\ref{sec:additional_qualitative_comparisons}.

\subsection{Ablation Studies}
\begin{table}[!t]
\begin{subtable}[t]{0.53\textwidth}
    \centering
    \small
    \setlength{\tabcolsep}{2pt}
    \renewcommand{\arraystretch}{0.95}
    \begin{tabular}{lcccccc} 
        \specialrule{0.75pt}{0pt}{1pt}
        & \multicolumn{3}{c}{\textbf{Replica360\_4K}} & \multicolumn{3}{c}{\textbf{Stanford2D3DS}} \\
        & \multicolumn{3}{c}{\textit{(Seam Eval.)}} & \multicolumn{3}{c}{\textit{(SI Depth Eval.)}} \\
        \cline{2-4} \cline{5-7}
        Method & SDD $\downarrow$ & SP $\downarrow$ & SS $\downarrow$ & AbsRel $\downarrow$ & RMSE $\downarrow$ & $\delta_1$ $\uparrow$ \\
        \hline
        DA3 & 33.71 & 44.29 & 68.97 & 6.58 & 35.79 & 95.94 \\
        \rowcolor{gray!15} \textbf{PaGeR} & \textbf{\phantom{0}0.94} & \textbf{\phantom{0}0.38} & \phantom{0}\textbf{2.31} & \textbf{5.93} & \textbf{35.34} & \textbf{96.10} \\
        \specialrule{0.75pt}{1pt}{1pt}
    \end{tabular}
    \vspace{-0.5em}
    \caption{Comparison against DA3 baseline.}
    \label{tab:vanilla_da3_vs_ours}
\end{subtable}%
\begin{subtable}[t]{0.47\textwidth}
    \centering
    \small
    \setlength{\tabcolsep}{2pt}
    \renewcommand{\arraystretch}{0.95}
    \begin{tabular}{lcccc}
        \specialrule{0.75pt}{0pt}{1pt}
        Method & Mean $\downarrow$ &  MSE $\downarrow$ & $\delta_{5\degree}$ $\uparrow$  & $\delta_{22.5\degree}$ $\uparrow$  \\
        \hline
        Frozen backbone & 6.21 & 251.1 & 77.00 & 90.52 \\
        W/o perceptual & 5.92 & 198.2 & 77.82 & 90.96 \\
        Rand. init. DPT & 6.02 & 191.9 & 78.02 & 89.76 \\
        \rowcolor{gray!15} \textbf{Full PaGeR} & \textbf{5.49} & \textbf{174.9} & \textbf{79.91} & \textbf{92.83} \\
        \specialrule{0.75pt}{1pt}{1pt}
    \end{tabular}
    \vspace{-0.5em}
    \caption{Surface Normals on Structured3D.}
    \label{tab:normals_ablation}
\end{subtable}
\vspace{-0.5em}
\caption{\textbf{Ablations for vanilla DA3 and surface normals}.}
\vspace{-2em}
\end{table}

\noindent\textbf{Comparison to Baseline.} 
We evaluate \name{} against the vanilla DA3 baseline using depth accuracy and cross-face geometric consistency metrics (\Cref{tab:vanilla_da3_vs_ours}; formal definitions in \Cref{sec:seam_metrics}). \name{} significantly improves upon both. While preserving the robust geometric priors of the base model, our framework resolves depth misalignment across face boundaries to yield seamless reconstructions, as qualitatively illustrated in \Cref{fig:da3-comparison}. Apart from the standard 6-face cubemap input, we ablate alternative input configurations to the DA3 backbone, provided in the supplementary material~\ref{sec:alternative_DA3}.

\begin{table}[!h]
\vspace{-0.2em}
\begin{subtable}[t]{0.5\textwidth}
    \centering
    \small
    \setlength{\tabcolsep}{2pt}
    \renewcommand{\arraystretch}{0.95}
    \begin{tabular}{lccc}
        \specialrule{0.75pt}{0pt}{1pt}
        Method & AbsRel $\downarrow$ & RMSE $\downarrow$ & $\delta_1$ $\uparrow$  \\
        \hline
        W/o camera cond. & 10.12 & 43.91 & 90.01 \\ 
        Linear-depth & \phantom{0}6.96 & 39.12 & 94.98 \\
        W/o sky segm. head & \phantom{0}6.88 & 40.74 & 93.59 \\ 
        Euclidean-depth & \phantom{0}6.34 & 36.91 & 95.48 \\
        W/o valid padding & \phantom{0}6.45 & 37.00 & 95.73 \\
        W/o structural losses & \phantom{0}6.11 & 36.72 & 95.93 \\
        Vanilla L1 loss & \phantom{0}6.15 & 36.02 & 96.00 \\
        W/o perspective data & \phantom{0}6.45 & 37.98 & 95.45 \\
        \rowcolor{gray!15} \textbf{PaGeR} & \phantom{0}\textbf{5.93} & \textbf{35.34} & \textbf{96.10} \\
        \specialrule{0.75pt}{1pt}{1pt}
    \end{tabular}%
    \vspace{-0.4em}
    \caption{Scale-invariant depth.}
    \label{tab:SI_depth_ablation}
\end{subtable}
\begin{subtable}[t]{0.5\textwidth}
    \centering
    \small
    \setlength{\tabcolsep}{2pt}
    \renewcommand{\arraystretch}{0.95}
    \begin{tabular}{lccc}
        \specialrule{0.75pt}{0pt}{1pt}
        Method & AbsRel $\downarrow$ & RMSE $\downarrow$ & $\delta_1$ $\uparrow$ \\
        \hline
        Single Global Scale & 12.67 & 47.14 & 88.05 \\   
        Dense, F=1 & 11.35 & 46.31 & 88.35 \\  
        Dense, F=2 & 11.34 & 45.98 & 90.01  \\
         \rowcolor{gray!15} \textbf{Dense, F=4 (Ours)} & \textbf{10.94} & \textbf{45.43} & \textbf{90.94} \\
        Dense, F=8 & 12.15 & 47.09 & 89.92 \\
        \specialrule{0.75pt}{1pt}{1pt}
    \end{tabular}
    \vspace{-0.5em}
    \caption{Metric depth.}
    \label{tab:metric_ablation}
\end{subtable}
\vspace{-0.4em}
\caption{\textbf{Ablations for scale-invariant and metric depth on Stanford2D3DS}.}
\vspace{-0.2em}
\end{table}
\vspace{-1em}
\noindent\textbf{Scale-Invariant Depth.} 
\Cref{tab:SI_depth_ablation} isolates the impact of individual architectural choices on the Stanford2D3DS dataset. Omitting explicit camera conditioning causes the most severe performance drop, increasing absolute relative error. Our log-space formulation also outperforms alternative linear-depth targets. Furthermore, decoupling infinite depth is critical; removing the auxiliary sky segmentation head causes a major spike in RMSE. Finally, simplifying the training objective or omitting structural data refinements, such as cross-face valid padding and joint perspective training, consistently degrades overall precision.

\noindent\textbf{Surface Normals.}
We validate our surface normal setup on the Structured3D dataset (\Cref{tab:normals_ablation}). Altering the encoder configuration proves highly detrimental; freezing the backbone causes a major spike in MSE, while optimizing a randomly initialized decoder yields the poorest threshold accuracy. Omitting the perceptual loss from the training objective causes a clear regression across all metrics, confirming that feature matching regularizers are necessary to prevent over-smoothing and recover sharp geometric boundaries.

\noindent\textbf{Metric Scale.} 
\Cref{tab:metric_ablation} evaluates our metric supervision strategy on the Stanford2D3DS dataset. Relying strictly on global scalar estimation rather than dense supervision across distributed anchor points yields the poorest baseline performance. Within our dense supervision framework, tuning the feature downsampling factor $F$ is critical; both insufficient ($F=1$) and extreme ($F=8$) downsampling undermine accuracy. Our final configuration successfully balances these factors to achieve optimal metric depth reconstruction.

\noindent\textbf{Architectural vs.\ Training Adaptations}

Our architectural adaptation (Spherically-Aware Decoder Padding, SADP) and mixed panoramic co-training target distinct failure modes. SADP is purely local: it replaces zero padding with features from geometrically adjacent faces, ensuring they agree exactly \textit{at} the boundary, but cannot force independent predictions to cohere as a unified scene. Conversely, panoramic co-training addresses global consistency, teaching the network to align the six faces into a single spatial layout with a unified metric scale.

Table~\ref{tab:adaptation_ablation} isolates these contributions. SADP alone nearly halves seam defects (SDD) but barely improves depth accuracy (AbsRel). Panoramic co-training alone improves accuracy slightly, but seam artifacts persist. Only their combination yields optimal results. Notably, the absolute error improvements appear small because the unadapted perspective foundation model already provides a highly efficient baseline (6.58 AbsRel)---explaining why our approach achieves state-of-the-art performance with only $\sim$100k panoramas, while native equirectangular training requires orders of magnitude more data.

\begin{table}[h]
\centering
\small
\renewcommand{\arraystretch}{1.1}
\setlength{\tabcolsep}{8pt}
\begin{tabular}{lcc}
    \specialrule{0.75pt}{0pt}{1pt}
    & \textbf{AbsRel $\downarrow$} & \textbf{Seam Defects (SDD) $\downarrow$} \\
    \textbf{Variant} & (Stanford2D3DS) & (Replica360-4K) \\
    \hline
    Vanilla DA3 on cubemaps (No adaptation)  & 6.58 & 33.71 \\
    SADP only, no panoramic training         & 6.49 & 19.01 \\
    Panoramic training only, without SADP    & 6.45 & 12.67 \\
    \rowcolor{gray!15} \textbf{Panoramic training + SADP (Ours)} & \textbf{5.93} & \textbf{\phantom{0}0.94} \\
    \specialrule{0.75pt}{1pt}{1pt}
\end{tabular}
\caption{\textbf{Ablation of local structural adaptation (SADP) versus global panoramic co-training.} Metrics are scaled by $\times 100$. Both adaptations are necessary to achieve optimal global accuracy and eliminate local seam artifacts.}
\label{tab:adaptation_ablation}
\end{table}

\section{Related Work}

\noindent\textbf{Perspective Geometry Estimation.}
Monocular depth estimation has evolved into general-purpose architectures capable of robust zero-shot 3D geometry inference~\cite{ranftl2020towards, piccinelli2025unidepthv2, yang2024depth}, recently augmented by diffusion priors~\cite{ke2024repurposing, fu2024geowizard, garcia2025fine} and unified geometric models~\cite{hu2024metric3d, wang2025moge} for joint depth and normal prediction. Despite high structural fidelity, recovering absolute metric scale remains challenging. Naively regressing dense metric depth or localized scales~\cite{bhat2023zoedepth, wei2023fsdepthfocalandscaledepthestimation, yin2023metric3d} disrupts perspective priors, while directly regressing a single global scale factor~\cite{Zhu_2026} suffers from sparse gradients. However, for panoramic imagery with fixed 360° intrinsics, metric reconstruction mathematically reduces to estimating a single global scale. Exploiting this, we propose a framework that transfers rich perspective priors to panoramas, outputting scale-invariant geometry alongside a robustly decoupled global metric scale.

\noindent\textbf{Visual Geometry Foundation Models.}
Visual Geometry Foundation Models (VGFMs)~\cite{wang2024dust3rgeometric3dvision, leroy2024groundingimagematching3d, wang2025vggtvisualgeometrygrounded, lin2025depth3recoveringvisual,dens3r} have shifted 3D reconstruction from task-specific estimators to unified, feed-forward architectures. By treating reconstruction as dense correspondence regression, they bypass traditional iterative Structure-from-Motion. A key advantage of VGFMs is their input flexibility, seamlessly transitioning from monocular to multi-view regimes and natively incorporating varying camera intrinsics or extrinsics. We build directly upon these robust, flexible priors, adapting their multi-view feed-forward capabilities to the panoramic domain.

\noindent\textbf{Panoramic Geometry Estimation.}
Panoramic geometry estimation historically struggles with equirectangular distortions and data scarcity. Previous methods range from designing specialized architectures~\cite{wang2022bifuse++, shen2022panoformer, huang2024panonormal, lee2025hush} to adapting perspective models via multi-projection formats~\cite{wang2024depth, ai2024elite360d}. Recently, training-free cubemap adaptations~\cite{kalischek2025cubediff, huang2025dreamcube, jung2025rpg360robust360depth} mitigate distortions by predicting mutually consistent faces, though they often struggle to merge views back into a seamless equirectangular projection (ERP). Consequently, the field faces a trade-off: training-free multi-face methods exhibit seam artifacts, while fully-supervised continuous ERP models~\cite{li20252, lin2025depthpanoramasfoundationmodel} incur prohibitive training costs. Even concurrent efforts adapting VGFMs to 360° panoramas~\cite{guo2026panovggtfeedforward3dreconstruction, yuan2026vggt360geometryconsistentzeroshotpanoramic} require extensive retraining. In contrast, our approach retains and reuses the original perspective VGFM priors, extending them to panoramas without requiring massive retraining or sacrificing expressivity. We additionally provide extensive analysis and the comparison of our method to similar and relevant prior works in the supplementary~\ref{sec:additional_related_work}.
\section{Conclusion}
\label{sec:conclusion}

We have presented \name{}, a unified framework for panoramic geometry estimation that successfully lifts the robust representations of perspective depth models into the spherical domain. By combining a synchronized cubemap representation with separate decoding heads, our architecture transfers established perspective priors to multiple 360° geometric tasks without requiring extensive retraining. To guide this adaptation, we introduced a joint panoramic/perspective training regime. Extensive evaluations demonstrate that \name{} establishes a new state of the art in zero-shot panoramic reconstruction across diverse indoor and outdoor environments. Although our instantiation leverages DA3, the architectural principles of our framework maintain compatibility with alternative geometric transformers~\cite{wang2025vggtvisualgeometrygrounded, wang2026pi3permutationequivariantvisualgeometry}, offering a generalized methodology for bridging the domain gap between perspective and panoramic 3D scene understanding.

\section{Limitations}
\label{sec:limitations}
While the unbounded $360^{\circ}$ context of panoramas reduces global structural ambiguities---such as truncated objects or unconstrained floor layouts---it cannot resolve local pixel ambiguities inherent to monocular depth prediction. First, \name{} struggles with specular, reflective, or transparent surfaces, typically reconstructing the geometry reflected or seen through them (e.g., interpreting a mirror as an adjoining room). Resolving this likely requires a fundamentally different algorithmic approach. Second, accuracy naturally degrades in large textureless regions or under out-of-distribution conditions (e.g., complex outdoor lighting), where the model must rely on generalized layout priors.
Furthermore, our outdoor evaluation is constrained by the scarcity of comprehensive, real-world panoramic benchmarks, relying primarily on a limited set of 100 European urban panoramas (ZüriPano) alongside synthetic data (Holo360D). Consequently, we are currently unable to fully validate the model's zero-shot generalization across highly diverse, natural, or non-urban outdoor landscapes, representing a shared limitation in both our evaluation and the broader field.

Finally, despite cross-face padding and global attention mechanisms, subtle geometric misalignments can occasionally emerge at face boundaries in structurally complex scenes. This is primarily a data limitation, as the network's attention must cover a long tail of complex edge configurations. While failures in textureless regions, out-of-distribution appearances, and seam artifacts are expected to improve with dataset scaling, guaranteeing absolute boundary consistency may ultimately require explicit mathematical continuity constraints in future work.

\newpage
{
\section*{Acknowledgments}

This work was supported by ETHAR (funded by a Google donation) and the Swiss National Supercomputing Center (Swiss AI Initiative, projects a144, a0175, a0182). We thank Hothifa Smair and Parametra for authorizing the use of the iCity Blender add-on to generate and release the \emph{PanoInfinigen} urban split under an academic license. We also thank Veljko Bozic for assembling the urban scenes.

    \small
    \bibliographystyle{ieeenat_fullname}
    \bibliography{main}
}

\newpage

\appendix

\section{PanoInfinigen}
\label{sec:panoinfinigen}

High-quality datasets are the cornerstone of robust monocular depth estimation. While perspective depth estimation has benefited immensely from large-scale, diverse data collections, the panoramic domain suffers from limited training data. Existing standard benchmarks, such as Stanford2D3DS~\cite{https://doi.org/10.57761/gmhc-wx10} and Matterport3D360~\cite{bata1126}, rely on real-world scanners. Consequently, they are restricted to static indoor environments and often contain acquisition artifacts, such as missing depth values in reflective or distant regions. Synthetic alternatives like Structured3D~\cite{Structured3D} offer dense, complete ground truth but are similarly confined to low resolutions and indoor scenes with limited variability.
 To address this gap, we introduce \textit{PanoInfinigen}, a large-scale synthetic dataset designed to unlock the potential of high-resolution, general-purpose panoramic depth estimation.

\paragraph{Data Generation and Characteristics.}

PanoInfinigen provides complete, high-quality ground truth for both depth and surface normals across a variety of indoor and outdoor scenes. The dataset is intended to support the development of high-performance models; moreover, it is extensible through our accompanying open-source generation tool.
To construct PanoInfinigen, we build upon Infinigen~\cite{infinigen2023infinite, infinigen2024indoors}, a procedural content generation framework capable of synthesizing realistic, unbounded 3D environments. In contrast to methods that depend on fixed 3D asset libraries, Infinigen procedurally generates both geometry and textures, thereby enabling effectively unlimited variability in scene layout, object placement, and illumination conditions. We extend its rendering pipeline to support 360° equirectangular projection and generate 70,000 unique panoramas from 20,000 distinct scenes.

To extend coverage to urban environments, we employ the iCity~\cite{iCity} procedural city-generation add-on for Blender, which automates the creation of realistic and diverse synthetic urban scenes. We synthesize 20 urban environments with varying configurations and visual styles, and render approximately 400 panoramas per city using our existing Infinigen-based auto-rendering tool. This results in roughly 7,000 high-quality outdoor panoramas, each accompanied by dense depth and surface normal maps.

\begin{figure}
    \centering
    \includegraphics[width=0.332\linewidth]{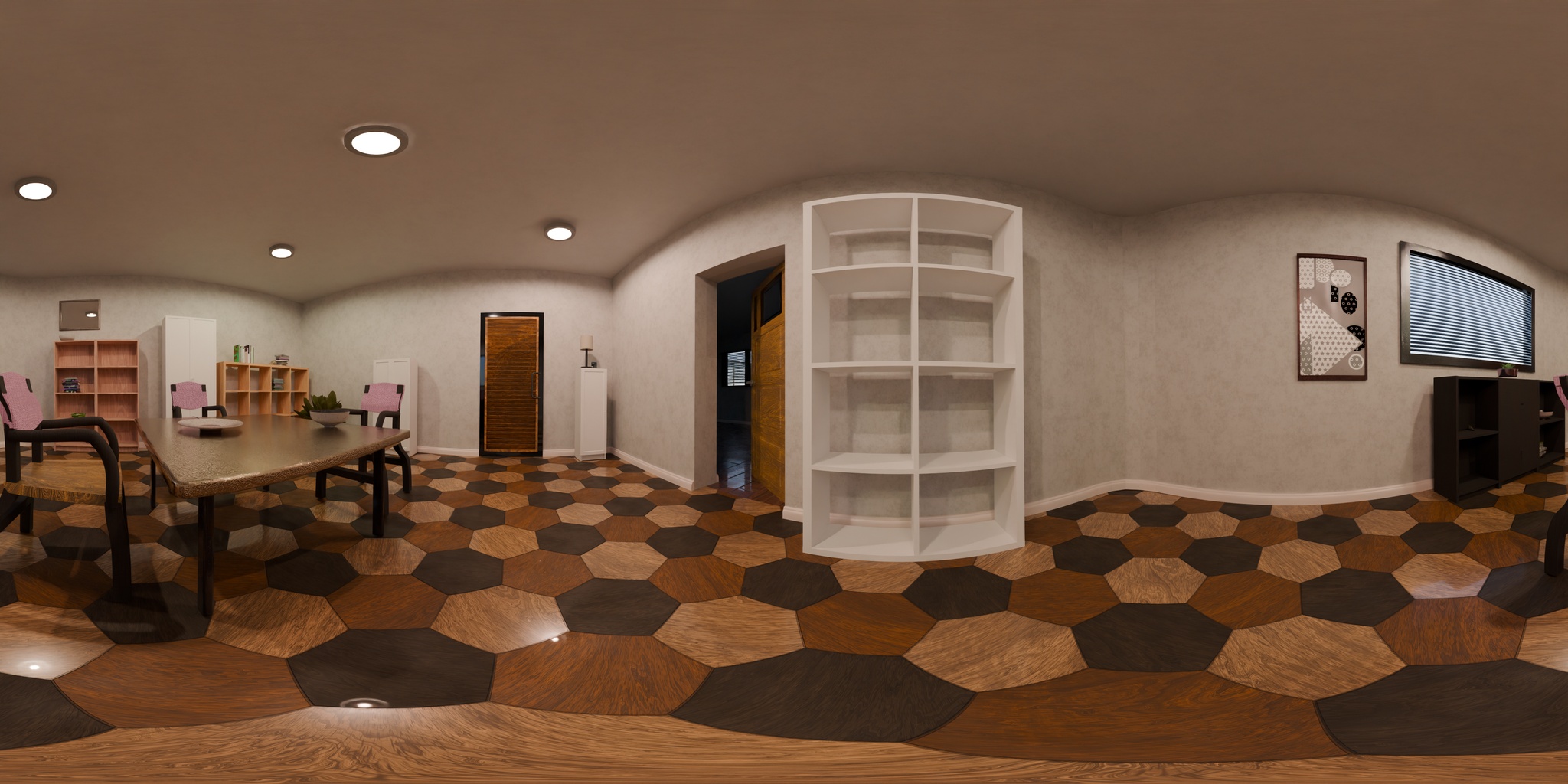}%
    \hfill
    \includegraphics[width=0.332\linewidth]{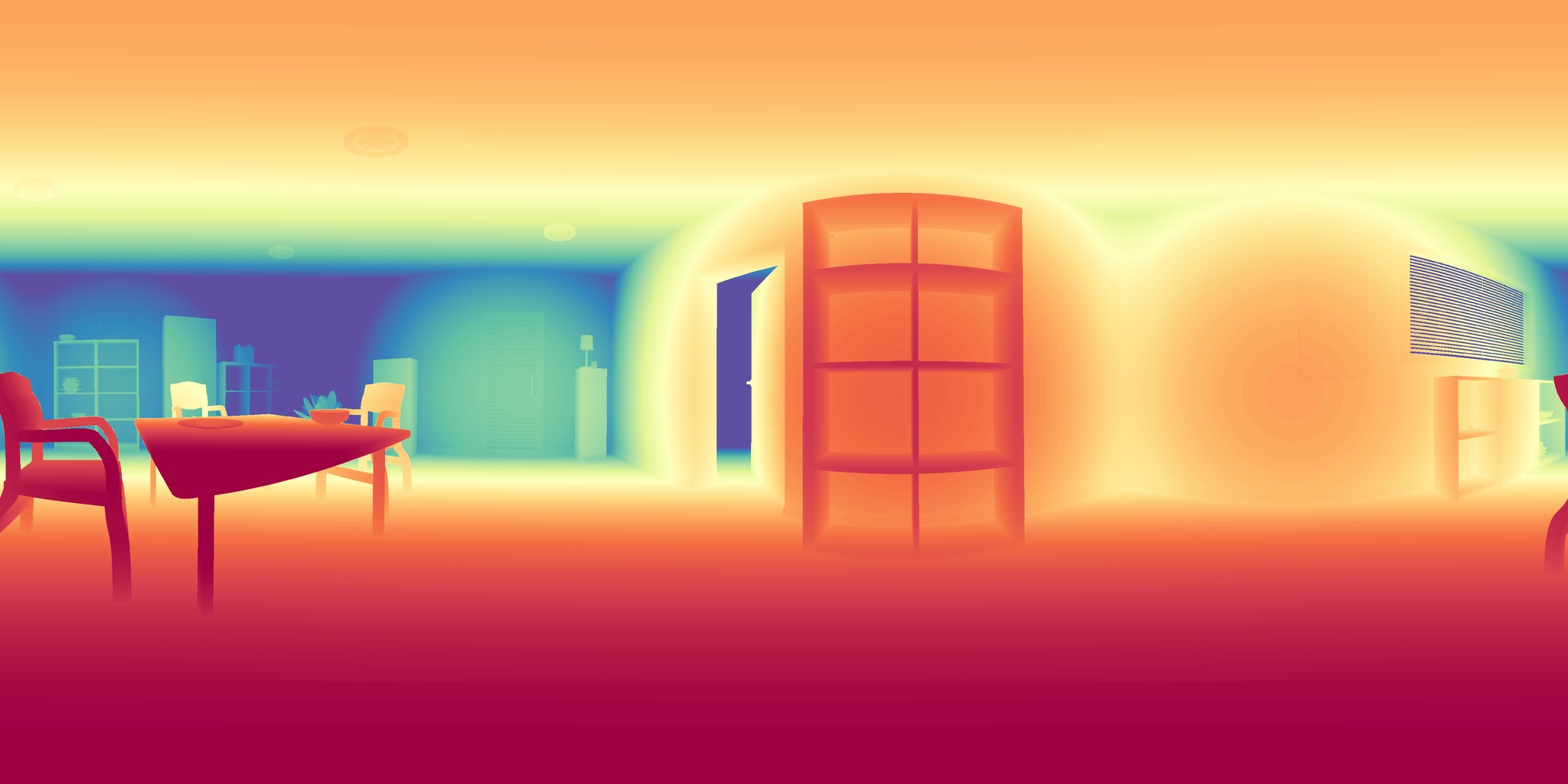}%
    \hfill
    \includegraphics[width=0.332\linewidth]{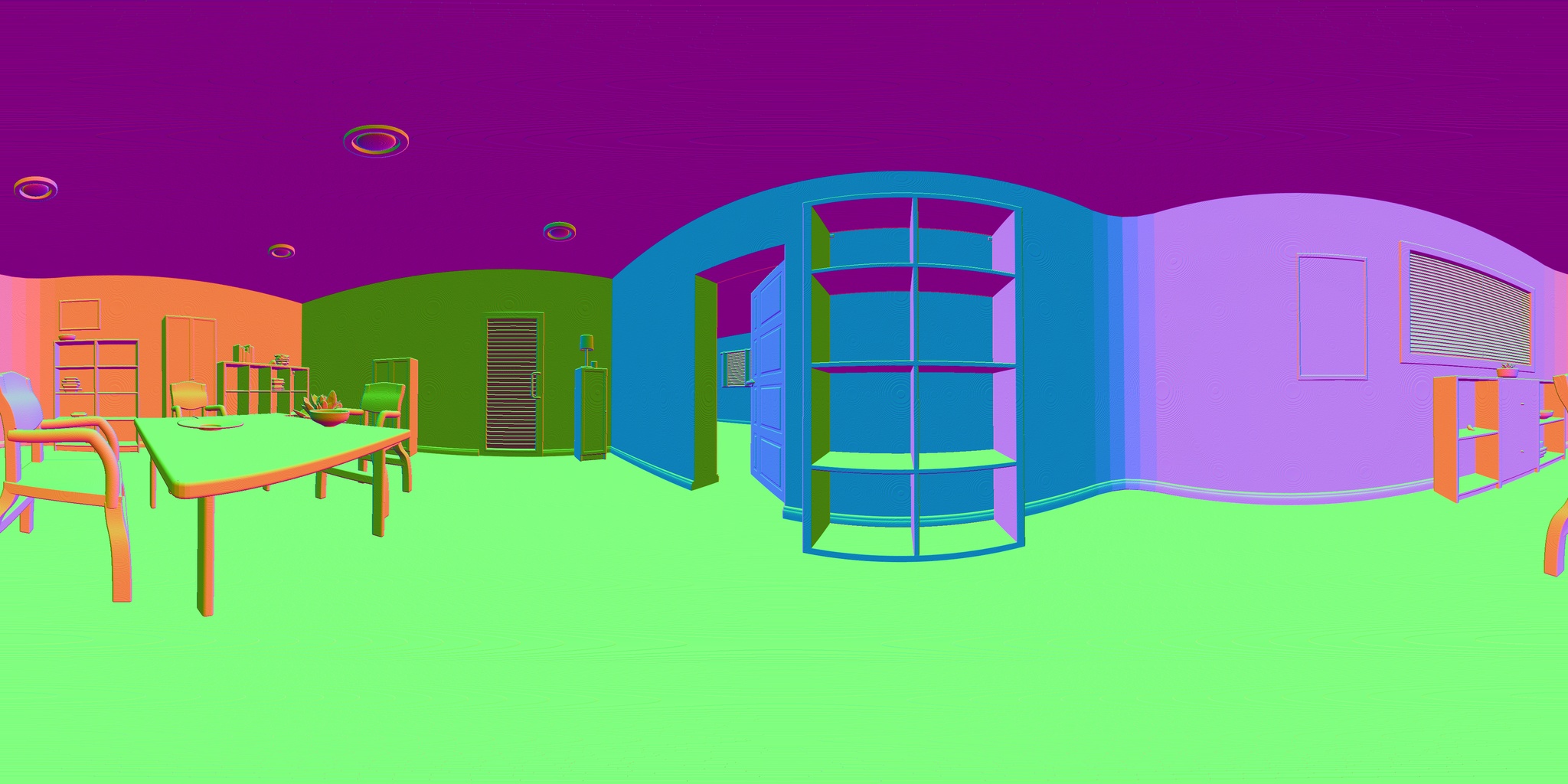}%
    \hfill
    \includegraphics[width=0.332\linewidth]{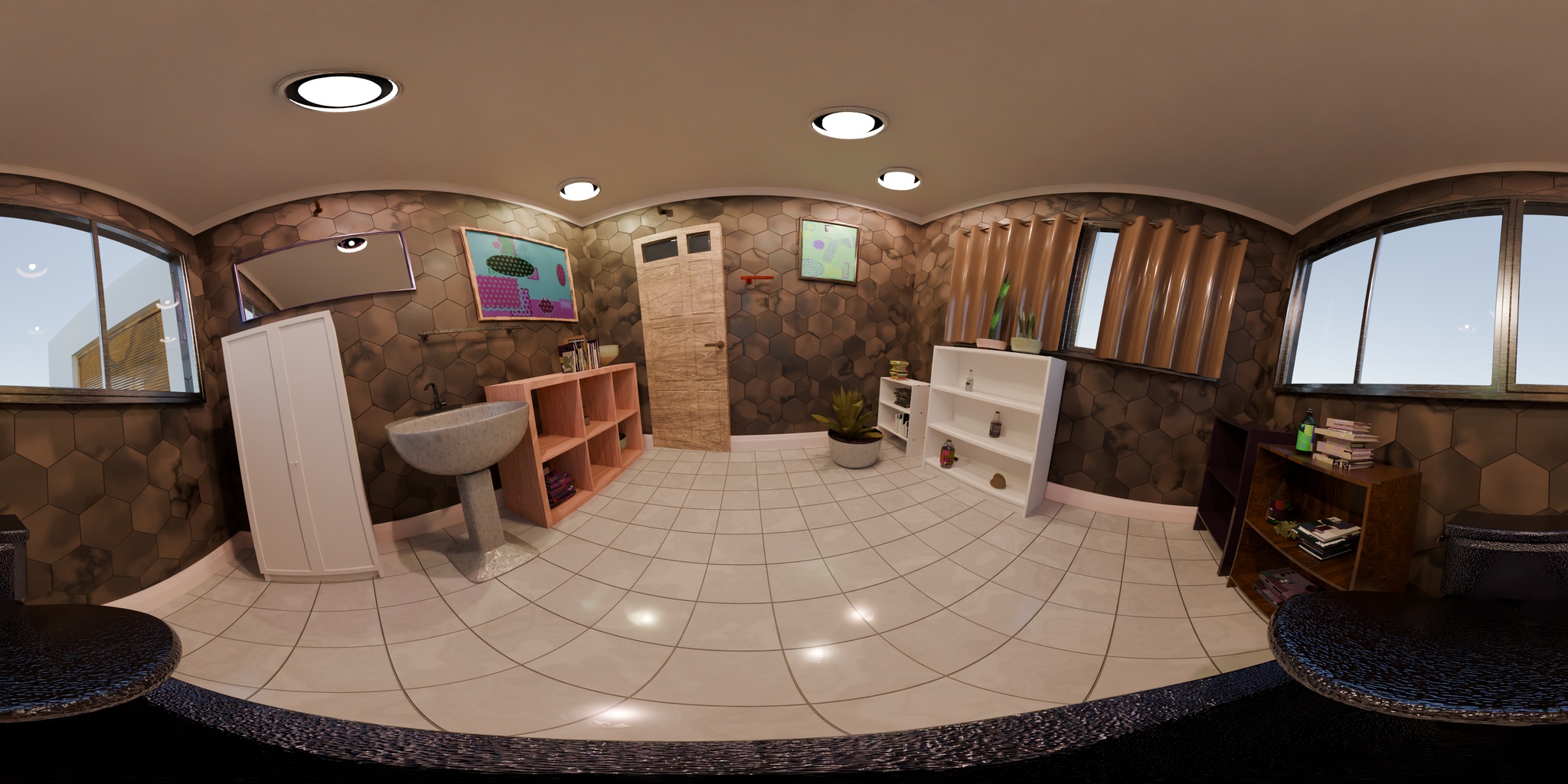}%
    \hfill
    \includegraphics[width=0.332\linewidth]{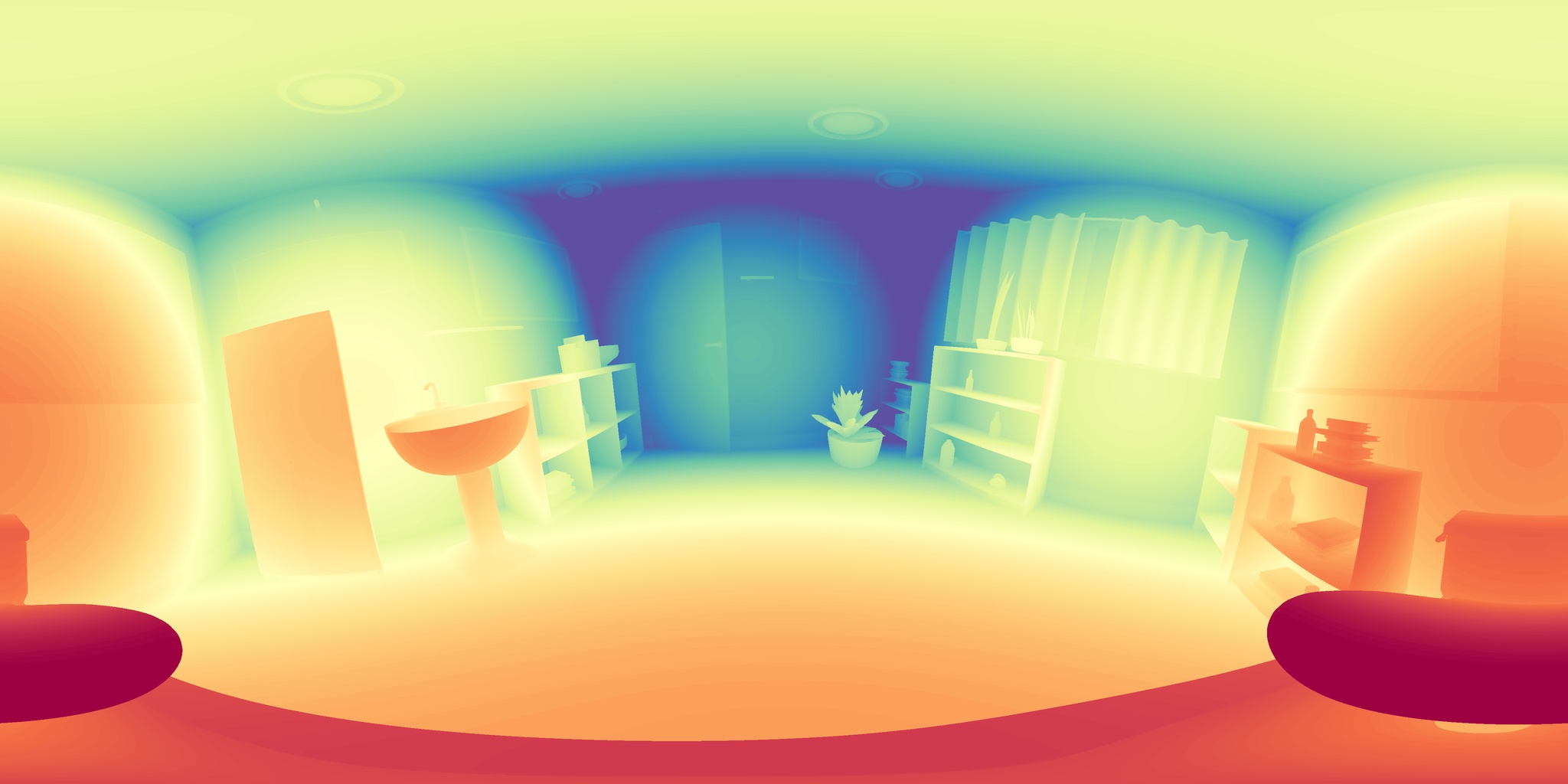}%
    \hfill
    \includegraphics[width=0.332\linewidth]{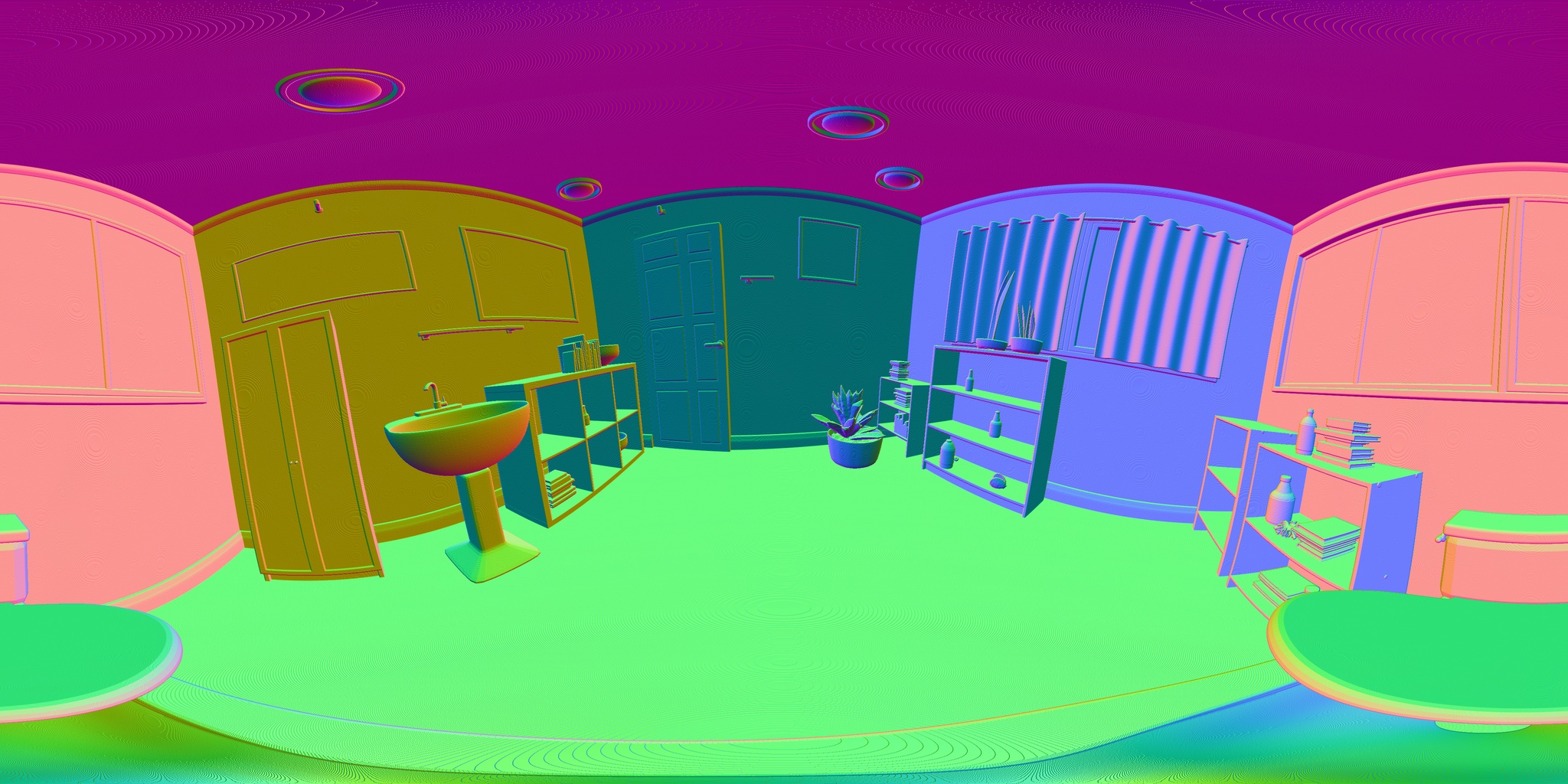}%
    \hfill
    \includegraphics[width=0.332\linewidth]{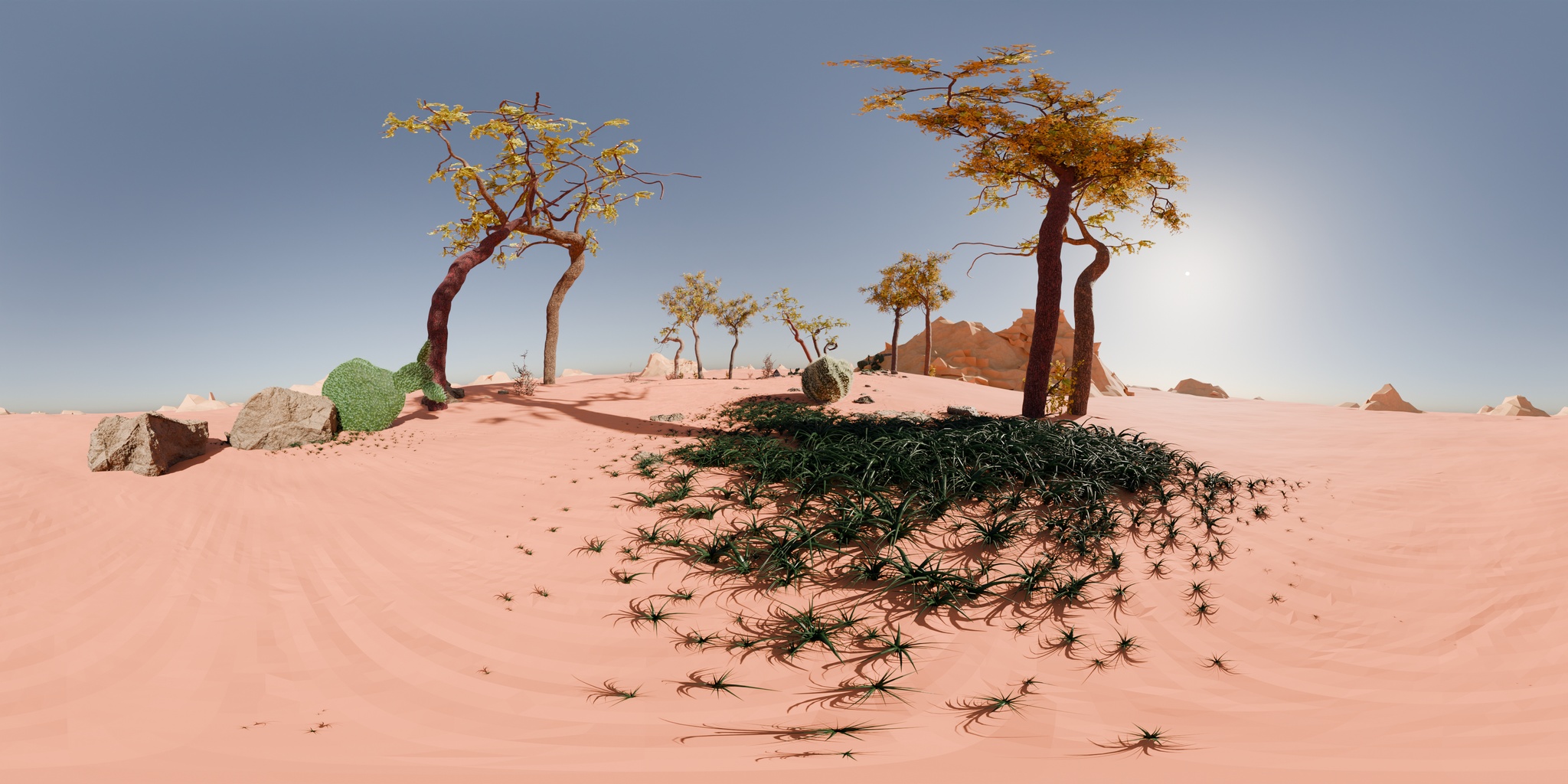}%
    \hfill
    \includegraphics[width=0.332\linewidth]{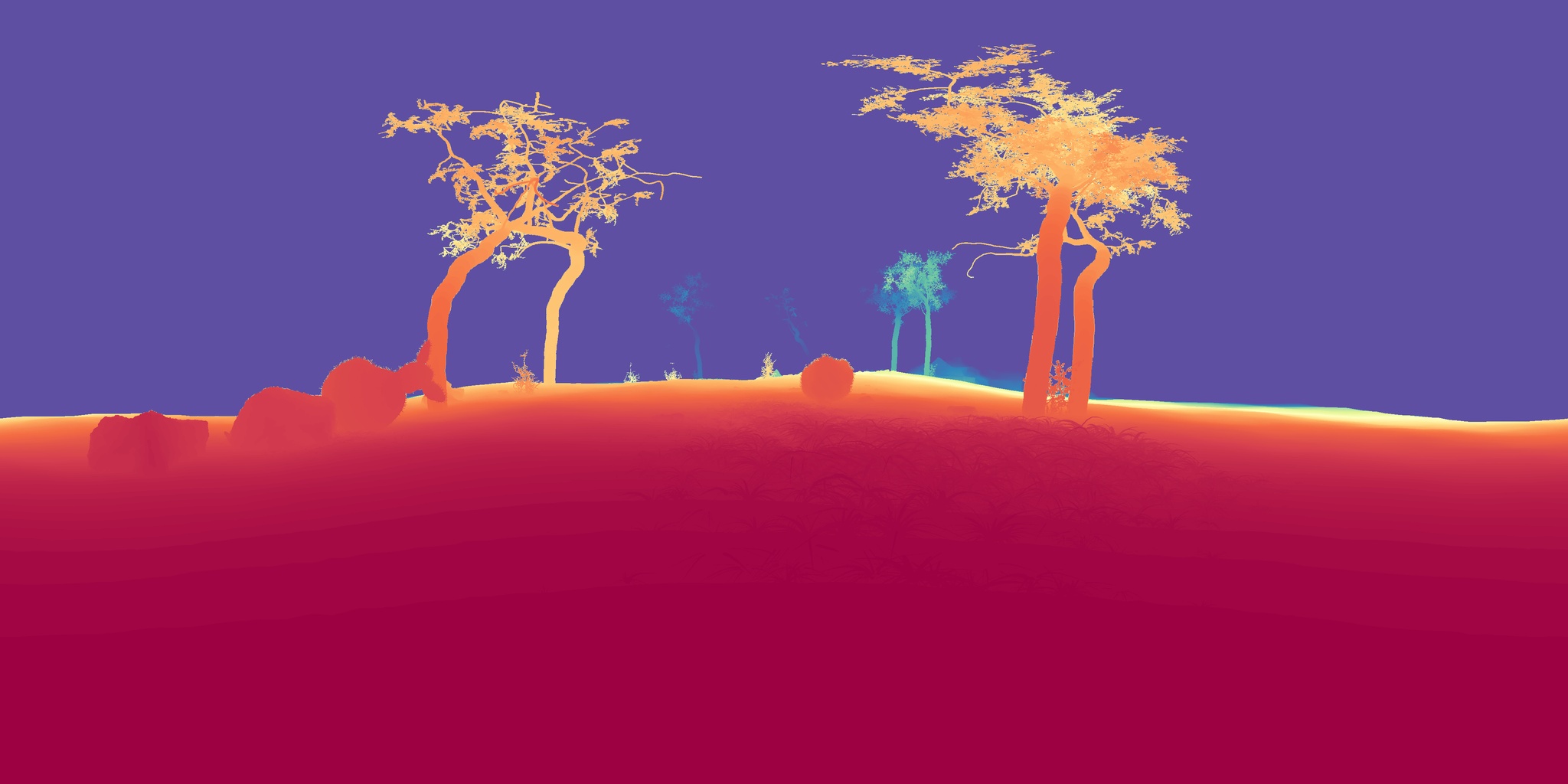}%
    \hfill
    \includegraphics[width=0.332\linewidth]{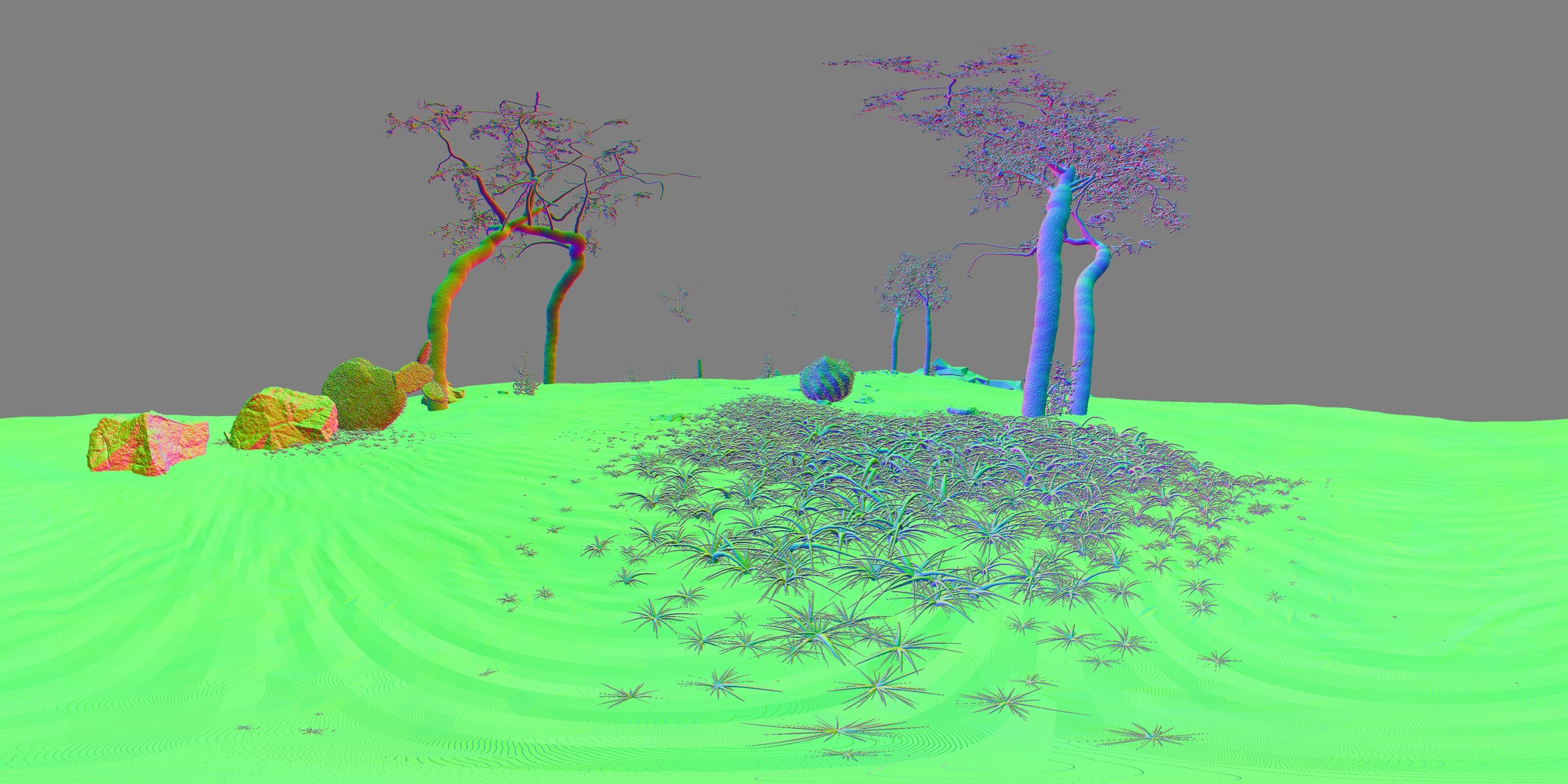}%
    \hfill
    \includegraphics[width=0.332\linewidth]{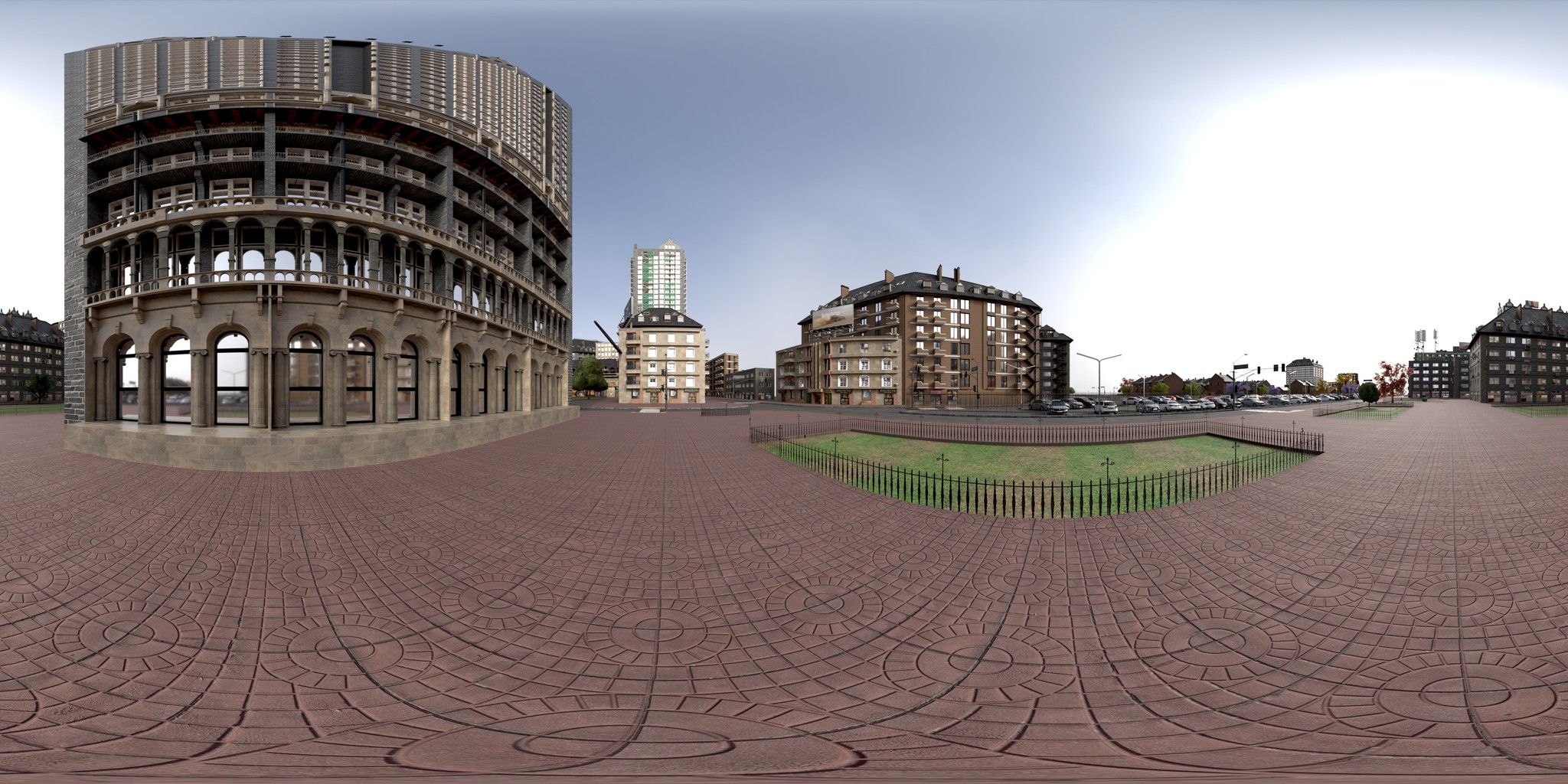}%
    \hfill
    \includegraphics[width=0.332\linewidth]{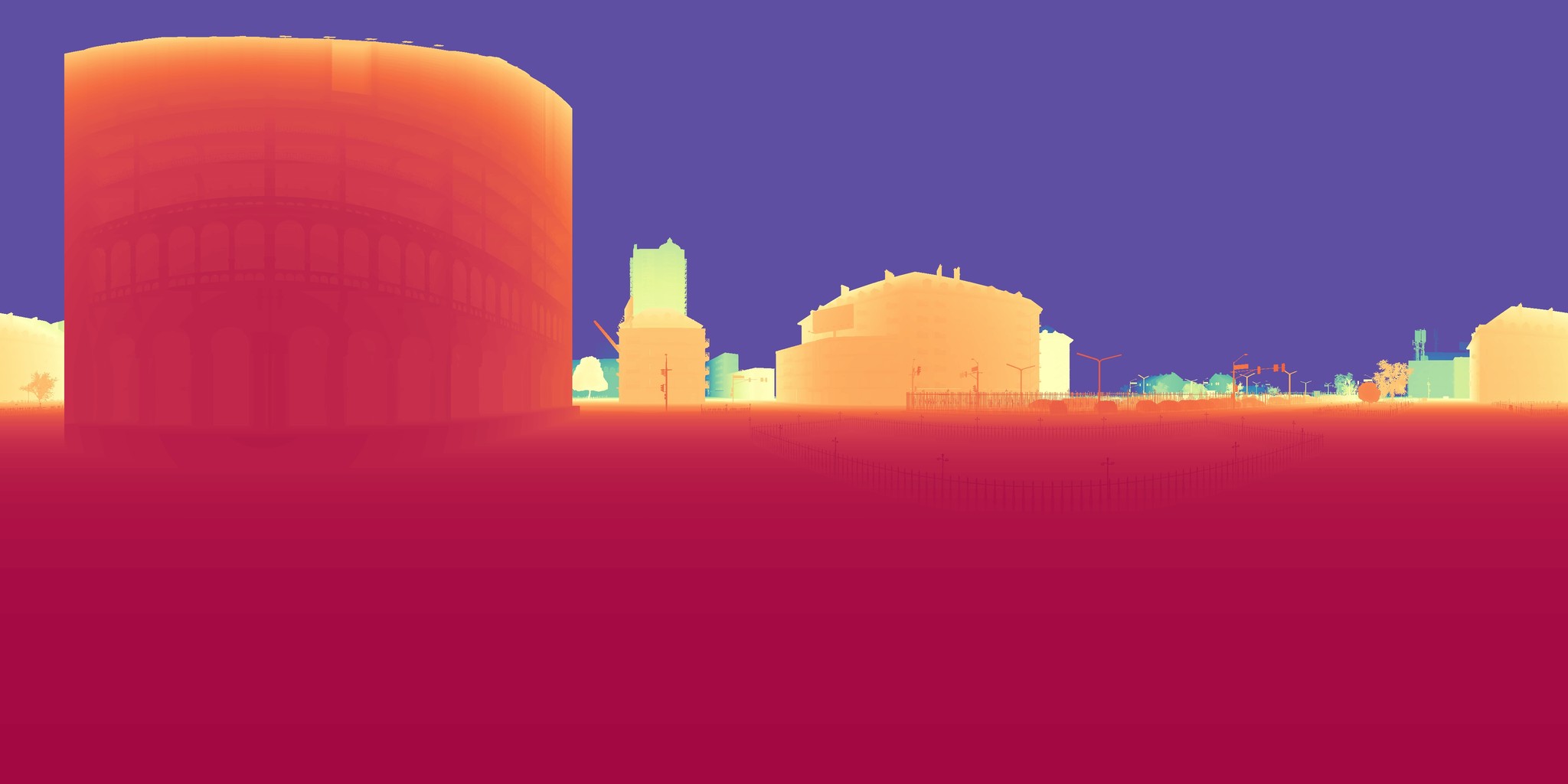}%
    \hfill
    \includegraphics[width=0.332\linewidth]{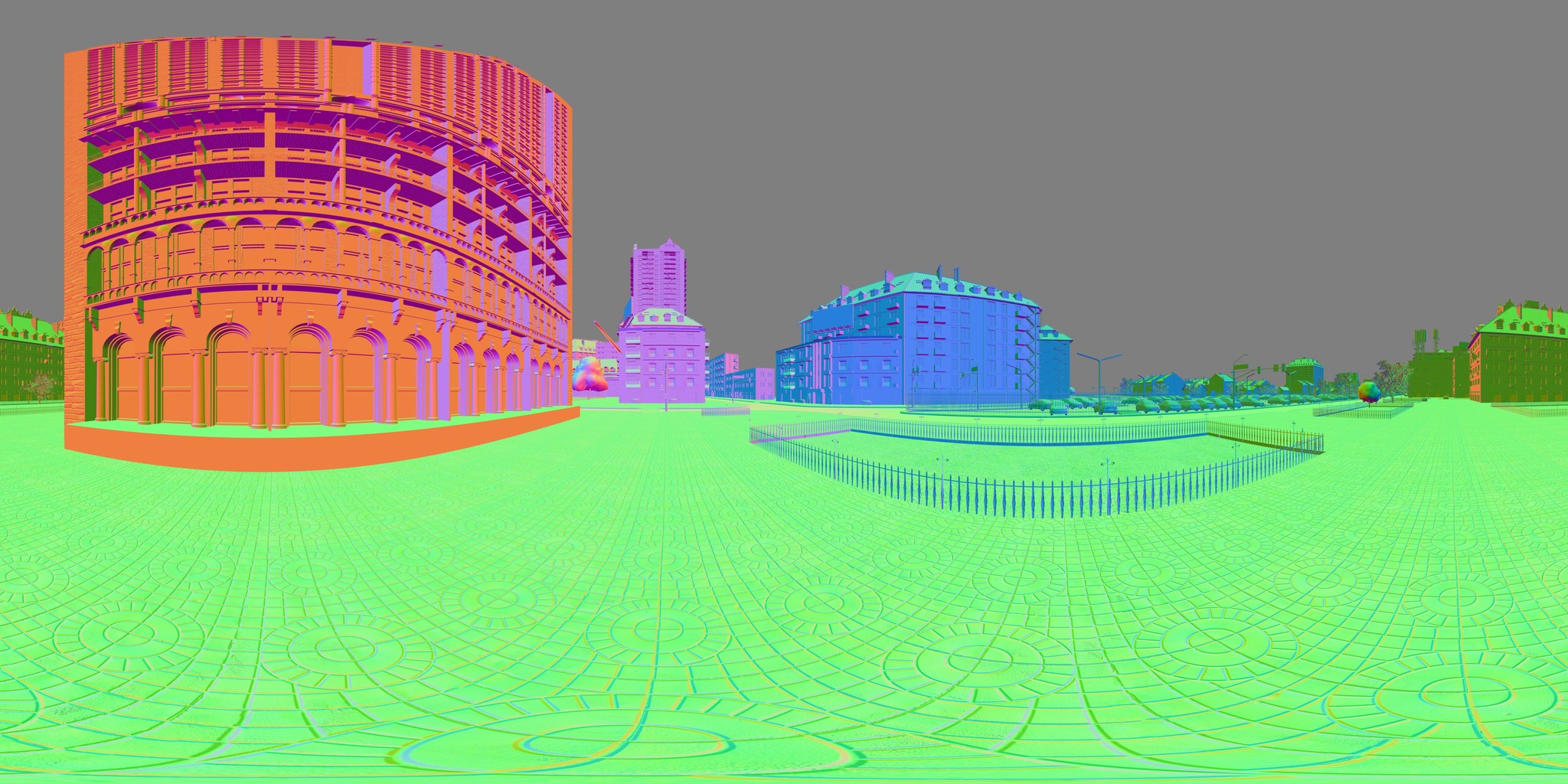}%
    \hfill
    \caption{Examples of RGB, depth, and surface normal panoramas from our PanoInfinigen dataset.}
    \label{fig:panoinfinigen}
\end{figure}

The resulting dataset spans a vast range of semantic domains, from standard indoor environments (\eg, kitchens, bedrooms) to complex natural landscapes (\eg, forests, deserts) and urban surroundings (\eg, skyscrapers, small houses, historical buildings, parks). Crucially, every sample is rendered at native 4K resolution with complete, pixel-perfect ground truth for metric depth and surface normals. Examples of the dataset are shown in Fig.~\ref{fig:panoinfinigen}.

\section{ZüriPano}
\label{sec:zuripano}

Outdoor panoramic data collection introduces significant complexities absent in indoor environments, primarily due to dynamic occlusions (e.g., pedestrians, vehicular traffic, and urban activity) and the limited availability of high-resolution, long-range sensing hardware. To date, these challenges have hindered the development of a reliable, LiDAR-based outdoor panoramic evaluation benchmark. To our knowledge, ZüriPano directly addresses this gap by serving as the first real-world outdoor 360$^\circ$ benchmark with survey-grade ground truth.

For acquisition, we employ the Leica RTC360 LiDAR scanner \cite{leica_rtc360}, a high-performance reality capture system. The device features an effective operating range of 130 meters and utilizes advanced HDR imaging alongside automated double-scan routines to effectively reject transient moving objects. Our collection encompasses 100 terrestrial laser scans across 11 distinct urban locations in Zürich, Switzerland, capturing a diverse array of architectural styles and open-space geometries. Scans are registered using Leica Cyclone REGISTER 360, with each alignment verified against the exported E57 point cloud. The accompanying panoramic imagery is natively captured at 8K resolution and released at 4K ($4096 \times 2048$).

During post-processing, we meticulously filter the data to ensure metric reliability. Because the scanner and camera share a single optical center, the dense Euclidean distance rendered from the co-located scan is fundamentally free of cross-view occlusions. The ground truth is stored as 16-bit depth over a $[0, 200]\,\text{m}$ range; this yields a quantization step of $\approx 3\,\text{mm}$, ensuring that storage precision does not limit the scanner's native accuracy. Infinite depth regions (such as the sky) and annotated reflective surfaces (e.g., windows and glass facades, where LiDAR returns are unreliable) are strictly masked out. Finally, the validity masks are eroded by 5 pixels around all depth discontinuities. We believe this dataset provides a rigorous testbed for evaluating the robustness of panoramic depth estimation models, specifically regarding long-range accuracy and structural consistency in complex urban environments.

\begin{figure}
    \centering
    \includegraphics[width=0.497\linewidth]{images/zuripano/rgb/Arianestrasse_s005.jpg}%
    \hfill
    \includegraphics[width=0.497\linewidth]{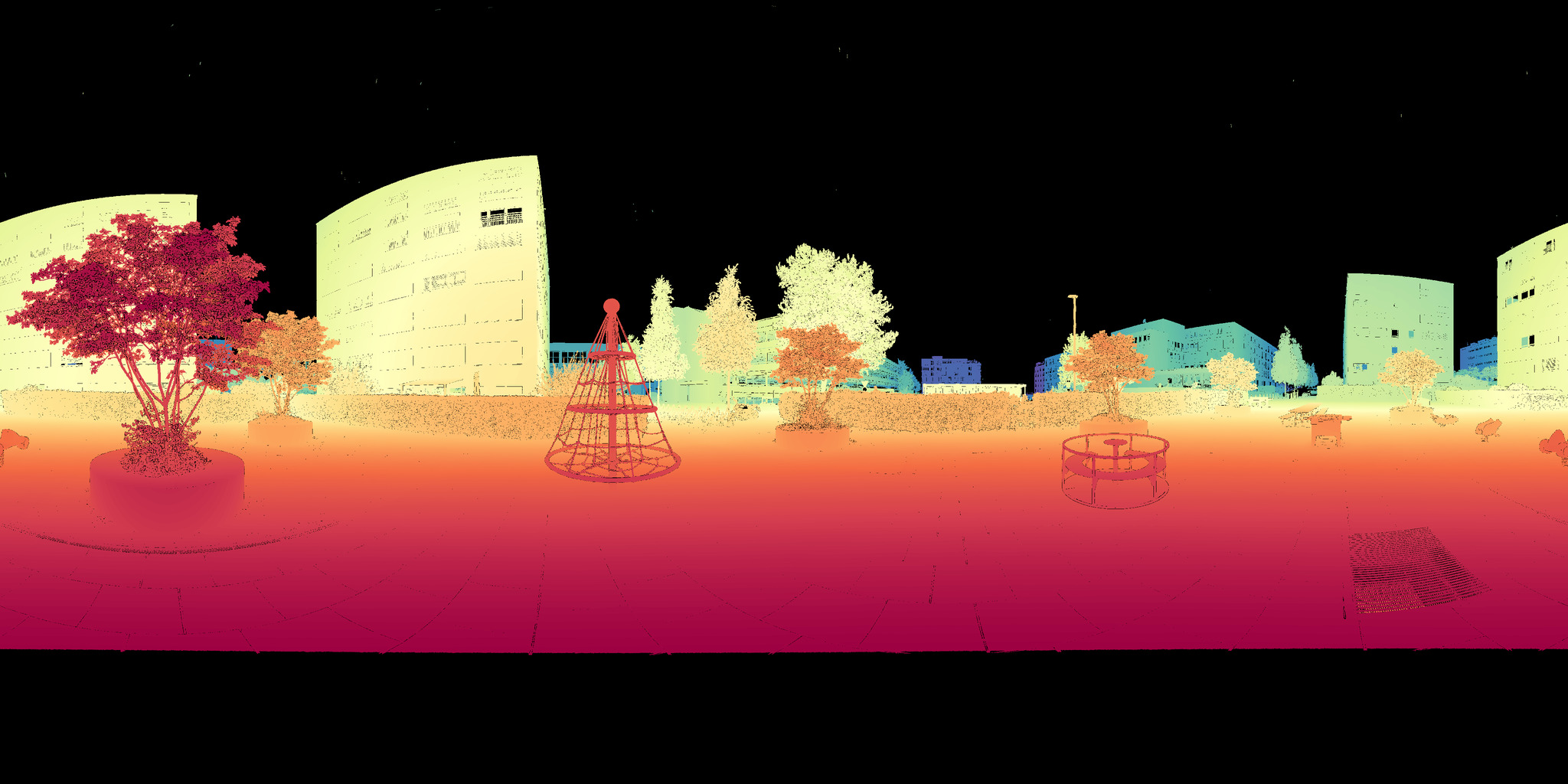}
    \hfill
    \includegraphics[width=0.497\linewidth]{images/zuripano/rgb/Franklinstrasse_s002.jpg}%
    \hfill
    \includegraphics[width=0.497\linewidth]{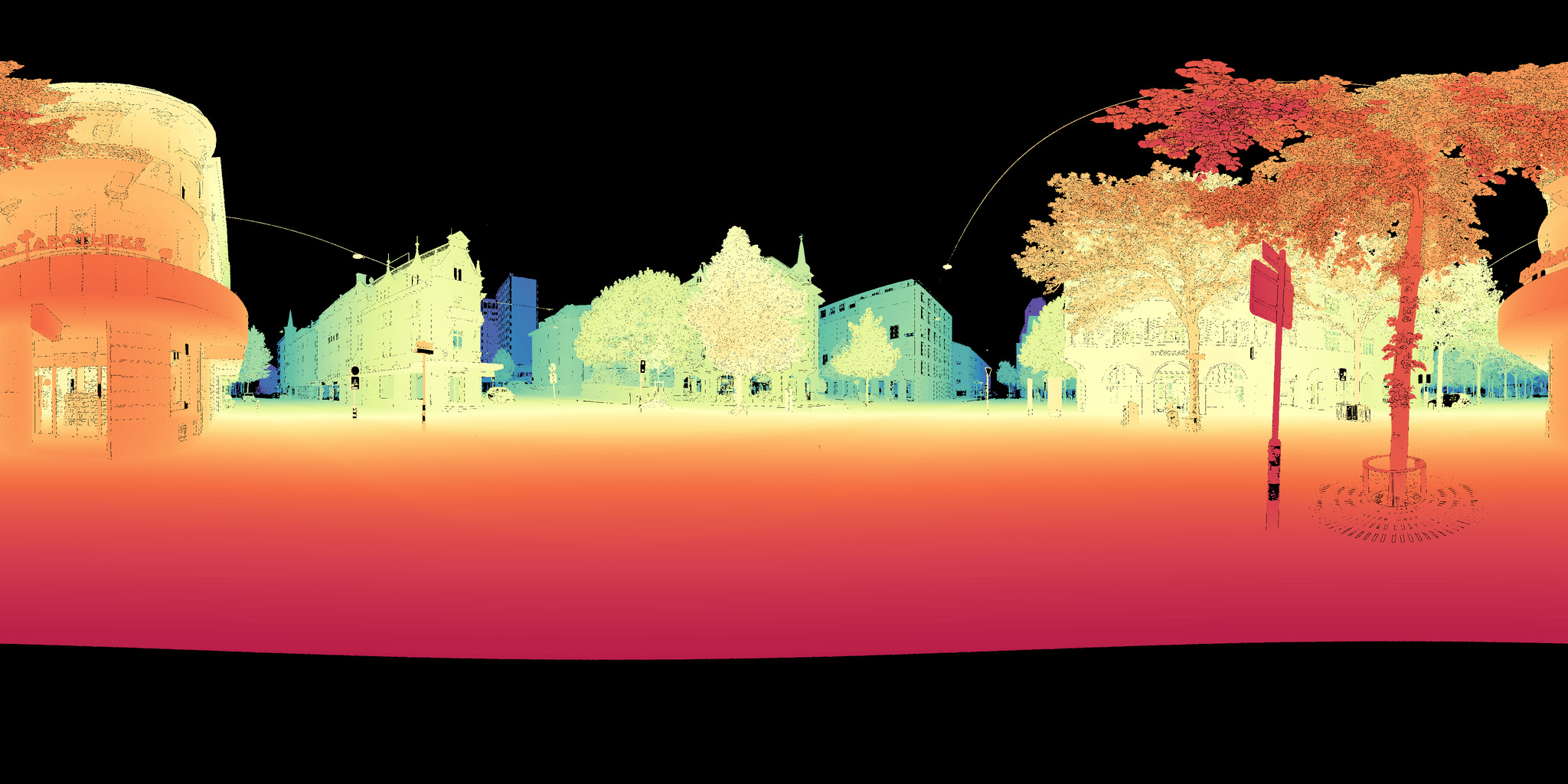}
    \hfill
    \includegraphics[width=0.497\linewidth]{images/zuripano/rgb/Hoenggerberg1_s008.jpg}%
    \hfill
    \includegraphics[width=0.497\linewidth]{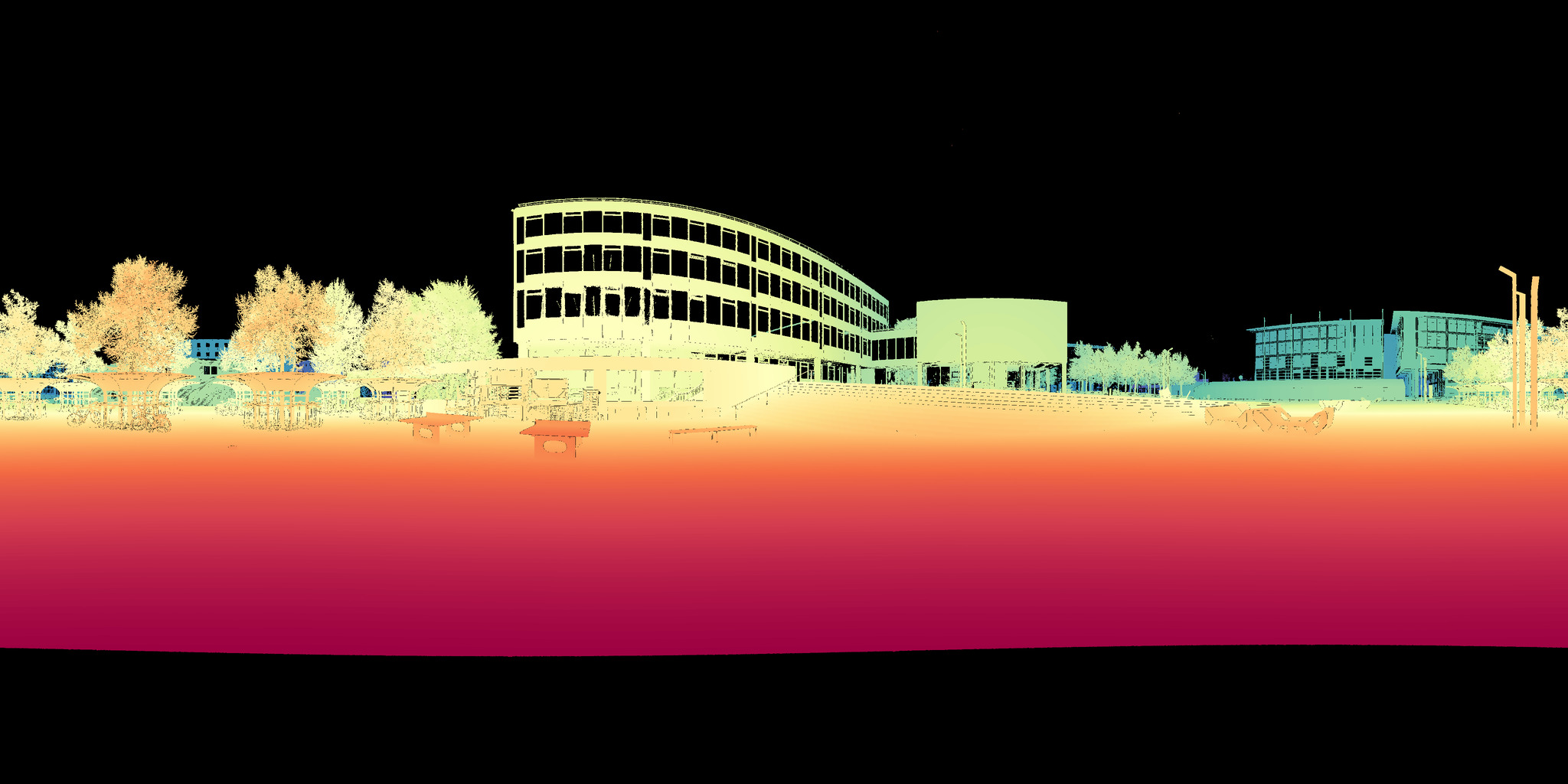}
    \caption{Samples of our ZüriPano dataset.}%
    \hfill
    \label{fig:zuripano}
\end{figure}

\section{Additional Related Work Analysis}
\label{sec:additional_related_work}
In this section, we provide further contextualization of our architectural design choices relative to recent methods, expanding upon the discussion in the main text.

\textbf{Relationship to Recent Multi-Task and Sky-Masking Models.} 
Recent advances in perspective geometry estimation, such as Metric3D v2~\cite{hu2024metric3d} and MoGe~\cite{wang2025moge}, have demonstrated the benefits of joint depth and normal prediction, as well as the utility of masking unbounded regions. Building on these insights, alongside the DA3 architecture, our framework adopts joint multi-task prediction and explicit sky masking. Our specific architectural contribution in this context lies in efficiency. We demonstrate that delegating unbounded sky regions to a lightweight branch branching off the final depth decoder successfully stabilizes metric regression at a negligible computational cost (as evidenced by the significant RMSE spike when omitted, shown in our ablation studies). Furthermore, regarding joint prediction, we show that a parallel normal decoder initialized from depth weights achieves state-of-the-art panoramic normals with minimal fine-tuning. This indicates that the underlying pre-trained perspective transformer (DA3 ViT) already encodes a geometric prior rich enough to support several modalities simultaneously, without necessitating massive joint retraining from scratch.

\textbf{Handling Discontinuities: ERP Padding vs. Cubemap Consistency.} 
Some recent approaches to panoramic geometry, such as 360Anything~\cite{wu2026360anythinggeometryfreeliftingimages} address spatial continuity by employing circular padding (e.g., Circular Latent Encoding) to resolve artifacts at the $\pm 180^{\circ}$ equirectangular (ERP) wrap meridian. However, while ERP-side padding resolves the left-right boundary, it neglects the fact that when multi-face representations (like cubemaps) are used, seams manifest throughout the interior of the unprojected ERP image. 

Recent training-free cubemap adaptations attempt to resolve these interior inconsistencies. Most notably, RPG360~\cite{jung2025rpg360robust360depth} converts the panorama to a six-face cubemap and predicts depth for each face independently using a frozen Metric3D v2 model. It then repairs inconsistencies post-hoc via test-time graph optimization over per-face scales and shifts (requiring up to 480 Adam iterations per inference). Because this correction relies on global scalars per face, it fundamentally cannot resolve dense, per-pixel misalignments at the boundaries. In contrast, PaGeR establishes seamless consistency directly during prediction through a joint feed-forward inference pass. By fine-tuning cross-view attention to synchronize features globally and employing cross-face padding to enforce dense local continuity, our method avoids test-time optimization entirely. This approach roughly halves the Absolute Relative error compared to RPG360 across all benchmarks, operating at significantly lower runtime and higher resolution.

\newpage
\section{Seam Consistency Metrics}
\label{sec:seam_metrics}
To quantify the geometric consistency of our cubemap projections, we evaluate seam artifacts at three distinct granularities: \textbf{Seam Defect Density (SDD)}, \textbf{Seam Prevalence (SP)}, and \textbf{Seam Severity (SS)}. These metrics monitor depth discontinuities across the $N=12$ shared face boundaries of the cubemap.

Let $\mathcal{E}$ be the set of all pixel pairs $(p, q)$ that are spatially adjacent across a cubemap boundary, and let $\{E_k\}_{k=1}^N$ be the partition of $\mathcal{E}$ into $N$ disjoint sets representing each edge. For any pair $(p, q) \in \mathcal{E}$, we define the log-depth jump as:
\begin{equation}
    \Delta_{pq} = |\log \hat{d}_p - \log \hat{d}_q|
\end{equation}
where $\hat d_p$ and $\hat d_q$ are the Euclidean ERP linear depths at pixels $p$ and $q$, respectively.

\paragraph{Seam Defect Density (SDD)} measures the global frequency of artifacts by calculating the fraction of total boundary pixels whose depth jump exceeds a tolerance threshold $\tau$:
\begin{equation}
    \mathrm{SDD} = \frac{1}{|\mathcal{E}|} \sum_{(p,q) \in \mathcal{E}} \mathbb{I}(\Delta_{pq} > \tau)
\end{equation}

\paragraph{Seam Prevalence (SP)} assesses the distribution of defects across the cubemap structure. An edge $E_k$ is considered prevalent with defects if more than 10\% of its constituent pixels exceed $\tau$. The SP metric is the fraction of such "corrupted" edges:
\begin{equation}
    \mathrm{SP} = \frac{1}{N} \sum_{k=1}^{N} \mathbb{I}\left( \frac{1}{|E_k|} \sum_{(p,q) \in E_k} \mathbb{I}(\Delta_{pq} > \tau) > 0.1 \right)
\end{equation}

\paragraph{Seam Severity (SS)} captures systemic geometric misalignment by measuring the fraction of edges whose \textit{mean} jump across the entire boundary exceeds a strict magnitude threshold $\gamma$:
\begin{equation}
    \mathrm{SS} = \frac{1}{N} \sum_{k=1}^{N} \mathbb{I}\left( \frac{1}{|E_k|} \sum_{(p,q) \in E_k} \Delta_{pq} > \gamma \right)
\end{equation}
where $\mathbb{I}(\cdot)$ is the indicator function that equals 1 if the condition is true and 0 otherwise.

In plain words, Seam Defect Density (SDD) measures the local density of pixel-level defects; Seam Prevalence (SP) captures geometric coverage by identifying how many of the twelve edges show visible artifacts; and Seam Severity (SS) identifies systemic collapses by measuring the fraction of edges with high average depth jumps. This triplet allows us to distinguish between widespread, mild jitter and isolated catastrophic failures.

\section{Performance Trade-offs of Joint vs. Independent Model Training}
While our unified panorama estimation model provides the significant advantage of simultaneously outputting scale-invariant (SI) depth, metric depth, surface normals, and sky segmentation from a shared ViT backbone, this multi-task setting inherently requires a compromise in individual task performance. To quantify the impact of joint training and analyze the network's capacity limits, we conducted an ablation study comparing our unified framework against independently trained models.

For this ablation, we trained isolated models for SI depth, surface normals, and a full metric depth model. The independent full metric depth model was trained using dense supervision without depth alignment, thereby directly outputting metric depth. 

As expected in multi-task architectures, the results demonstrate that the independent SI depth and surface normal models modestly outperform their counterparts in the unified model. However, the most substantial performance gap is observed between our independent full metric depth model and the metric scale head within the unified setting. 

This discrepancy is directly tied to gradient flow and backbone freezing constraints. For the independent full metric model, the ViT backbone is fully unlocked, allowing the network to extract metric-specific cues early in the encoder stages. This deep integration greatly enhances metric depth accuracy and effectively obviates the need for separate, heavy metric heads. 

Conversely, in the unified model, the ViT backbone must remain frozen with respect to the metric scale head. We found that allowing gradients from the metric scale head to flow back into the shared ViT during joint training fundamentally conflicts with the optimization of the other tasks, leading to a severe degradation in the quality of both SI depth and surface normal predictions. Thus, keeping the ViT frozen preserves the integrity of the SI depth and normals but creates a bottleneck for metric depth performance.

These findings suggest that while our shared representation is highly efficient, it is currently capacity-constrained when balancing the extraction of both scale-invariant and purely metric cues. A careful adaptation of the ViT backbone---such as introducing task-specific routing or slightly expanding capacity to better accommodate these competing gradients---represents a highly promising direction for future work.

\begin{table*}[t]
    \centering
    
    \begin{subtable}{\textwidth}
        \centering
        \resizebox{\textwidth}{!}{
            \begin{tabular}{l ccc c ccc c ccc}
                \toprule
                & \multicolumn{3}{c}{\textbf{Matterport3D360}} && \multicolumn{3}{c}{\textbf{Stanford2D3DS}} && \multicolumn{3}{c}{\textbf{ZüriPano}} \\
                \cmidrule{2-4} \cmidrule{6-8} \cmidrule{10-12}
                \textbf{Model} & AbsRel$\downarrow$ & RMSE$\downarrow$ & $\delta_1$$\uparrow$ && AbsRel$\downarrow$ & RMSE$\downarrow$ & $\delta_1$$\uparrow$ && AbsRel$\downarrow$ & RMSE$\downarrow$ & $\delta_1$$\uparrow$ \\
                \midrule
                
                \multicolumn{12}{l}{\textit{SI Depth}} \\
                Unified     & \phantom{0}9.67 & \phantom{0}64.69 & 90.87 && \phantom{0}5.93 & \phantom{0}35.34 & 96.10 && \phantom{0}9.36 & 299.61 & 94.75 \\
                 
                Specialized & \textbf{\phantom{0}9.57} & \textbf{\phantom{0}63.22} & \textbf{91.02} && \textbf{\phantom{0}5.87} & \textbf{\phantom{0}34.07} & \textbf{96.15} && \textbf{\phantom{0}9.07} & \textbf{274.92} & \textbf{95.31} \\
                \midrule
                
                \multicolumn{12}{l}{\textit{Metric Depth}} \\
                Unified     & 21.83 & 123.48 & 69.50 && 10.94 & \textbf{\phantom{0}45.43} & 90.94 && 31.97 & 530.85 & 39.30 \\
                 
                Specialized & \textbf{17.71} & \textbf{\phantom{0}91.27} & \textbf{86.56} && \textbf{10.01} & \phantom{0}46.28 & \textbf{92.07} && \textbf{18.19} & \textbf{335.90} & \textbf{77.90} \\
                \bottomrule
            \end{tabular}
        }
        \caption{Depth Estimation (SI and Metric)}
    \end{subtable}
    
    \vspace{4mm}
    
    \begin{subtable}{\textwidth}
        \centering
        \begin{tabular}{l cc cc}
            \toprule
            & \multicolumn{4}{c}{\textbf{Structured3D}} \\
            \cmidrule{2-5}
            \textbf{Model} & Mean$\downarrow$ & MSE$\downarrow$ & $5^\circ$$\uparrow$ & $22.5^\circ$$\uparrow$ \\
            \midrule
            
            \multicolumn{5}{l}{\textit{Surf. Normals}} \\
            Unified     & \phantom{0}5.49 & 174.9 & 79.91 & 92.83 \\
             
            Specialized & \textbf{\phantom{0}4.96} & \textbf{157.4} & \textbf{82.18} & \textbf{93.53} \\
            \bottomrule
        \end{tabular}
        \caption{Surface Normals}
    \end{subtable}
    
    \vspace{2mm}
    \caption{\textbf{Unified vs. Specialized Models.} We compare our multi-task unified PaGeR model against individually trained specialized models for each modality. \textbf{(a)} Evaluation of Scale-Invariant (SI) and Metric depth across three datasets. \textbf{(b)} Surface normal estimation evaluated on the Structured3D dataset. Top-performing metrics for each modality pair are highlighted in \textbf{bold}.}
    \label{tab:ablation_modalities}
\end{table*}

\clearpage
\newpage
\section{Loss Weights and Hyperparameters}

To facilitate full reproducibility and provide transparency regarding our training setup, we detail the complete configuration and computational infrastructure of our framework. This includes scaling factors for the scale-invariant depth, metric scale, surface normal, and sky segmentation heads. For clarity and ease of reference, we present the precise loss weights $\lambda$, optimization schedules, learning rates, and architectural choices across all modalities in Table~\ref{tab:hyperparam}.
\begin{table}[!bh]
\centering
\begin{tabular}{llcc}
\toprule
\textbf{Modality} & \textbf{Loss Term} & \textbf{Symbol} & \textbf{Weight} \\
\midrule
\textbf{Depth}    & Confidence L1 Loss & $\lambda_{L1}$   & \phantom{0}1.0  \\
                  &   Lambda Confidence  & $\lambda_{c}$ & \phantom{0}0.2  \\
                  & Gradient Loss & $\lambda_{grad}$ & 40.0  \\
                  & Normals Consistency Loss & $\lambda_{norm}$ & \phantom{0}0.6  \\
\midrule
\textbf{Normals}  & Cosine Similarity Loss & $\lambda_{cos}$  &  \phantom{0}1.0 \\
                  & Perceptual Loss & $\lambda_{perc}$  & \phantom{0}0.5  \\
\midrule
\textbf{Metric Scale} & Confidence L1 Loss    & $\lambda_{L1}$   & \phantom{0}1.0  \\
&   Lambda Confidence  & $\lambda_{c}$ & \phantom{0}0.2  \\
&  DPT Subsampling Factor & $F$ & \phantom{0}4.0  \\
\midrule
\textbf{Sky Segmentation} & Binary Cross-Entropy Loss   & $\lambda_{BCE}$   & \phantom{0}1.0  \\
& Focal Loss & $\lambda_{Focal}$   & \phantom{0}0.4  \\
& Dice Loss  & $\lambda_{Dice}$   & \phantom{0}1.0  \\
\bottomrule
\end{tabular}
\caption{Hyperparameter configuration for our default model.}
\label{tab:hyperparam}
\end{table}

\section{Evaluation of Alternative DA3 Input Configurations}
\label{sec:alternative_DA3}
To ensure a fair and rigorous comparison against the vanilla DA3 baseline presented in Table 4a of the main text, we investigated whether alternative multi-view projection strategies could yield better zero-shot panoramic performance. While our framework relies on a non-overlapping 6×90° cubemap, it is a reasonable hypothesis that a pre-trained multi-view model like DA3 might benefit from explicit overlapping regions, denser view sampling, or independent single-view processing. 

To test this, we implemented and evaluated several alternative input configurations for the vanilla DA3 model on the Stanford2D3DS scale-invariant depth benchmark:

\begin{itemize}
    \item \textbf{(a) 6×110° overlapping cubemap (joint):} The standard six cubemap faces widened to a 110° field of view (FoV), providing approximately 20° of overlap between adjacent views.
    \item \textbf{(b) 12×90° icosahedral (joint):} A twelve-view arrangement distributed across an icosahedron, mirroring the official panorama processing pipeline of recent methods like MoGe.
    \item \textbf{(c) 20×70° icosahedral (joint):} A denser, smaller-FoV configuration using tangent views at the centers of icosahedron faces, resulting in roughly 28° of overlap. This configuration processes the views jointly but consumes approximately 3.3× the pixel budget of our standard cubemap.
    \item \textbf{(d) 20×70° independent:} Using the same dense 20-view configuration as (c), but predicting depth for each view in complete isolation. The final panorama is reconciled by solving for per-view scale factors via a global least-squares optimization over the overlap graph.
\end{itemize}

\begin{table}[h]
\centering
\small
\renewcommand{\arraystretch}{1.1}
\setlength{\tabcolsep}{8pt}
\begin{tabular}{lcc}
    \specialrule{0.75pt}{0pt}{1pt}
    \textbf{DA3 Input Configuration} & \textbf{AbsRel $\downarrow$} & \textbf{$\delta_1$ $\uparrow$} \\
    \hline
    \textbf{6×90° cubemap, joint (Table 4a)} & \textbf{\phantom{0}6.58} & \textbf{95.94} \\
    (a) 6×110° overlapping cubemap, joint & \phantom{0}9.37 & 93.58 \\
    (b) 12×90° icosahedral, joint & \phantom{0}8.33 & 94.58 \\
    (c) 20×70° icosahedral, joint & \phantom{0}7.65 & 94.95 \\
    (d) 20×70° independent, LS-aligned stitch & \phantom{0}9.45 & 92.72 \\
    \hline
    \rowcolor{gray!15} PaGeR (Ours, reference) & \phantom{0}5.93 & 96.10 \\
    \specialrule{0.75pt}{1pt}{1pt}
\end{tabular}
\caption{\textbf{Comparison of alternative DA3 input configurations on the Stanford2D3DS dataset (scale-invariant depth).} Metrics are scaled by ×100. Contrary to expectations, the non-overlapping 6×90° cubemap provides the strongest baseline performance for the vanilla model.}
\label{tab:da3_ablations}
\end{table}

As shown in Table~\ref{tab:da3_ablations}, despite the intuition that explicit overlaps or denser view sampling might aid the pre-trained perspective model, the standard non-overlapping 6×90° cubemap achieves the best quantitative results among all vanilla DA3 configurations. 

Furthermore, to ensure that the stitching mechanism was not bottlenecking the performance of the overlapping configurations, we tested four distinct compositing rules on the best-performing alternative (configuration c): feathered blending, uniform mean, nearest view selection, and per-pixel median. Across all four compositing strategies, the variance in AbsRel was $\le 0.24$. Because the consensus choice has a negligible impact on overall performance, the standard 6×90° cubemap remains the most robust and computationally efficient baseline against which to evaluate our proposed adaptations.

\section{Analysis of Multi-View Consistency}

To comprehensively evaluate the robustness of our representations, we conducted an additional ablation to assess the multi-view geometric consistency of our framework relative to prior state-of-the-art methods. Specifically, we investigate whether the model predicts geometrically compatible 3D outputs for two distinct panoramic frames of the same scene captured with a small camera offset.

\textbf{Evaluation Protocol.} Given two panoramic frames (Frame A and Frame B) of a scene from nearby viewpoints with known relative poses, we predict the depth for each frame independently. Following our standard scale-invariant evaluation protocol, we scale-align each prediction to its respective ground truth. We then unproject the predicted depths into 3D point clouds and place both reconstructions into a shared world coordinate frame.

We quantify multi-view consistency using two complementary metrics:
\begin{itemize}
    \item \textbf{Inter-view Reprojection AbsRel:} This metric projects each predicted 3D point from Frame A into the camera view of Frame B. It then computes the Absolute Relative Error between the projected range of Point A and the actual depth predicted by Frame B along that specific ray.
    \item \textbf{Co-visible Chamfer Distance (\% of scene depth):} A correspondence-free nearest-neighbor distance metric computed between the two overlapping point clouds. This metric is particularly sensitive to global misalignments, such as scale drift or systemic warping between the two independent predictions.
\end{itemize}

\textbf{Results.} We evaluated this consistency on a zero-shot basis using 40 real-world panorama pairs extracted from 5 test buildings within the Matterport3D dataset (with baseline distances ranging from $0.30$ to $0.97$ meters). 

As demonstrated in Table~\ref{tab:multiview_consistency}, \name{} achieves the highest multi-view consistency among all compared methods. It significantly outperforms optimization-based approaches like RPG360 and continuous ERP models like $\mathrm{DA}^2$. This confirms that the strong perspective priors preserved by our cubemap architecture not only yield highly accurate single-view geometry but also naturally result in robust, geometrically stable structures across novel viewpoints without requiring multi-view test-time optimization.

\begin{table}[H]
\centering
\small
\renewcommand{\arraystretch}{1.1}
\setlength{\tabcolsep}{12pt}
\begin{tabular}{lcc}
    \specialrule{0.75pt}{0pt}{1pt}
    \textbf{Method} & \textbf{Reproj. AbsRel (\%) $\downarrow$} & \textbf{Chamfer (\% depth) $\downarrow$} \\
    \hline
    GT floor (Reference) & \phantom{0}0.07 & 0.95 \\
    \hline
    UniK3D      & 58.34 & 4.05 \\
    PanDA       & 25.78 & 8.15 \\
    DA$^2$      & \phantom{0}3.98 & 3.05 \\
    RPG360      & \phantom{0}3.71 & 3.71 \\
    \rowcolor{gray!15} \textbf{PaGeR (Ours)} & \textbf{\phantom{0}2.48} & \textbf{2.26} \\
    \specialrule{0.75pt}{1pt}{1pt}
\end{tabular}
\caption{\textbf{Multi-view consistency on real Matterport3D panorama pairs (Zero-shot).} Both metrics evaluate the alignment of two independent single-view predictions when unprojected into a shared 3D space.}
\label{tab:multiview_consistency}
\end{table}

\section{Computational Efficiency and Resource Benchmark}

To evaluate the practical utility and scalability of \name{}, we benchmark its computational footprint against other baselines. Table~\ref{tab:method_comparison} provides a comprehensive comparison across four key performance dimensions: inference runtime, peak memory consumption during evaluation, native processing resolution, and total training data requirements.

\begin{table}[!bh]
\centering
\begin{tabular}{lcccc} 
\toprule
Method name & Runtime [s] & Peak Memory [GB] & Resolution [px] & Data size [$10^3$] \\
\midrule
DreamCube      & \phantom{0}6.06 & 15.2 & 1024$\times$2048 & \phantom{0}N/A \\
DepthAnyCamera & \phantom{0}0.11 & \phantom{0}0.3 & \phantom{0}512$\times$1024 & \phantom{0}800 \\
MoGe           & 36.05 & \phantom{0}3.6 & 1024$\times$2048 & 8860 \\
DAP            & \phantom{0}0.10 & \phantom{0}3.7 & \phantom{0}512$\times$1024 & 1700 \\
RPG360         & \phantom{0}5.83 & 13.2 & \phantom{0}512$\times$1024 & \phantom{0}N/A \\
EGformer       & \phantom{0}0.31 & \phantom{0}1.2 & \phantom{0}512$\times$1024 & \phantom{00}70 \\
UniK3D         & \phantom{0}0.41 & \phantom{0}3.3 & \phantom{0}560$\times$1106 & \phantom{0}700 \\
PanDA          & \phantom{0}0.63 & \phantom{0}3.5 & \phantom{0}504$\times$1008 & \phantom{0}120 \\
DA$^2$         & \phantom{0}0.05 & \phantom{0}2.8 & \phantom{0}546$\times$1092 & \phantom{0}606 \\
\rowcolor{gray!15}
Ours           & \phantom{0}0.48 & 12.8 & 1008$\times$2016 & \phantom{0}100 \\
\bottomrule
\end{tabular}
\caption{Comparison of various methods based on performance metrics.}
\label{tab:method_comparison}
\end{table}
\newpage
\section{Additional Quantitative Results}
\label{sec:additional_quantitative_metrics}
\begin{table*}[th]
\centering
\small
\renewcommand{\arraystretch}{1.1}
\setlength{\tabcolsep}{3pt}

\resizebox{0.7\textwidth}{!}{%
\begin{tabular}{l ccc c ccc}
    \specialrule{0.75pt}{0pt}{1pt}
    & \multicolumn{3}{c}{\textbf{Matterport3D360}} & & \multicolumn{3}{c}{\textbf{Stanford2D3DS}} \\
    \hhline{~---~---}
    Method & AbsRel $\downarrow$ & RMSE $\downarrow$ & $\delta_1$ $\uparrow$ & & AbsRel $\downarrow$ & RMSE $\downarrow$ & $\delta_1$ $\uparrow$ \\
    \hline
    DAP~\cite{lin2025depthpanoramasfoundationmodel}   & 13.44 & 43.36 & 84.90 & & \phantom{0}8.08 & 25.57 & 93.97 \\
    RPG360~\cite{jung2025rpg360robust360depth}        & 11.79 & 38.78 & 87.83 & & 10.50 & 29.74 & 89.33 \\
    UniK3D$^\dagger$~\cite{piccinelli2025unik3d}      & 13.12 & \underline{38.36} & 86.69 & & \phantom{0}9.25 & 26.18 & 93.83 \\
    PanDA*~\cite{cao2025panda}                        & \underline{10.64} & 38.71 & \underline{88.39} & & \phantom{0}7.92 & 23.18 & 94.97 \\
    $\mathrm{DA}^2$~\cite{li20252}                    & 10.81 & 38.38 & 87.01 & & \phantom{0}\underline{6.99} & \underline{23.00} & \underline{95.95} \\
    \rowcolor{gray!15} \textbf{PaGeR (Ours)}          & \phantom{0}\textbf{9.95} & \textbf{37.90} & \textbf{88.73} & & \phantom{0}\textbf{5.33} & \textbf{21.93} & \textbf{96.06} \\
\specialrule{0.75pt}{1pt}{1pt}
\end{tabular}
}
\caption{\textbf{Quantitative comparison between \name{} and state-of-the-art baselines on Matterport3D360 and Stanford2D3DS in narrow-depth setting ($[0, 5]\,\text{m}$).} Best and second-best results are indicated in \textbf{bold} and \underline{underlined}, respectively. Methods optimized using in-domain training are marked with $\dagger$, and affine-invariant methods are denoted with \textsuperscript{*}.}\label{tab:narrow_depth_results}
\end{table*}

\begin{table*}[th]
\centering
\small
\renewcommand{\arraystretch}{1.1}
\setlength{\tabcolsep}{3pt}

\resizebox{\textwidth}{!}{%
\begin{tabular}{l ccc c ccc c ccc}
    \specialrule{0.75pt}{0pt}{1pt}
    & \multicolumn{3}{c}{\textbf{Matterport3D360}} & & \multicolumn{3}{c}{\textbf{Stanford2D3DS}} & & \multicolumn{3}{c}{\textbf{ZüriPano}} \\
    \hhline{~---~---~---}
    Method & AbsRel $\downarrow$ & RMSE $\downarrow$ & $\delta_1$ $\uparrow$ & & AbsRel $\downarrow$ & RMSE $\downarrow$ & $\delta_1$ $\uparrow$ & & AbsRel $\downarrow$ & RMSE $\downarrow$ & $\delta_1$ $\uparrow$ \\
    \hline
    DAP~\cite{lin2025depthpanoramasfoundationmodel}   & 15.76 & 85.04 & 82.61 & & \phantom{0}9.76 & 46.84 & 92.29 & & 19.83 & 570.90 & 73.01 \\
    RPG360~\cite{jung2025rpg360robust360depth}        & 15.41 & 79.65 & 82.39 & & 11.92 & 47.08 & 87.71 & & \underline{18.22} & \underline{457.16} & \underline{78.59} \\
    UniK3D$^\dagger$~\cite{piccinelli2025unik3d}      & 14.82 & 69.20 & 83.69 & & \phantom{0}9.92 & 40.69 & 93.32 & & 28.09 & 794.97 & 46.52 \\
    PanDA*~\cite{cao2025panda}                        & 13.83 & 82.87 & 84.20 & & 10.69 & 55.80 & 90.70 & & 25.05 & 507.48 & 56.83 \\
    $\mathrm{DA}^2$~\cite{li20252}                    & \underline{11.07} & \underline{67.77} & \underline{89.16} & & \phantom{0}\underline{7.68} & \underline{37.50} & \underline{95.57} & & 60.53 & 859.89 & \phantom{0}2.22 \\
    \rowcolor{gray!15} \textbf{PaGeR (Ours)}          & \phantom{0}\textbf{9.69} & \textbf{64.86} & \textbf{90.85} & & \phantom{0}\textbf{6.01} & \textbf{35.96} & \textbf{96.02} & & \phantom{0}\textbf{9.47} & \textbf{305.08} & \textbf{94.62} \\
\specialrule{0.75pt}{1pt}{1pt}
\end{tabular}
}
\caption{\textbf{Quantitative comparison between \name{} and state-of-the-art baselines across Matterport3D360, Stanford2D3DS, and ZüriPano on downscaled $512 \times 1024$ resolution.} Best and second-best results are indicated in \textbf{bold} and \underline{underlined}, respectively. Methods optimized using in-domain training are marked with $\dagger$, and affine-invariant methods are denoted with \textsuperscript{*}.}\label{tab:depth_results_downscaled}
\end{table*}

\clearpage
\newpage
\section{Additional Qualitative Examples}
\label{sec:additional_qualitative_comparisons}
\usetikzlibrary{spy,decorations.pathreplacing,angles,quotes,calc,arrows,positioning,}
\definecolor{spycolor}{RGB}{150,150,200}
\tikzset{
    rectspy/.default={lens={scale=3}, size=4cm},
    rectspy on/.style={#1,},
    rectspy/.style={
        draw=spycolor,
        connect spies,
        spy scope={
        every spy on node/.style={
            draw=spycolor,
            very thick,
            rectangle, 
            rectspy on,
        },
        every spy in node/.style={
            draw=spycolor,
            very thick,
            rectangle,
        },
        #1
        },
        spy connection path={\draw[spycolor, very thick] (tikzspyonnode) -- (tikzspyinnode);}
    }
}
\tikzset{
    sepbar/.style={
        very thick,
        black!30!white,
    }
}

\begin{figure}[!h]
\resizebox{\textwidth}{!}{%
\begin{tikzpicture}[inner sep=0,rectspy={lens={scale=2.8}, width=1.85cm, height=0.95cm}]

\node[label={[inner sep=0pt,anchor=mid,rounded corners,yshift=4em, xshift=3em]below:{Matterport3D360}}] at (4*0-0.7,0) {\includegraphics[width=1.95cm]{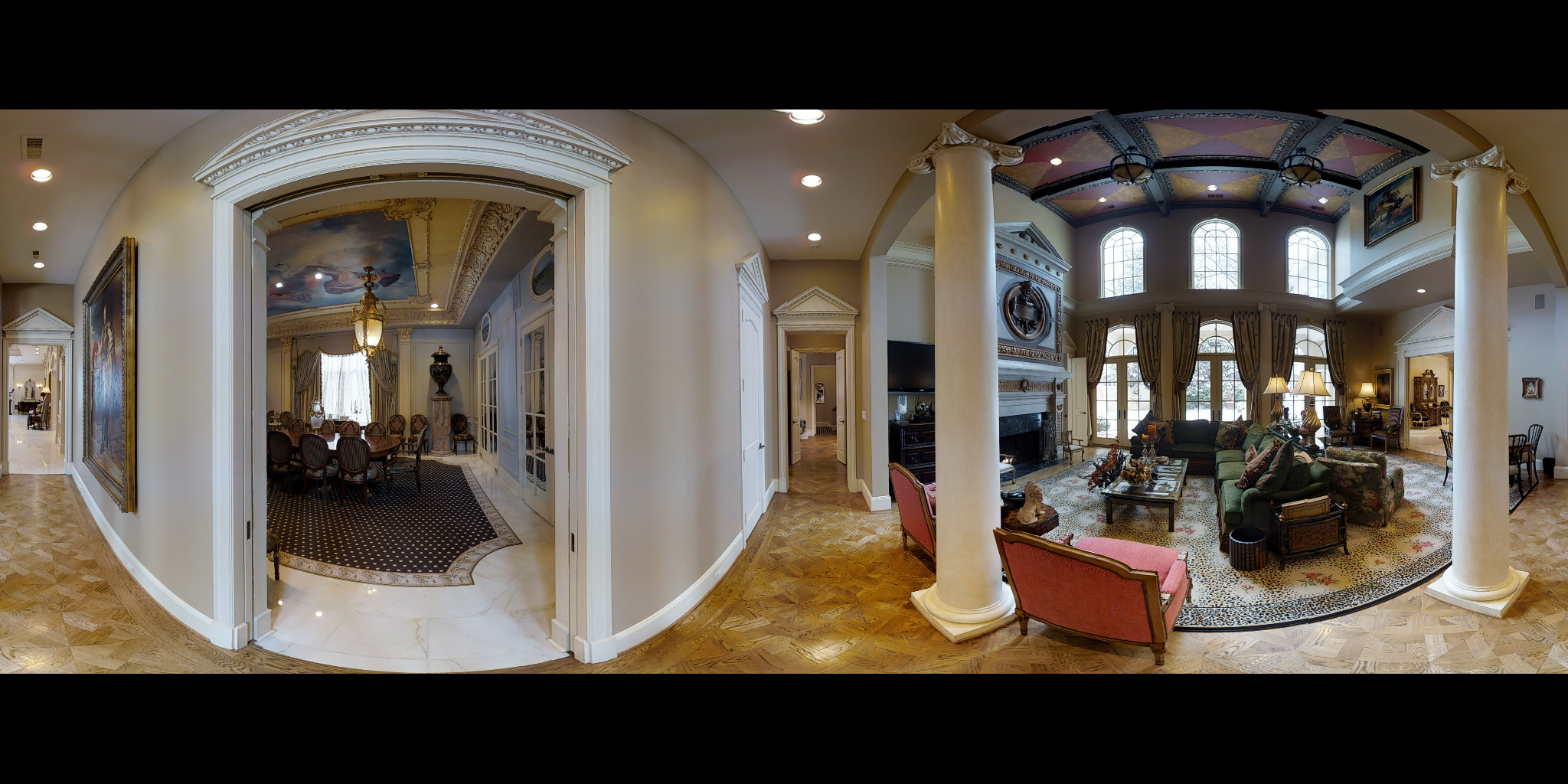}};
\node[label={[inner sep=0pt,anchor=mid,rounded corners,yshift=4em, xshift=3em]below:{Stanford2D3DS}}] at (4*1-0.7,0) {\includegraphics[width=1.95cm]{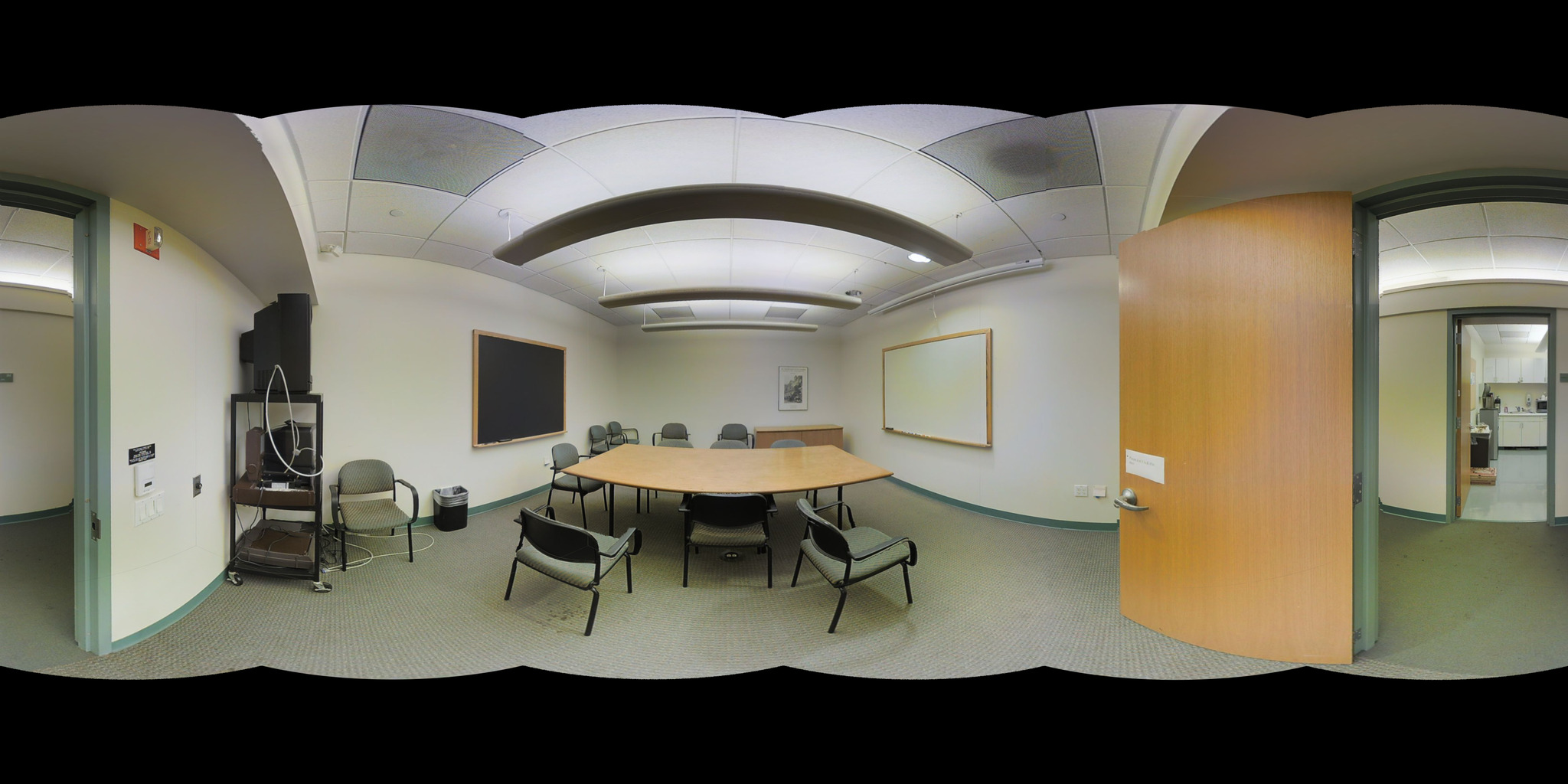}};
\node[label={[inner sep=0pt,anchor=mid,rounded corners,yshift=4em, xshift=3em]below:{ZüriPano}}] at (4*2-0.7,0) {\includegraphics[width=1.95cm]{images/depth_comparisons/Velopalast014/rgb.jpg}};
\node[rotate=90] at (-2.3,-0) {\footnotesize Input};

\node at (4*0-0.7,-1) {\includegraphics[width=1.95cm]{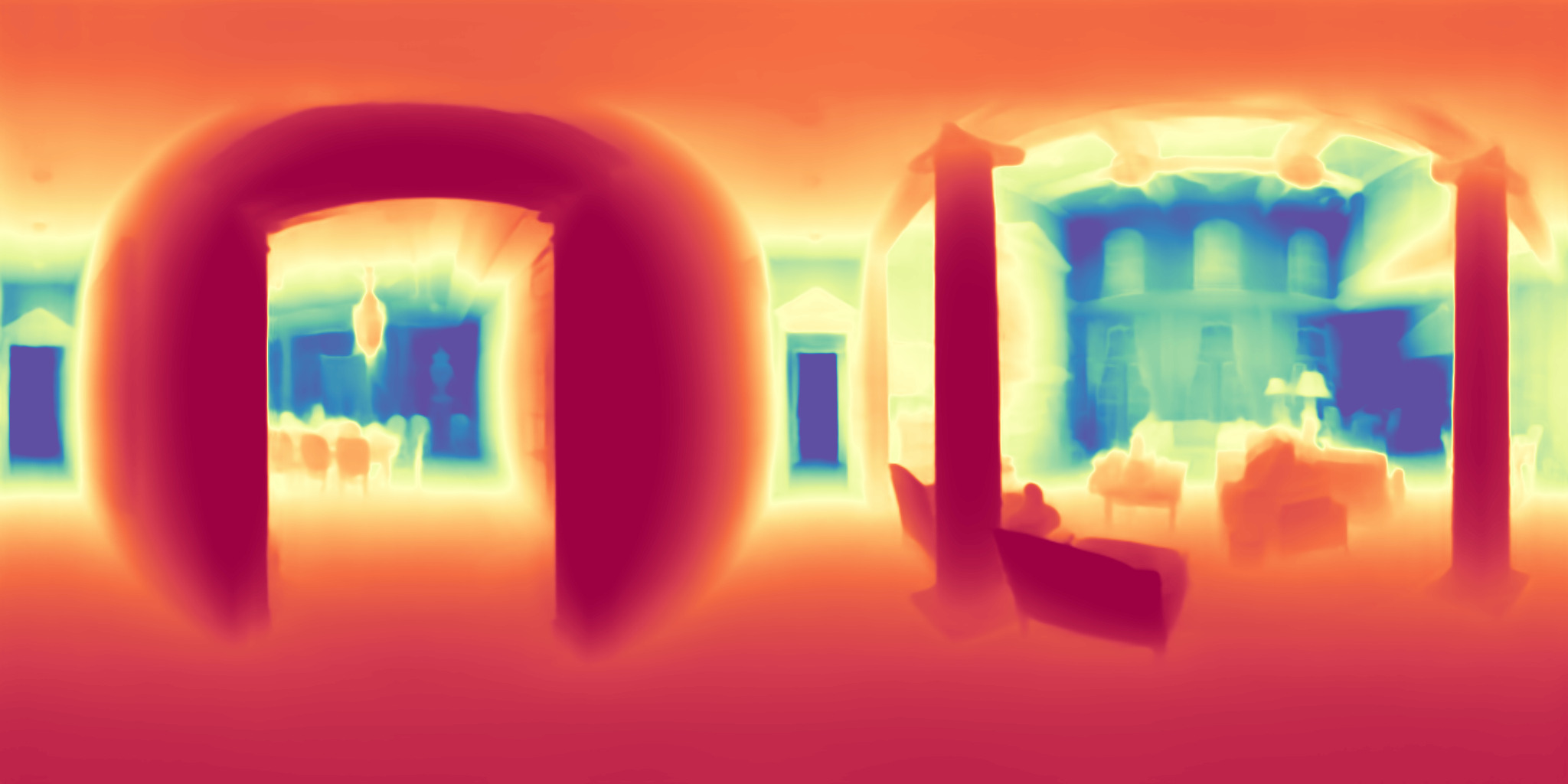}};
\node at (4*1-0.7,-1) {\includegraphics[width=1.95cm]{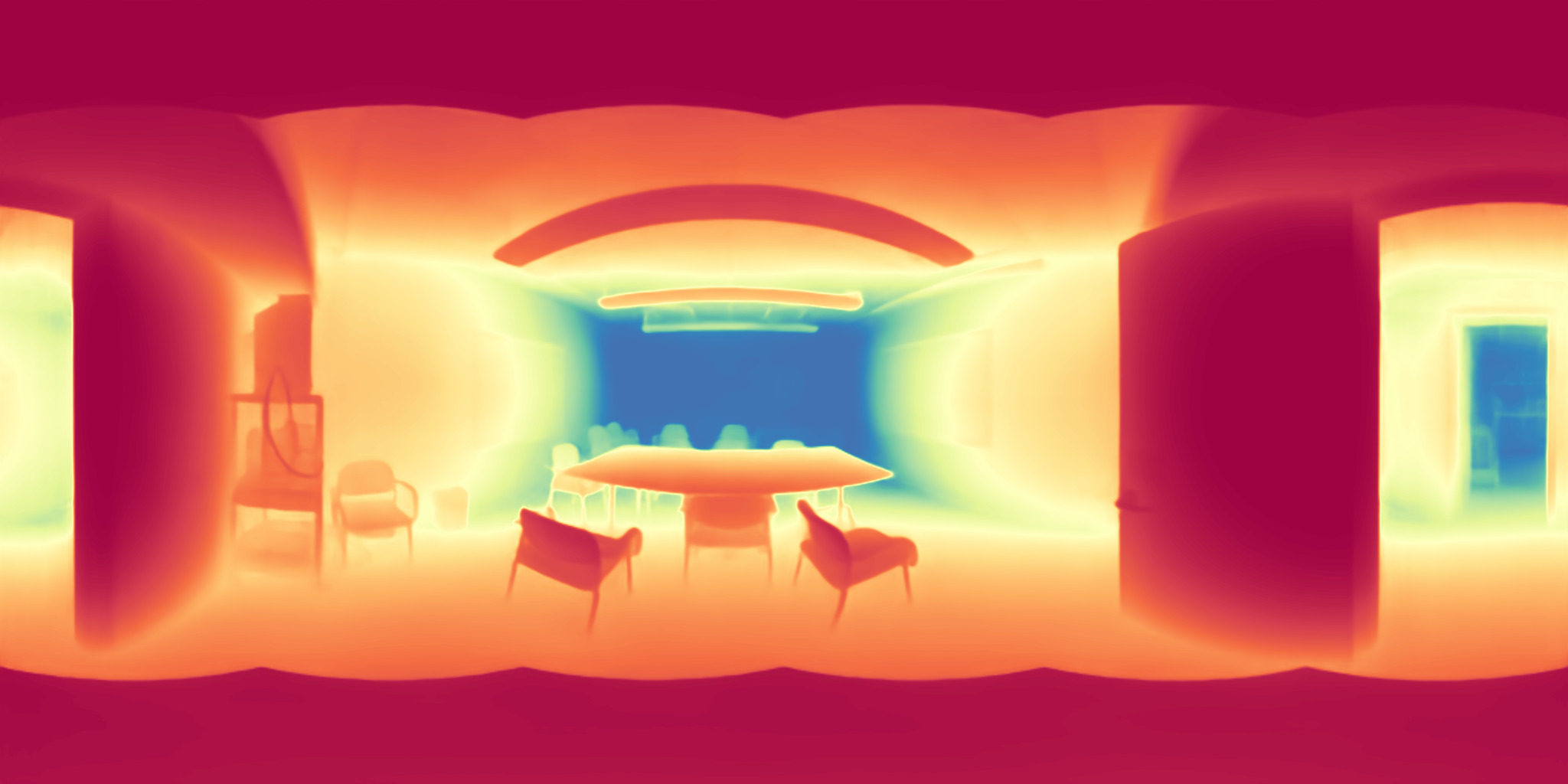}};
\node at (4*2-0.7,-1) {\includegraphics[width=1.95cm]{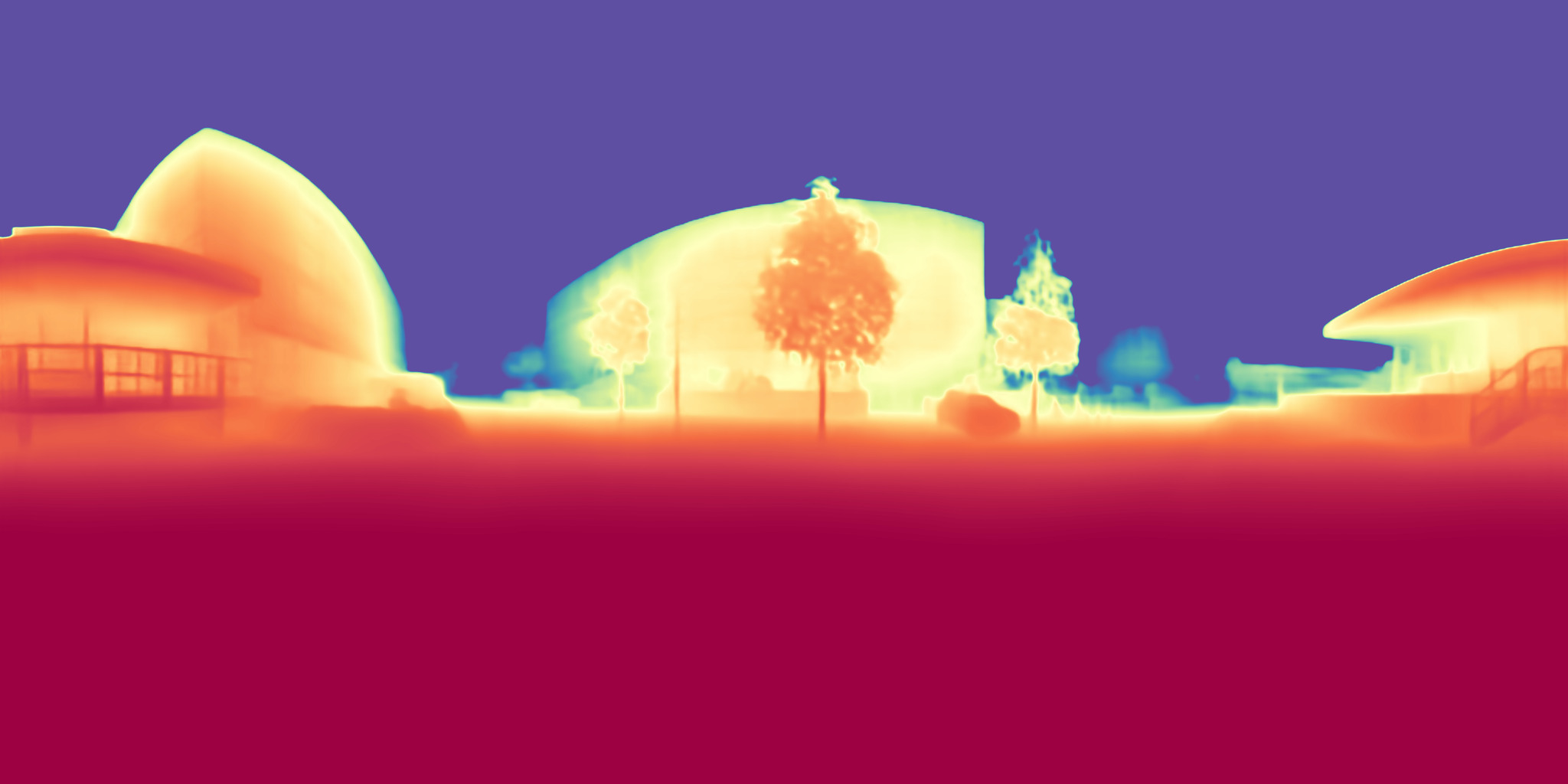}};
\node[rotate=90] at (-2.3,-1)
    {\scriptsize PanDA};

\node at (4*0-0.7,-2) {\includegraphics[width=1.95cm]{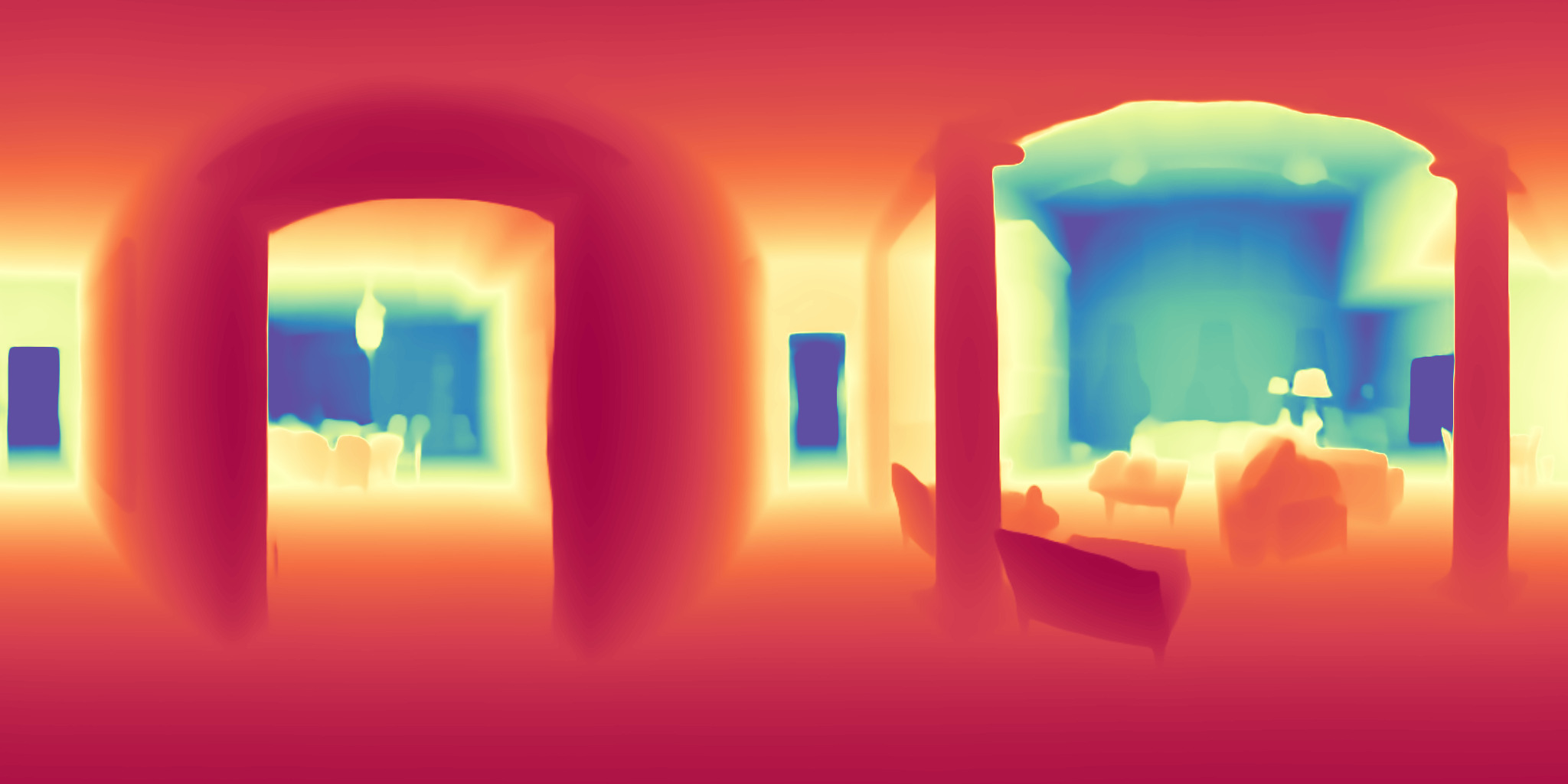}};
\node at (4*1-0.7,-2) {\includegraphics[width=1.95cm]{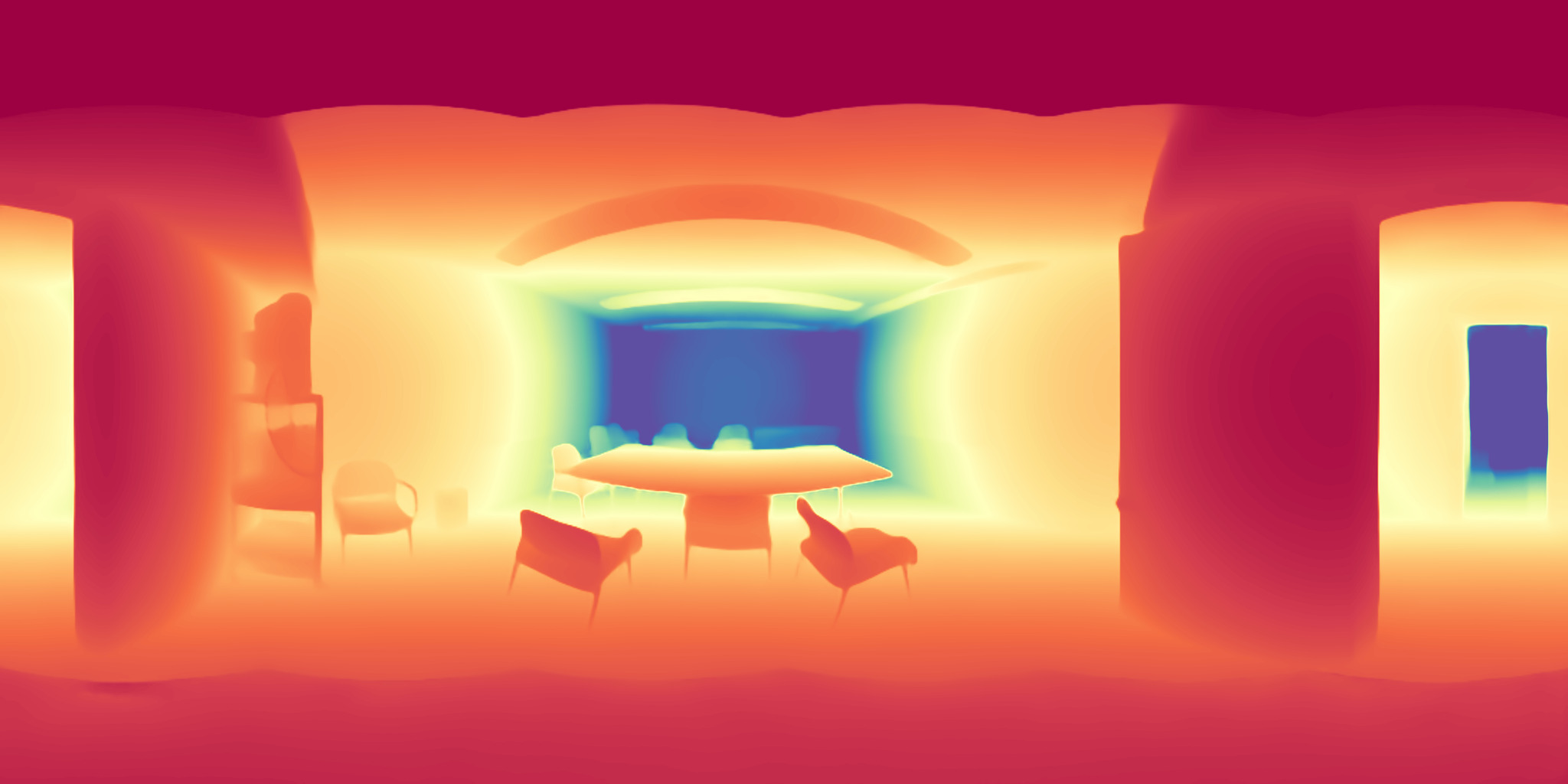}};
\node at (4*2-0.7,-2) {\includegraphics[width=1.95cm]{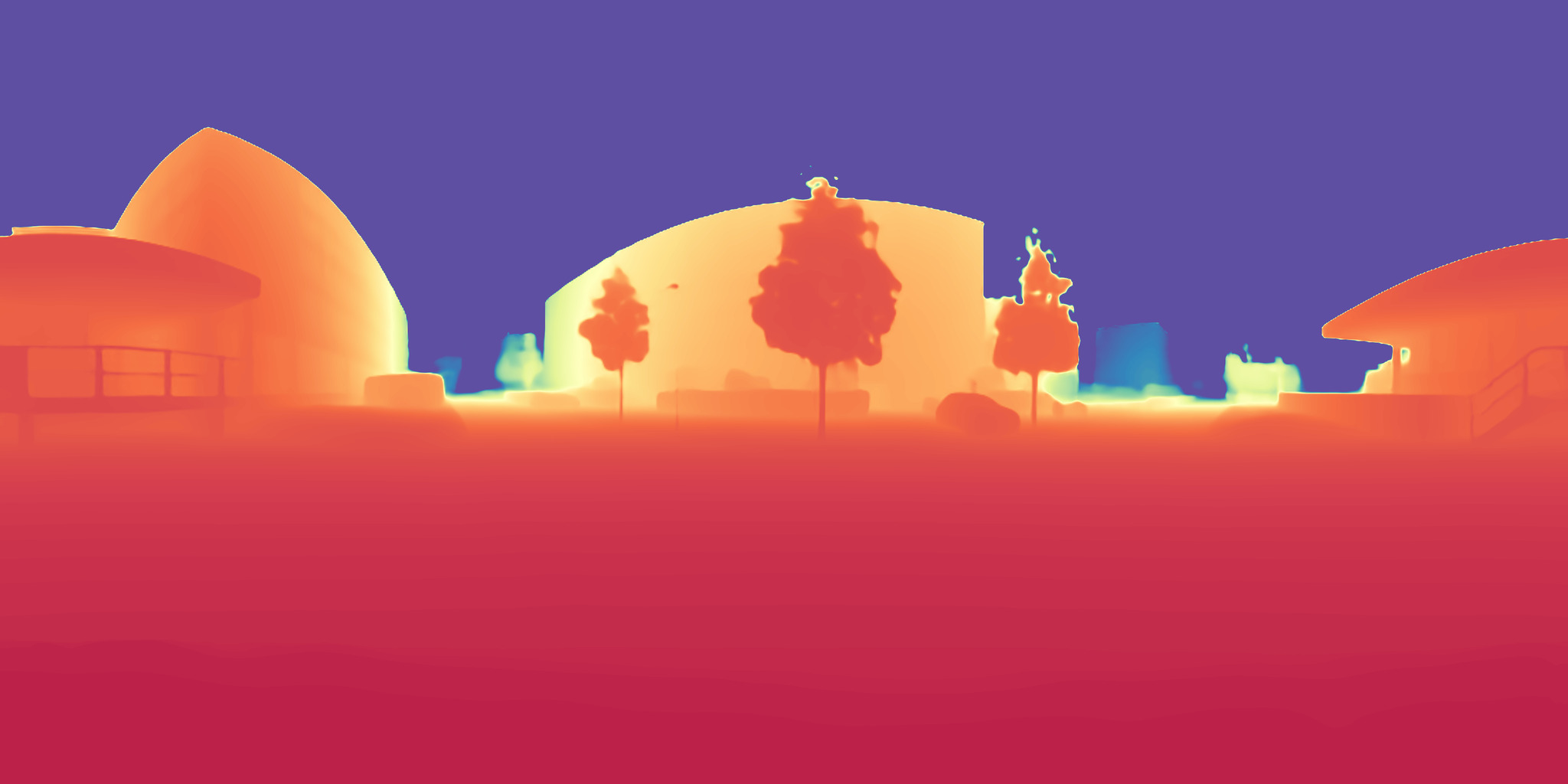}};
\node[rotate=90] at (-2.3,-2)
    {\scriptsize UniK3D};

\node at (4*0-0.7,-3) {\includegraphics[width=1.95cm]{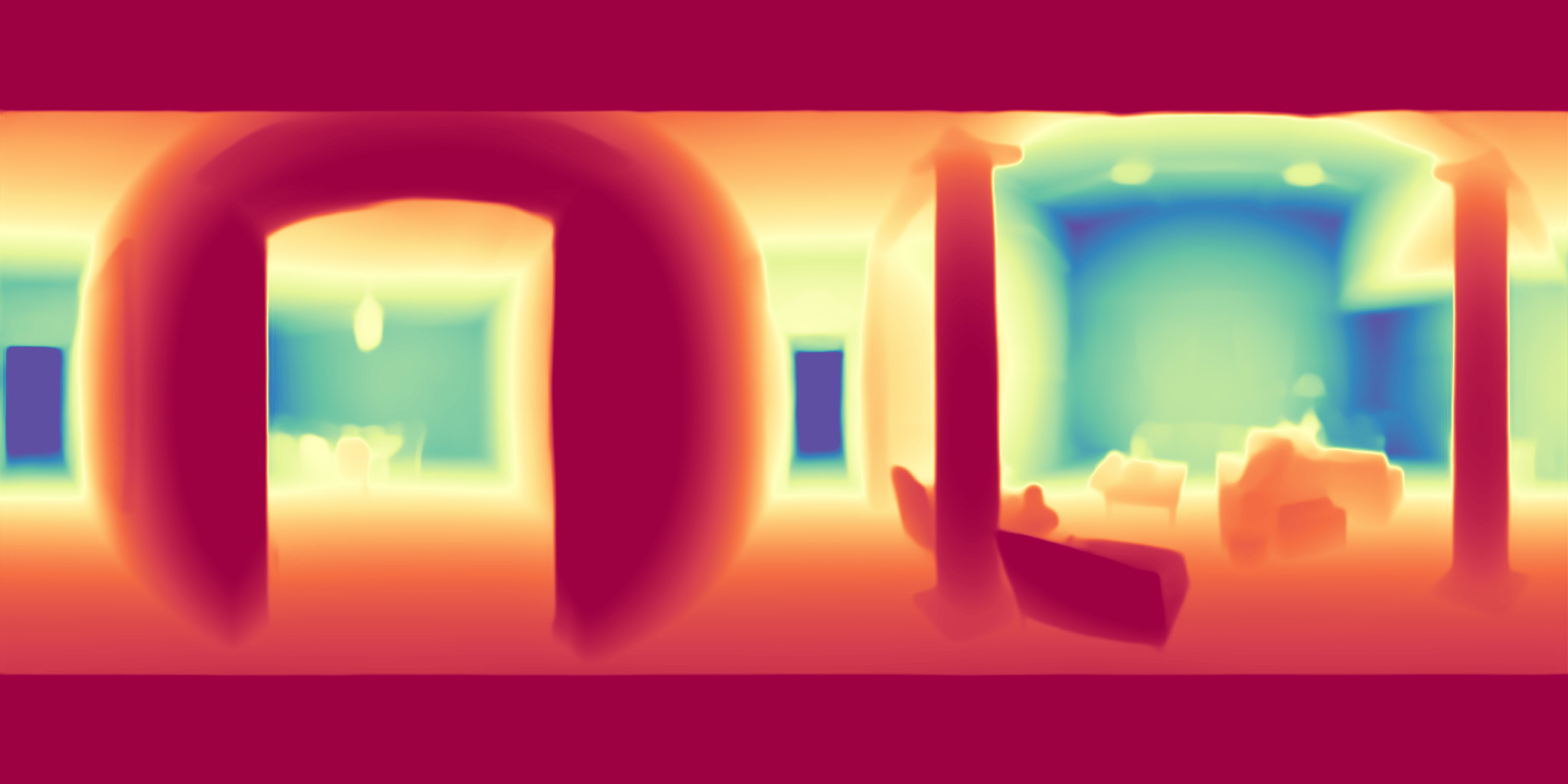}};
\node at (4*1-0.7,-3) {\includegraphics[width=1.95cm]{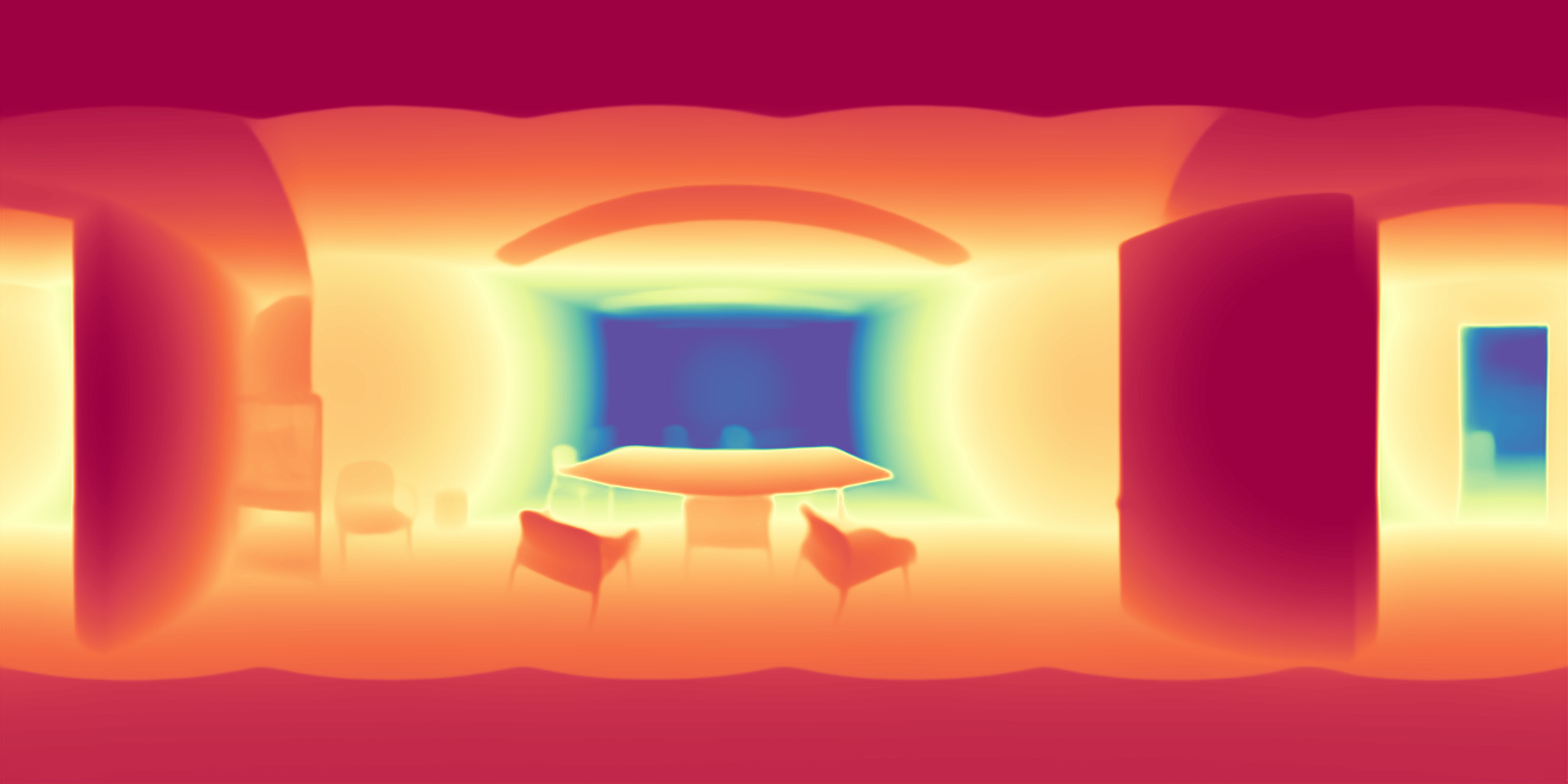}};
\node at (4*2-0.7,-3) {\includegraphics[width=1.95cm]{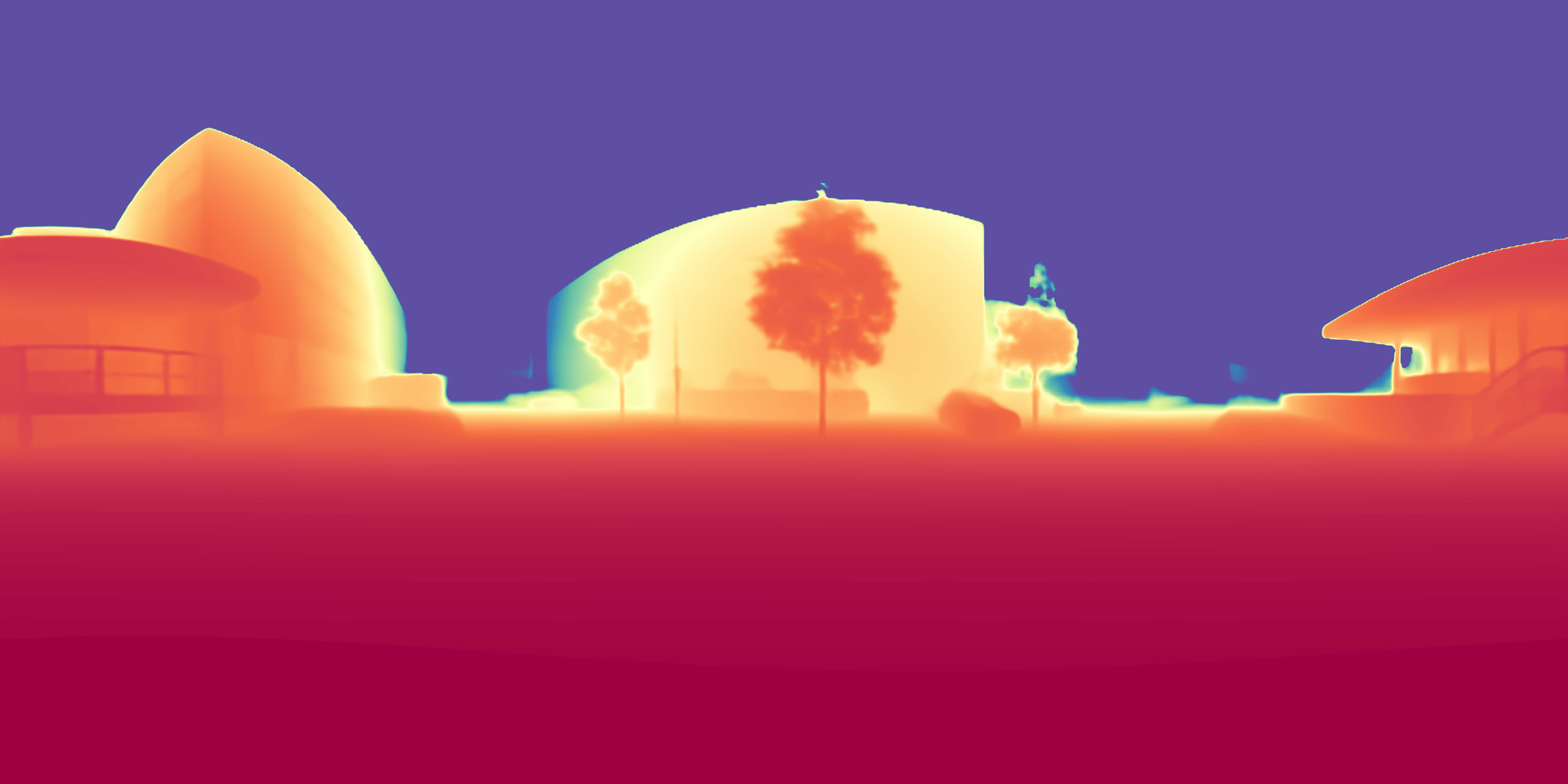}};
\node[rotate=90] at (-2.3,-3)
    {\scriptsize DAP};

\node at (4*0-0.7,-4) {\includegraphics[width=1.95cm]{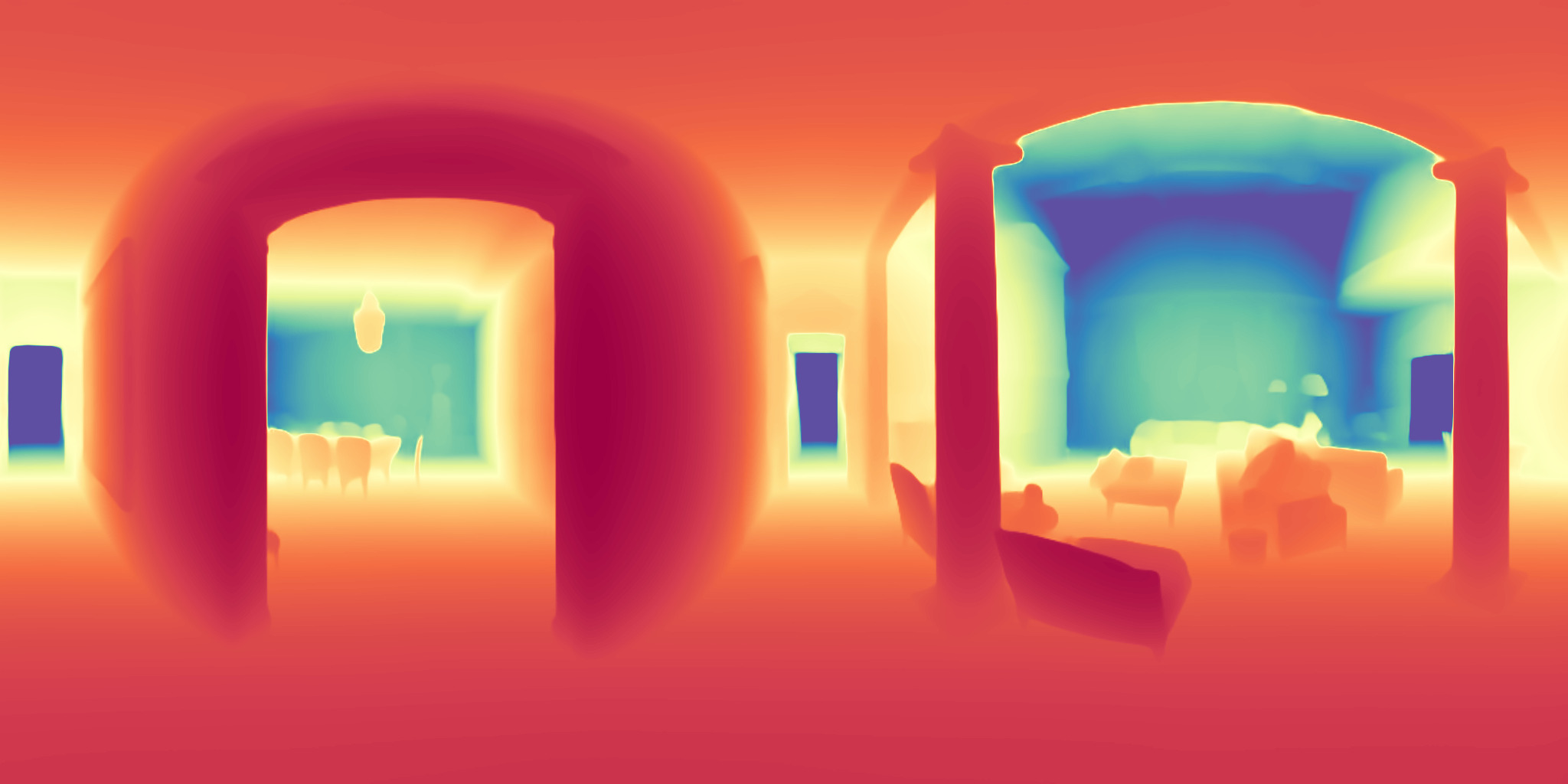}};
\node at (4*1-0.7,-4) {\includegraphics[width=1.95cm]{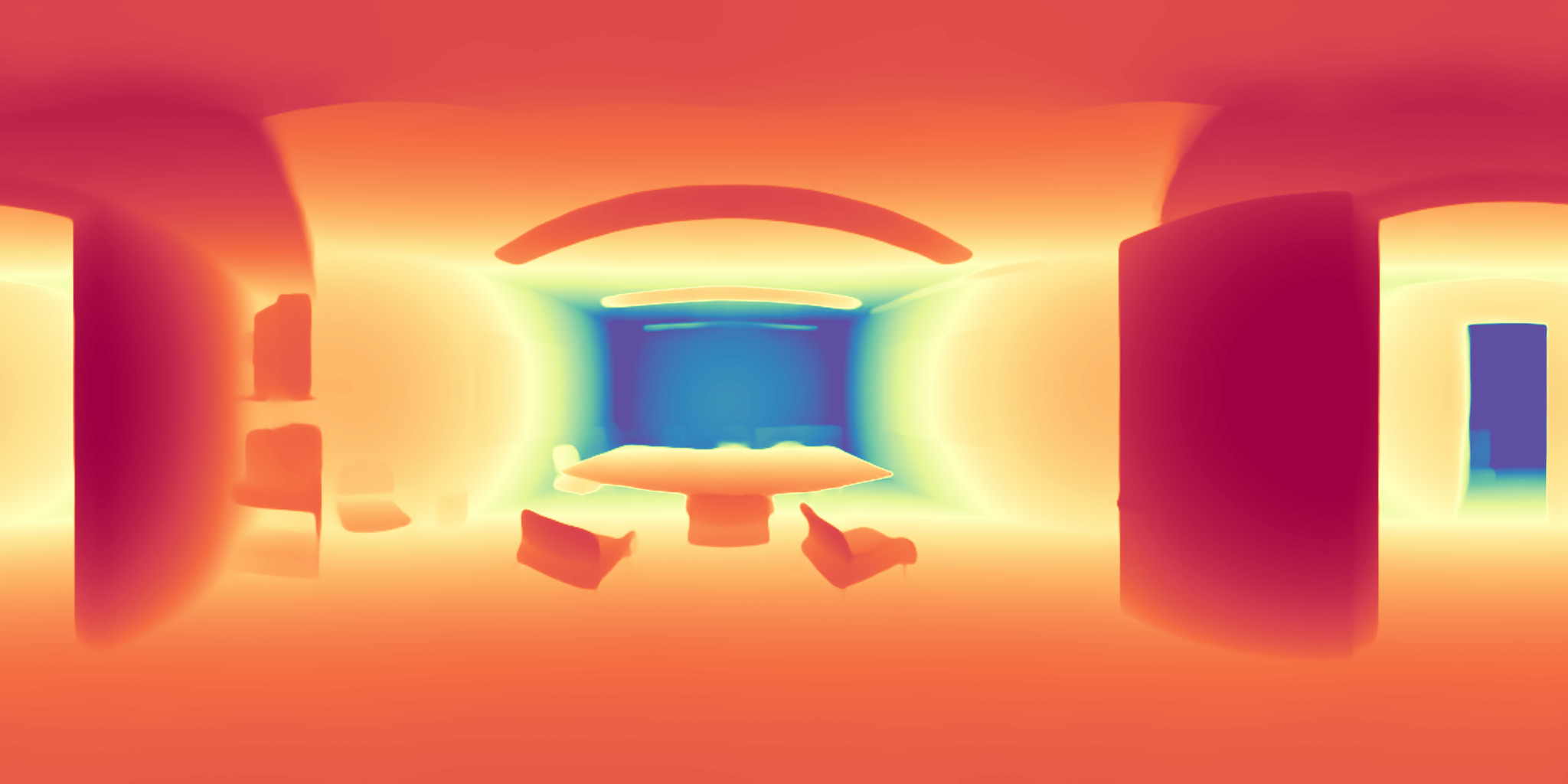}};
\node at (4*2-0.7,-4) {\includegraphics[width=1.95cm]{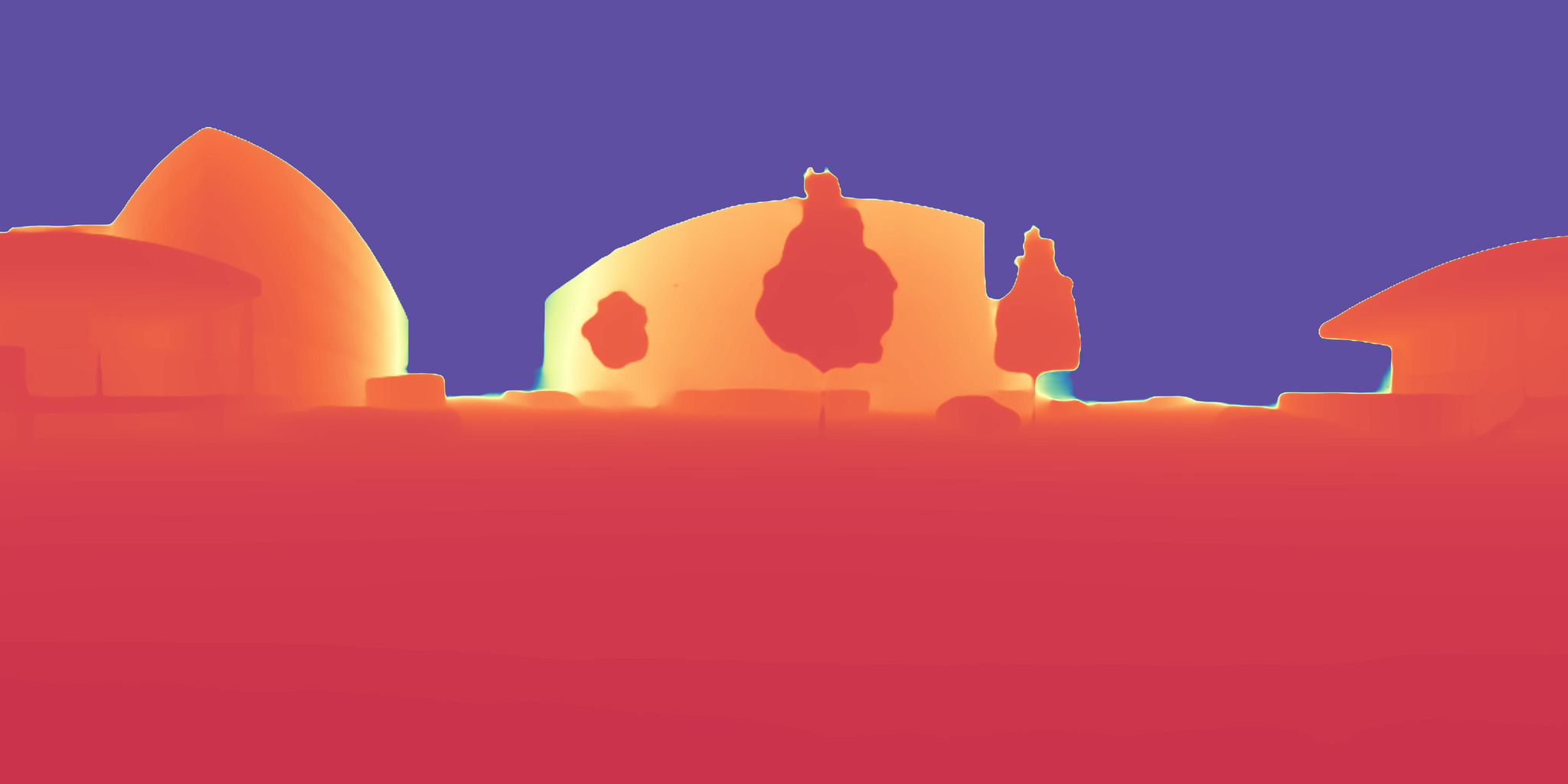}};
\node[rotate=90] at (-2.3,-4)
    {\scriptsize $\mathrm{DA}^2$};

\node at (4*0-0.7,-5) {\includegraphics[width=1.95cm]{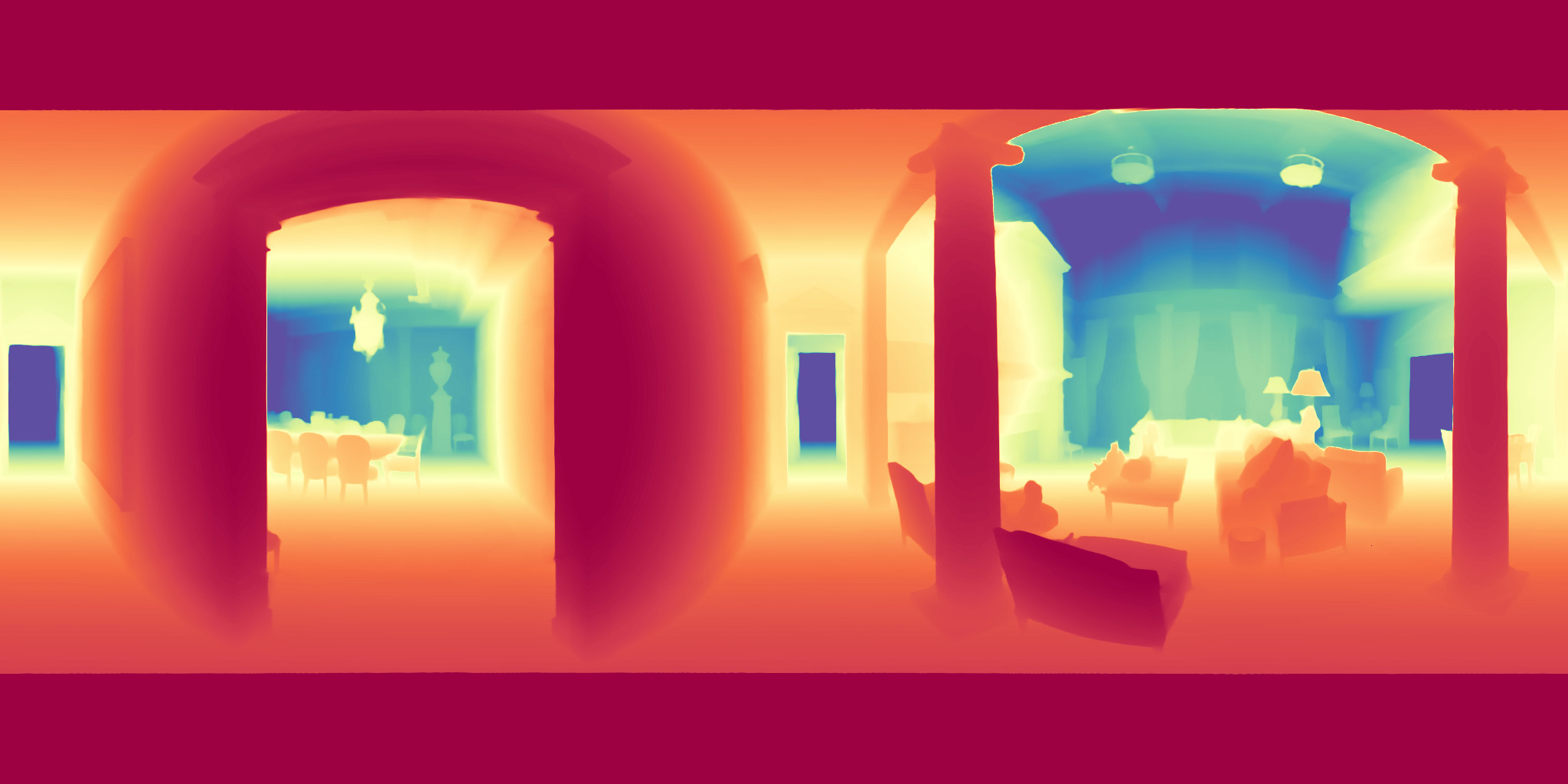}};
\node at (4*1-0.7,-5) {\includegraphics[width=1.95cm]{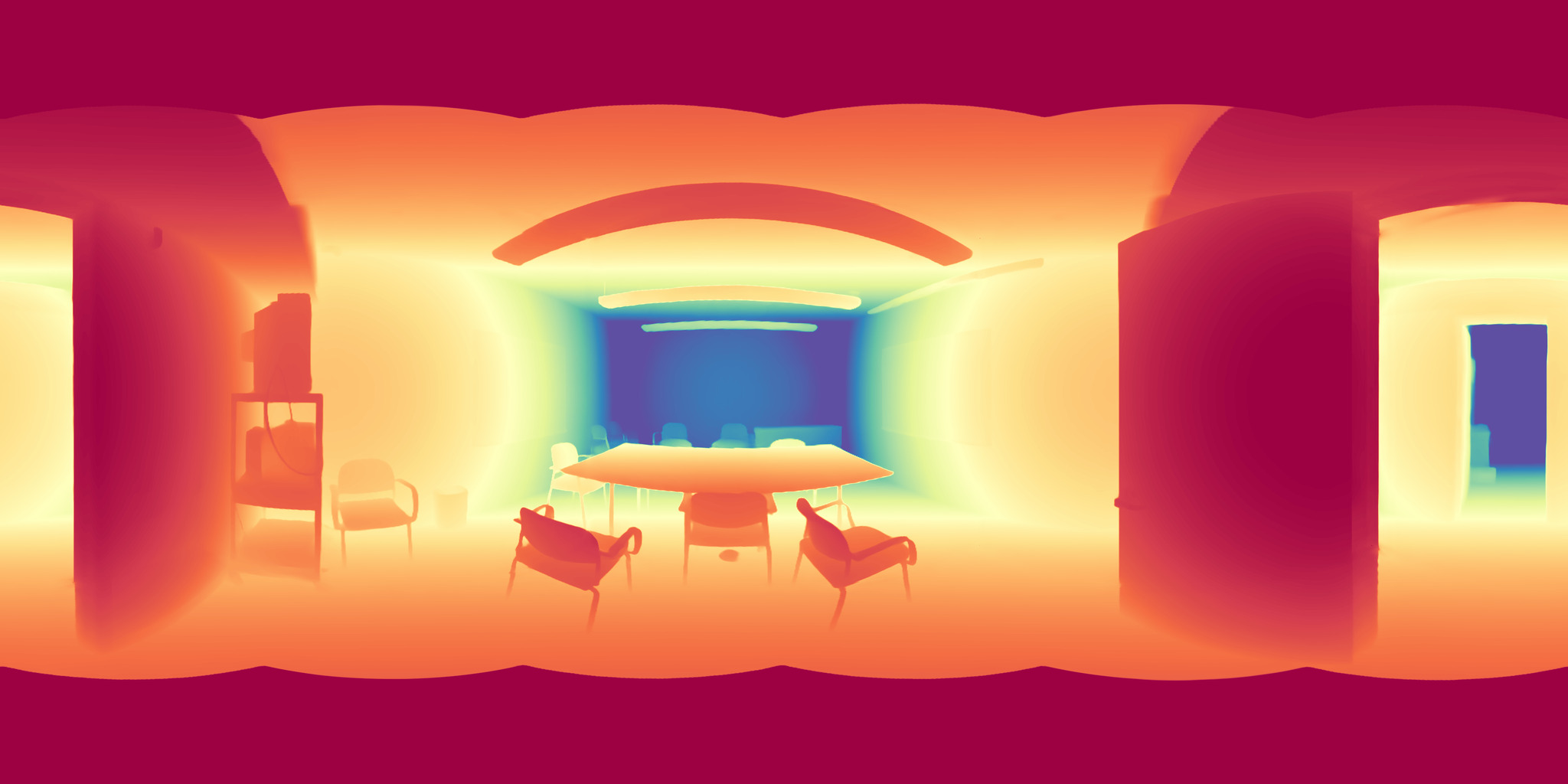}};
\node at (4*2-0.7,-5) {\includegraphics[width=1.95cm]{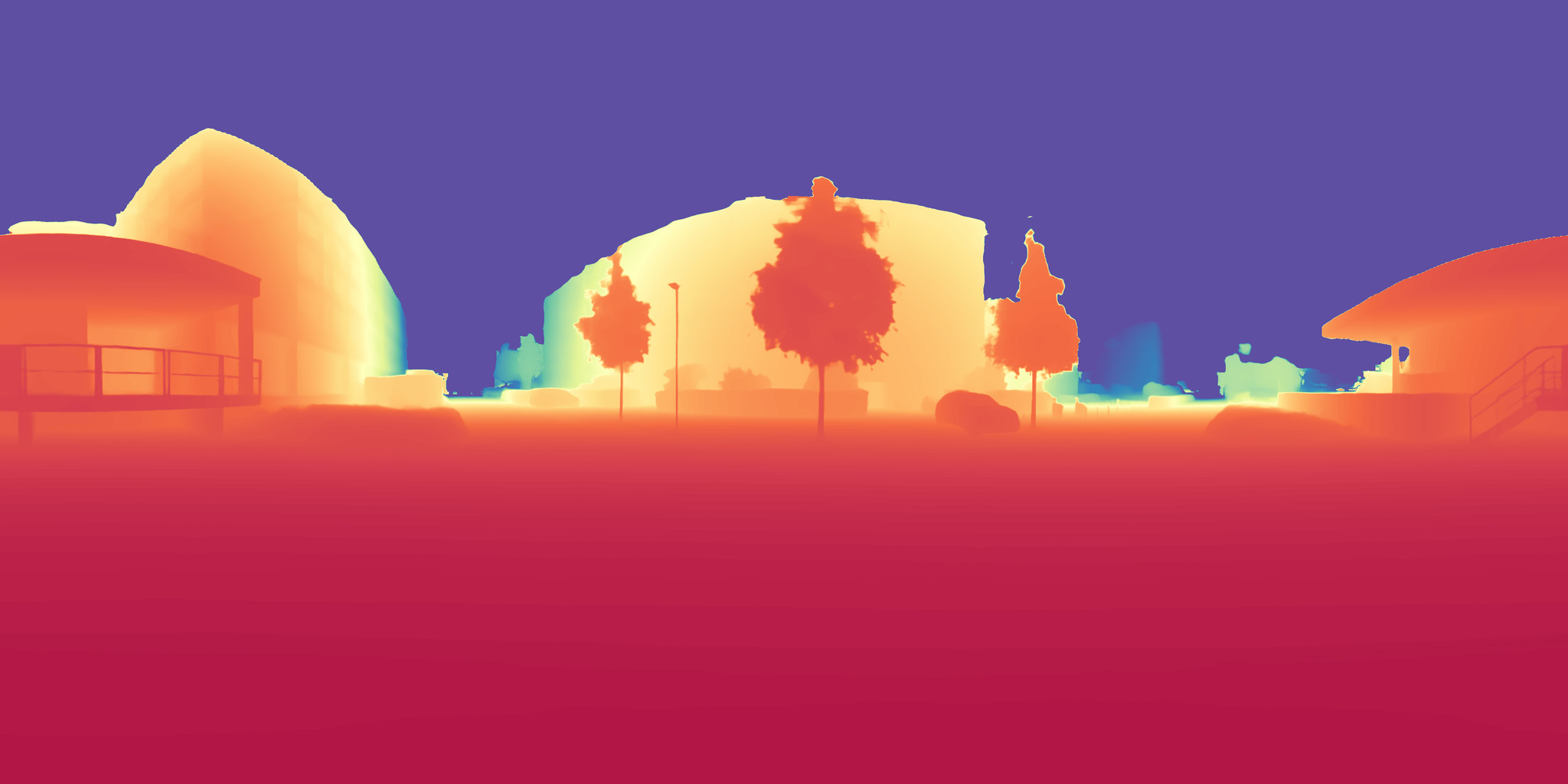}};
\node[rotate=90] at (-2.3,-5)
    {\scriptsize Ours};

    \node at (4*0-0.7,-6)
        {\includegraphics[width=1.95cm]{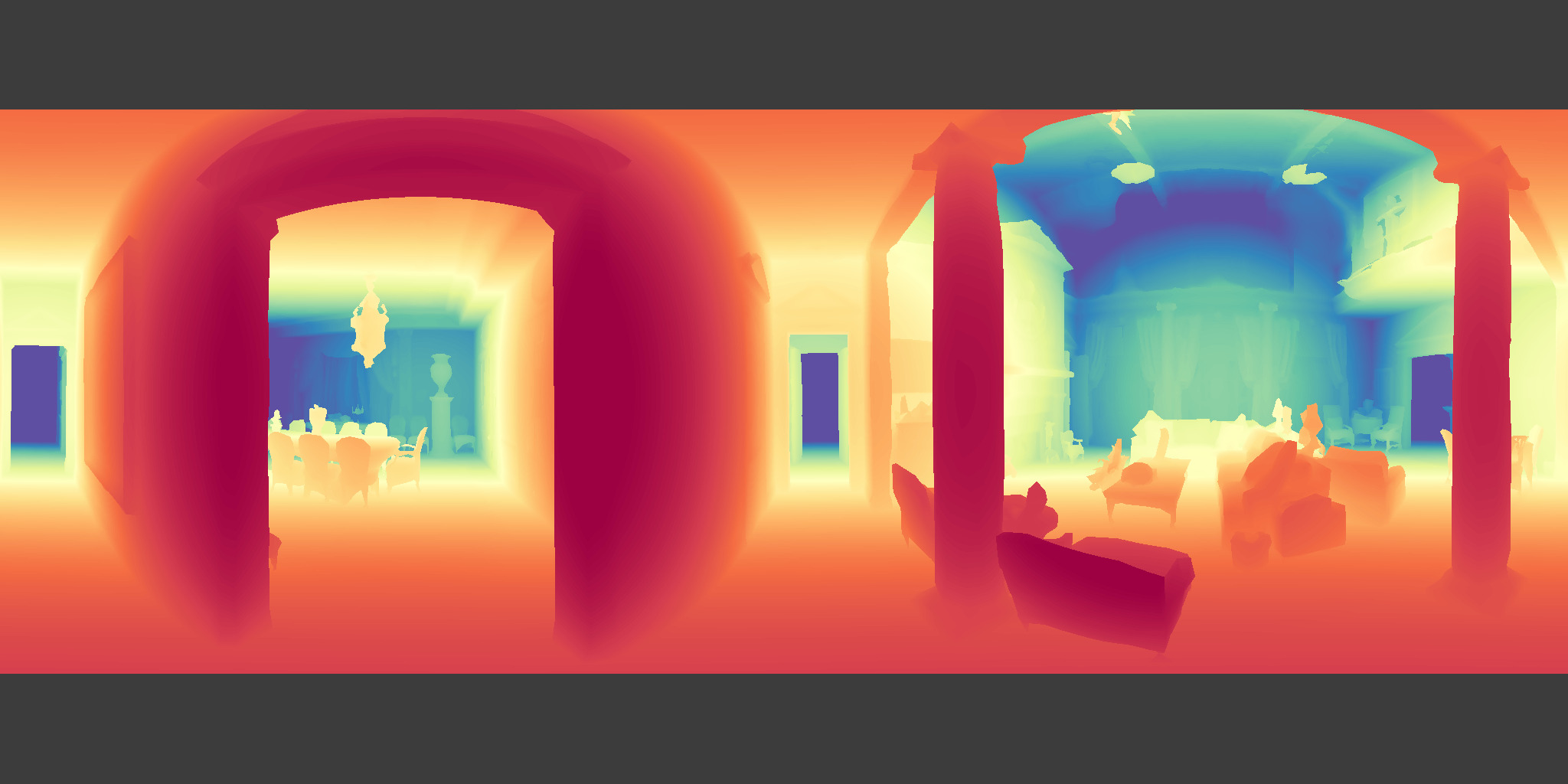}};
    \node at (4*1-0.7,-6)
        {\includegraphics[width=1.95cm]{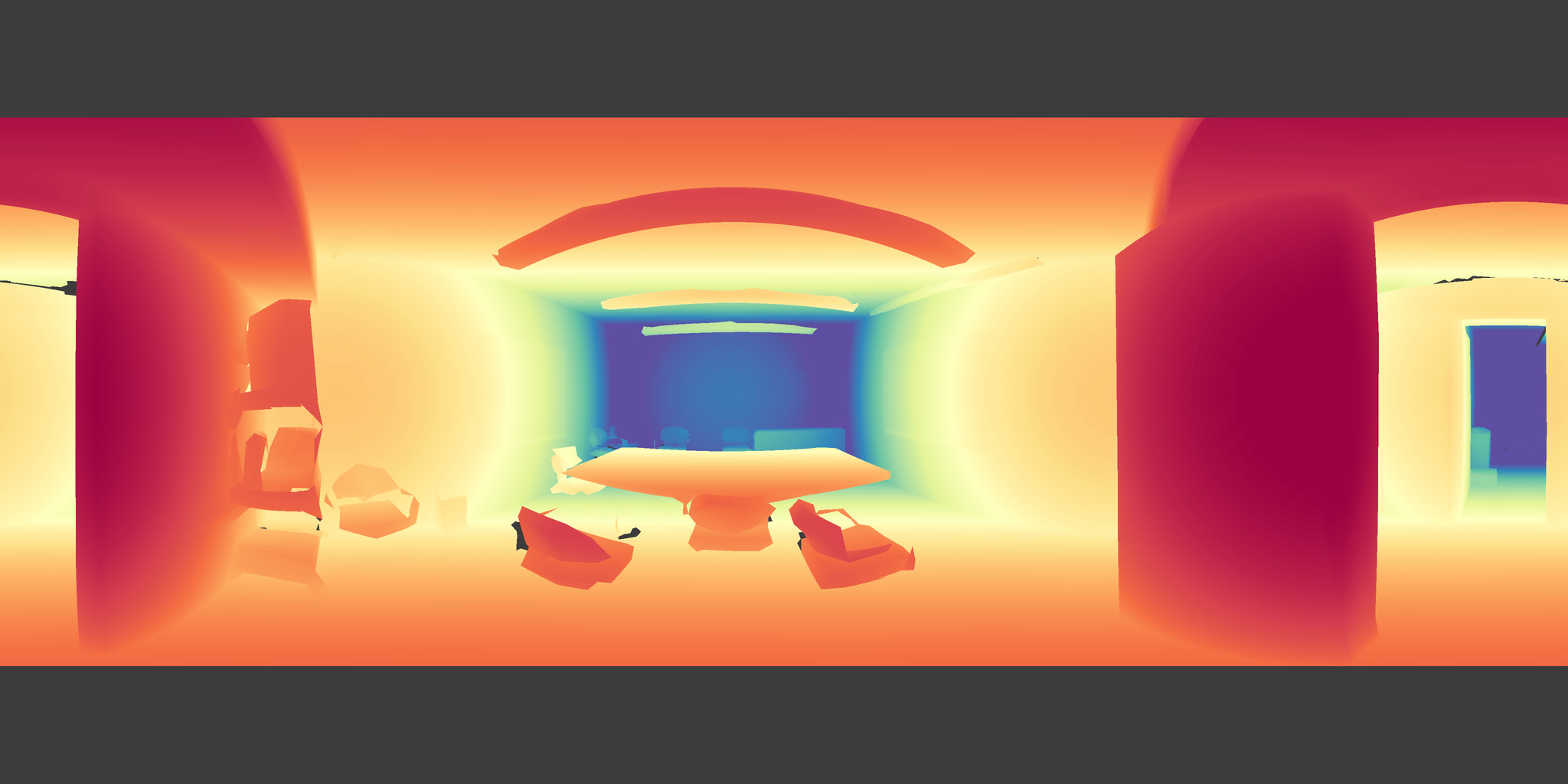}};
    \node at (4*2-0.7,-6)
        {\includegraphics[width=1.95cm]{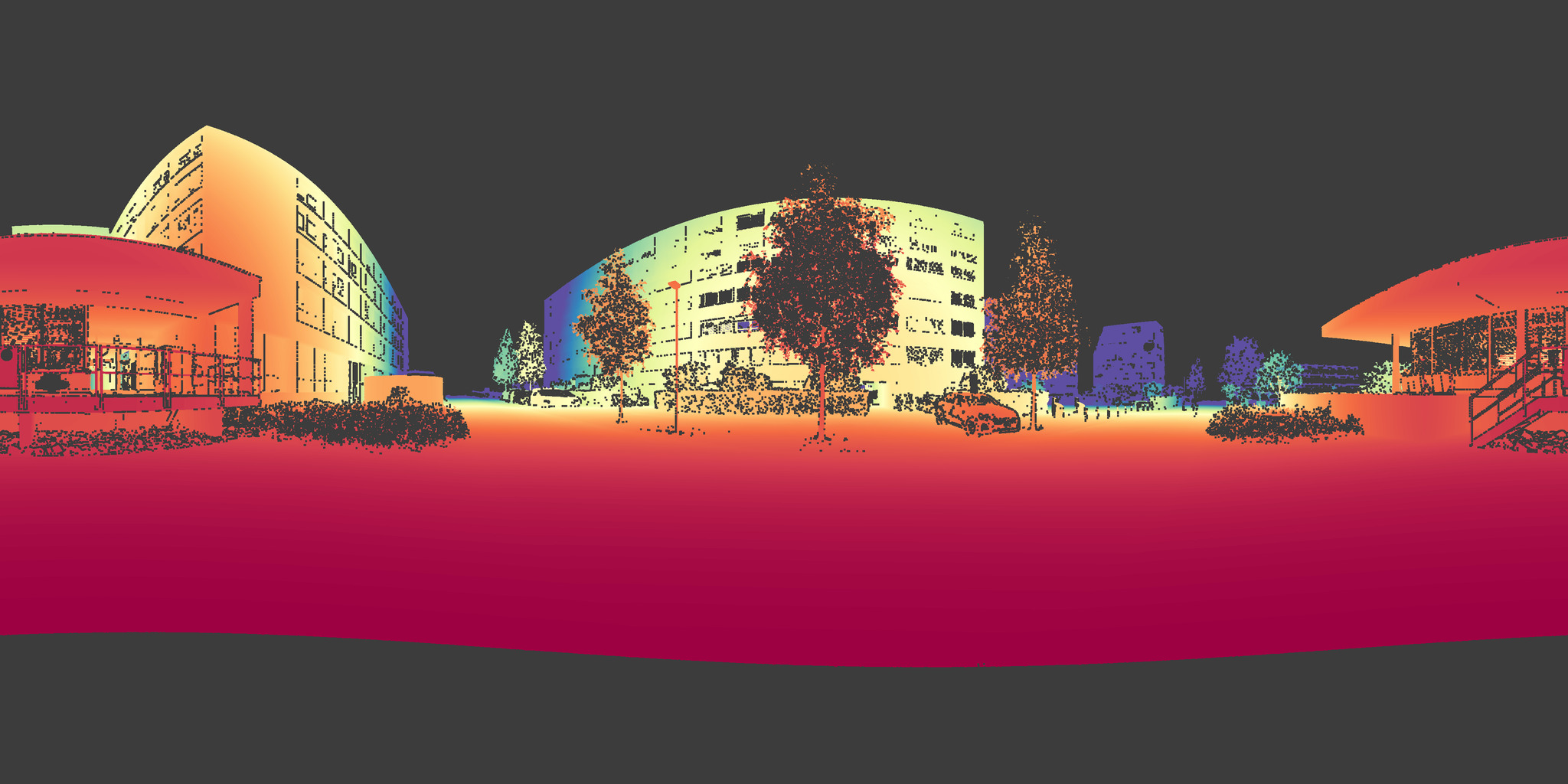}};
\node[rotate=90] at (-2.3,-6) {\scriptsize GT};

\spy on (-0.2,-0.1) in node at (1.32,0);
\spy on (2.9,-0.15) in node at (5.32,0);
\spy on (7.3,0) in node at (9.32,0);

\spy on (-0.2,-1.1) in node at (1.32,-1);
\spy on (2.9,-1.15) in node at (5.32,-1);
\spy on (7.3,-1) in node at (9.32,-1);

\spy on (-0.2,-2.1) in node at (1.32,-2);
\spy on (2.9,-2.15) in node at (5.32,-2);
\spy on (7.3,-2) in node at (9.32,-2);

\spy on (-0.2,-3.1) in node at (1.32,-3);
\spy on (2.9,-3.15) in node at (5.32,-3);
\spy on (7.3,-3) in node at (9.32,-3);

\spy on (-0.2,-4.1) in node at (1.32,-4);
\spy on (2.9,-4.15) in node at (5.32,-4);
\spy on (7.3,-4) in node at (9.32,-4);

\spy on (-0.2,-5.1) in node at (1.32,-5);
\spy on (2.9,-5.15) in node at (5.32,-5);
\spy on (7.3,-5) in node at (9.32,-5);

\spy on (-0.2,-6.1) in node at (1.32,-6);
\spy on (2.9,-6.15) in node at (5.32,-6);
\spy on (7.3,-6) in node at (9.32,-6);

\end{tikzpicture}
}
\vspace{-.5cm}
\caption{\textbf{Qualitative comparison of panoramic depth estimation.}. Visual results from \name{}, compared to the subset of the best evaluated baselines, shown alongside the RGB input and ground-truth depth on Matterport3D360, Stanford2D3DS, and ZüriPano panoramas. (Best viewed zoomed in.)}
\label{fig:comparison_supp}
\end{figure}

\begin{figure}[!h]
    \centering
        \begin{minipage}{0.2\linewidth}
            \centering 
            Input
        \end{minipage}%
        \begin{minipage}{0.4\linewidth}
            \centering 
            DA3
        \end{minipage}%
        \begin{minipage}{0.4\linewidth}
            \centering 
            Ours
        \end{minipage}%
        \hfill
        \begin{minipage}{0.198\linewidth}
            \centering 
            \includegraphics[width=\linewidth]{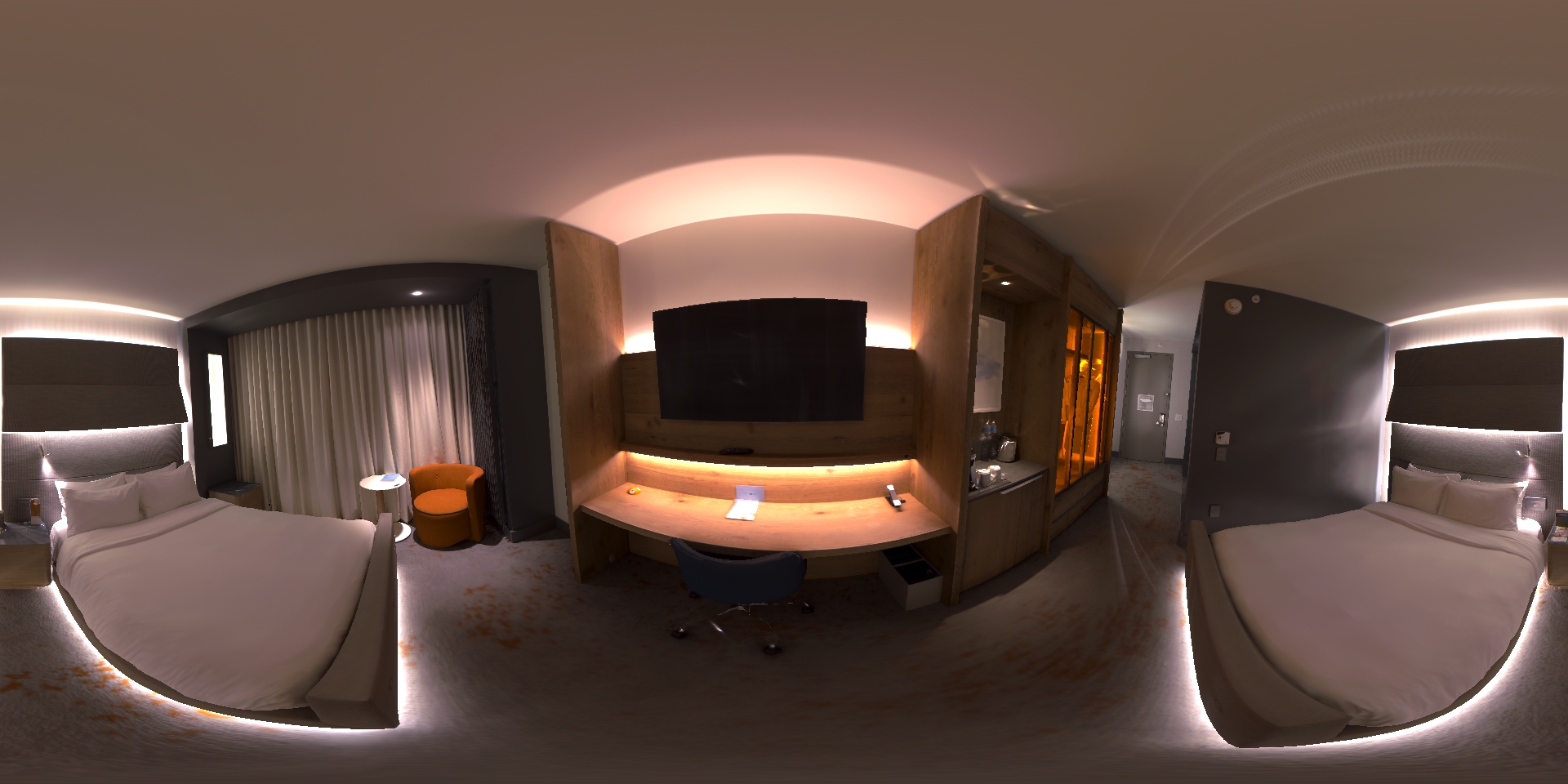}
        \end{minipage}%
        \hfill
        \begin{minipage}{0.198\linewidth}
            \centering 
            \includegraphics[width=\linewidth]{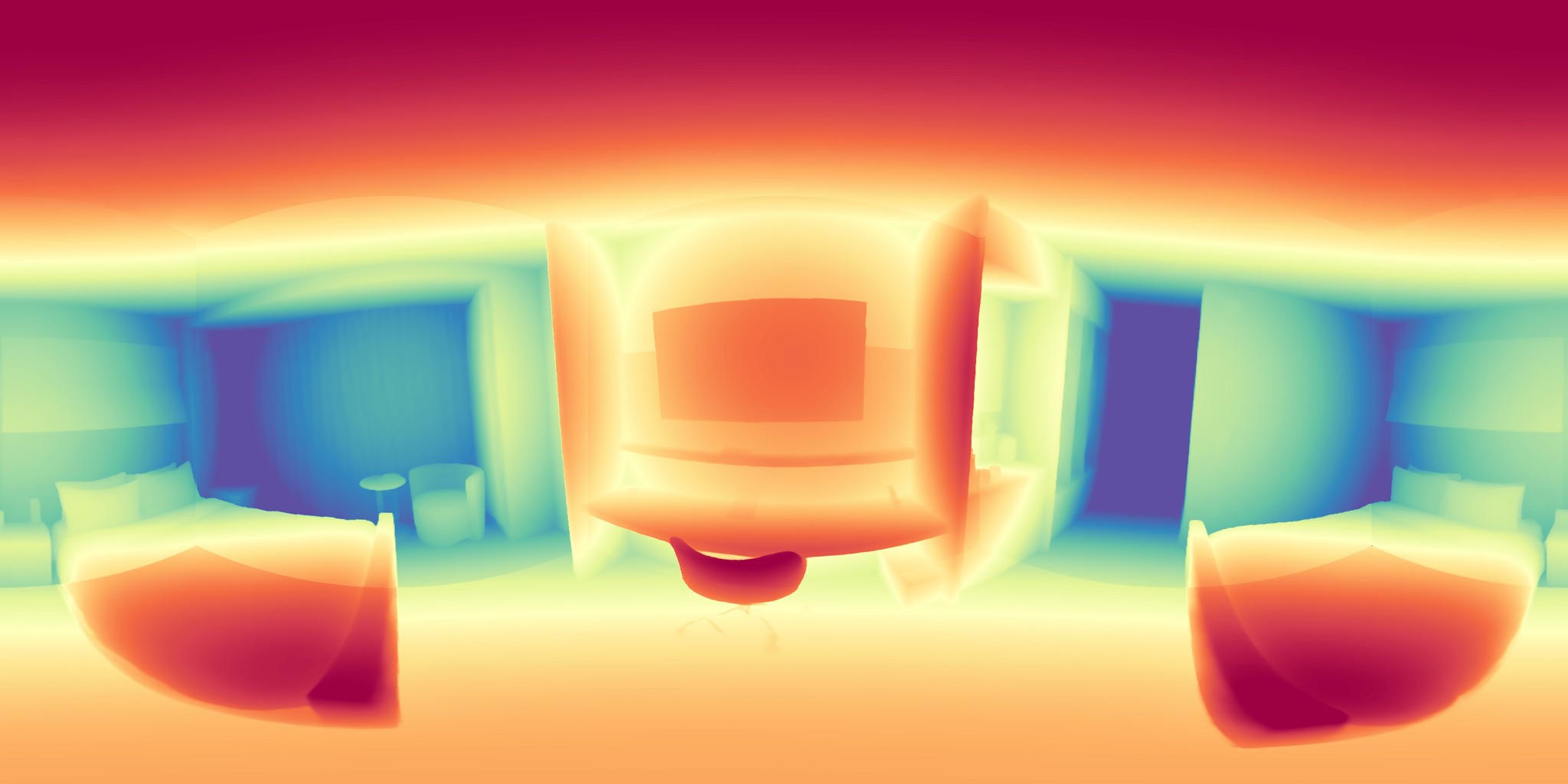}
        \end{minipage}%
        \begin{minipage}{0.198\linewidth}
            \centering 
            \includegraphics[width=\linewidth]{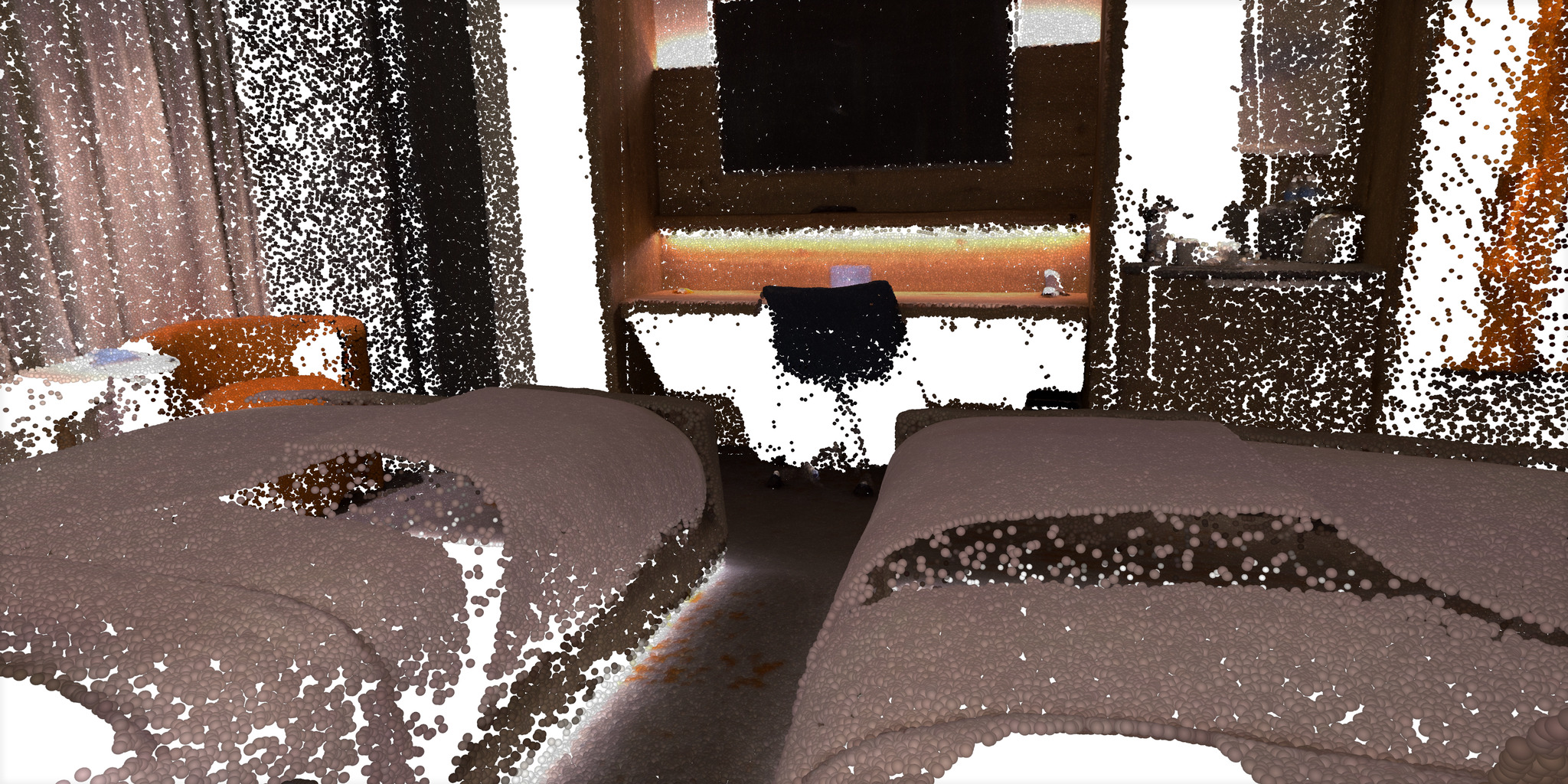}
        \end{minipage}%
        \hfill
        \begin{minipage}{0.198\linewidth}
            \centering 
            \includegraphics[width=\linewidth]{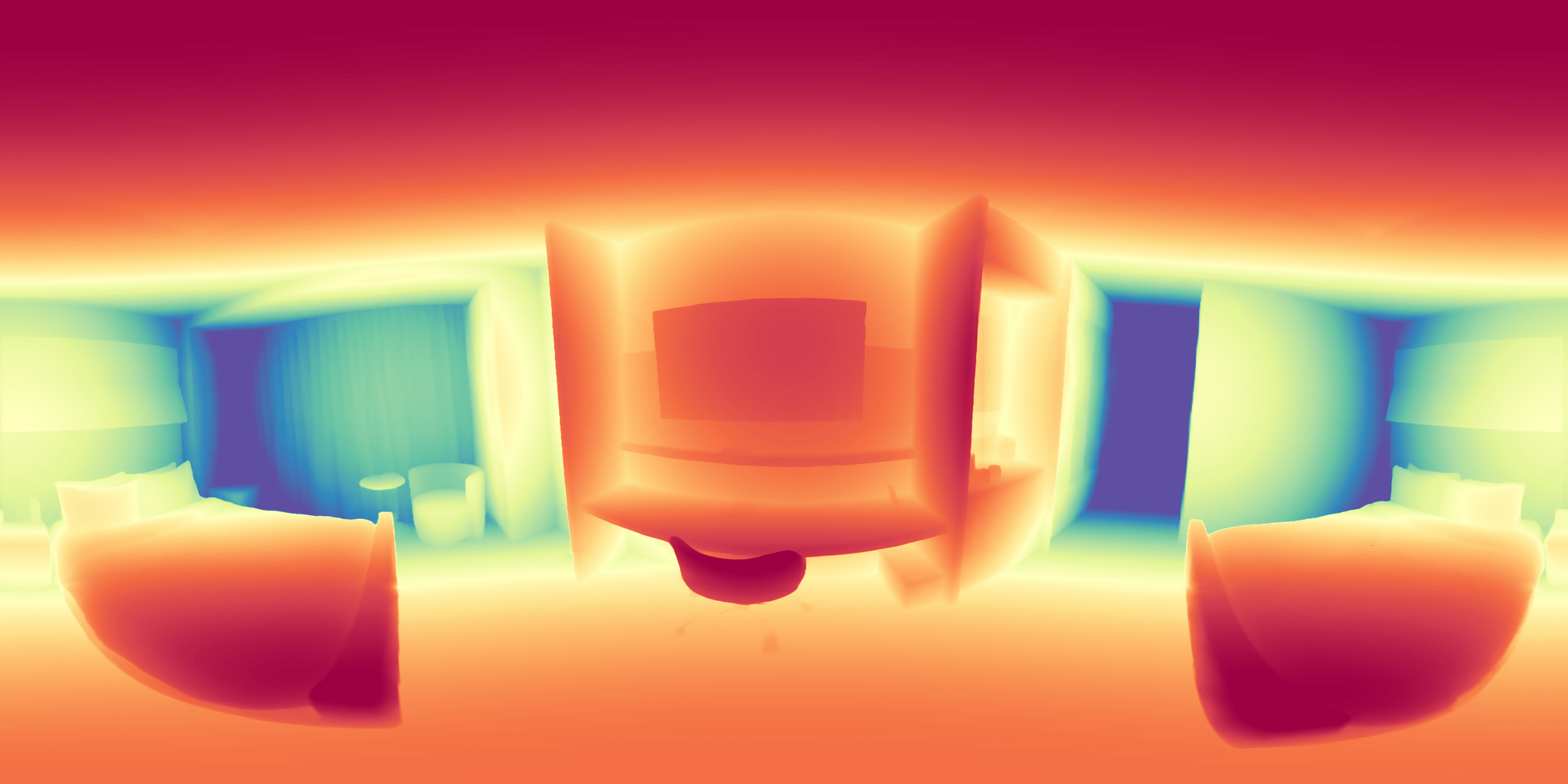}
        \end{minipage}%
        \begin{minipage}{0.198\linewidth}
            \centering 
            \includegraphics[width=\linewidth]{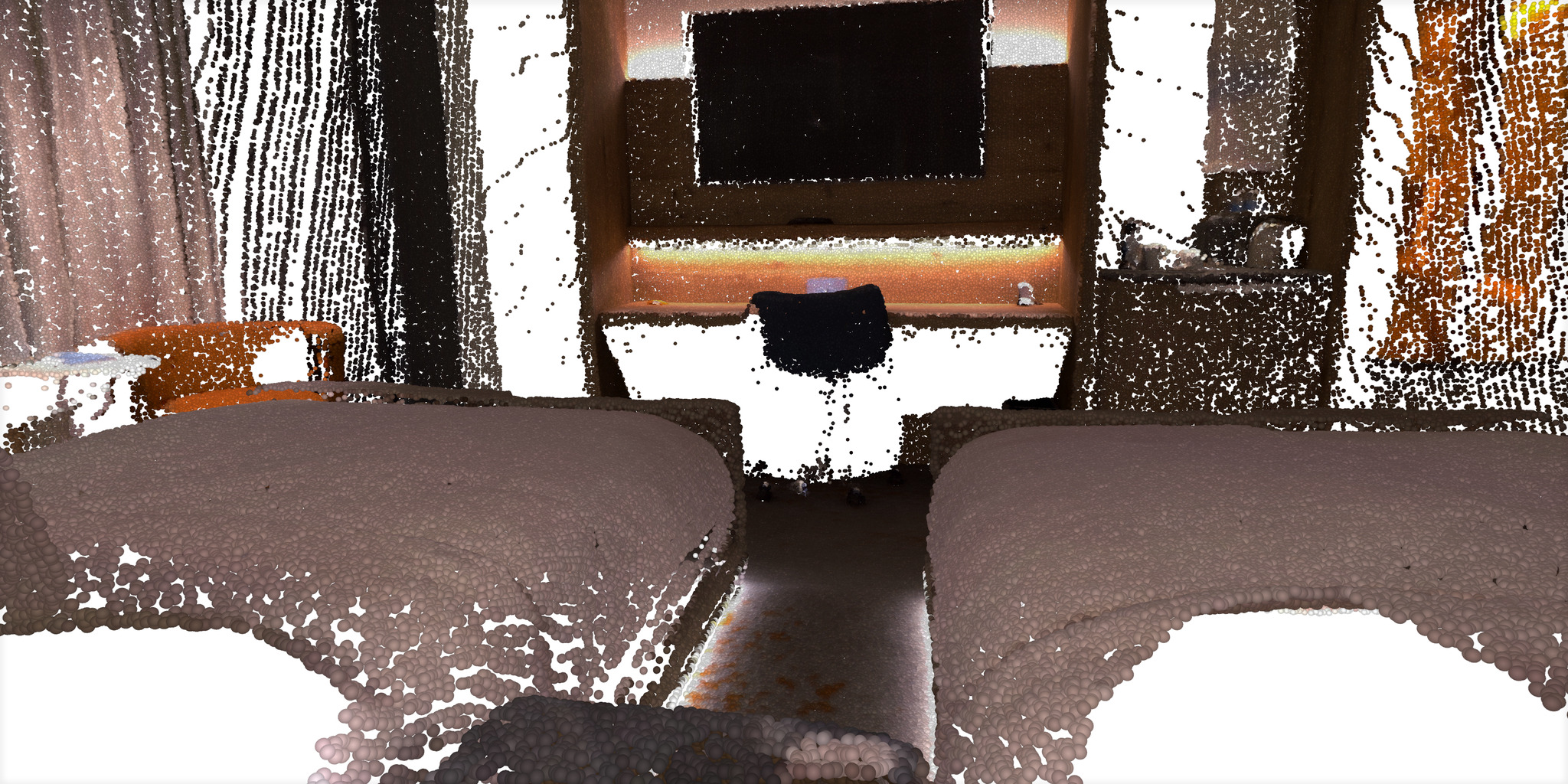}
        \end{minipage}%
        \hfill
        \begin{minipage}{0.198\linewidth}
            \centering 
            \includegraphics[width=\linewidth]{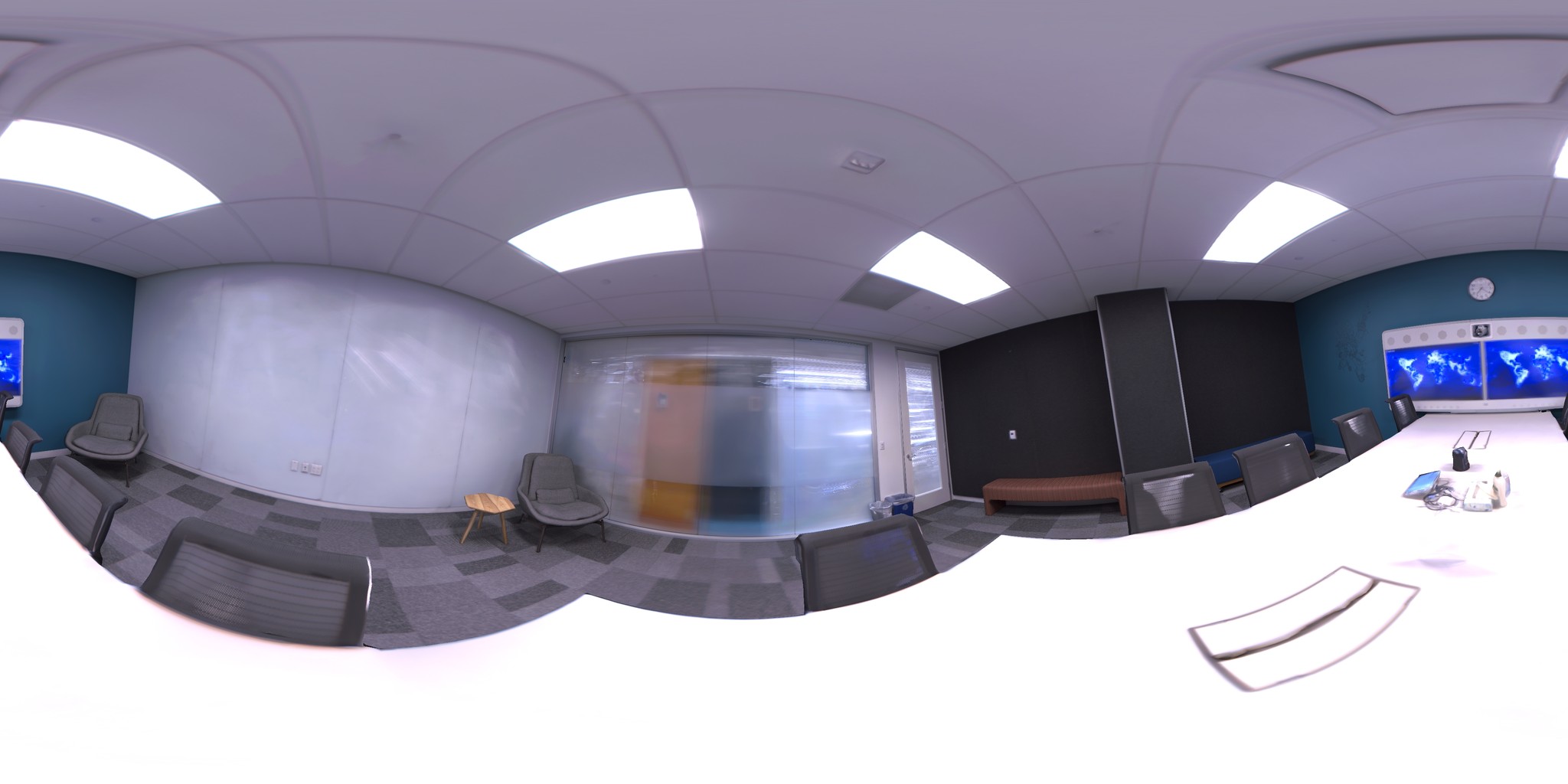}
        \end{minipage}%
        \hfill
        \begin{minipage}{0.198\linewidth}
            \centering 
            \includegraphics[width=\linewidth]{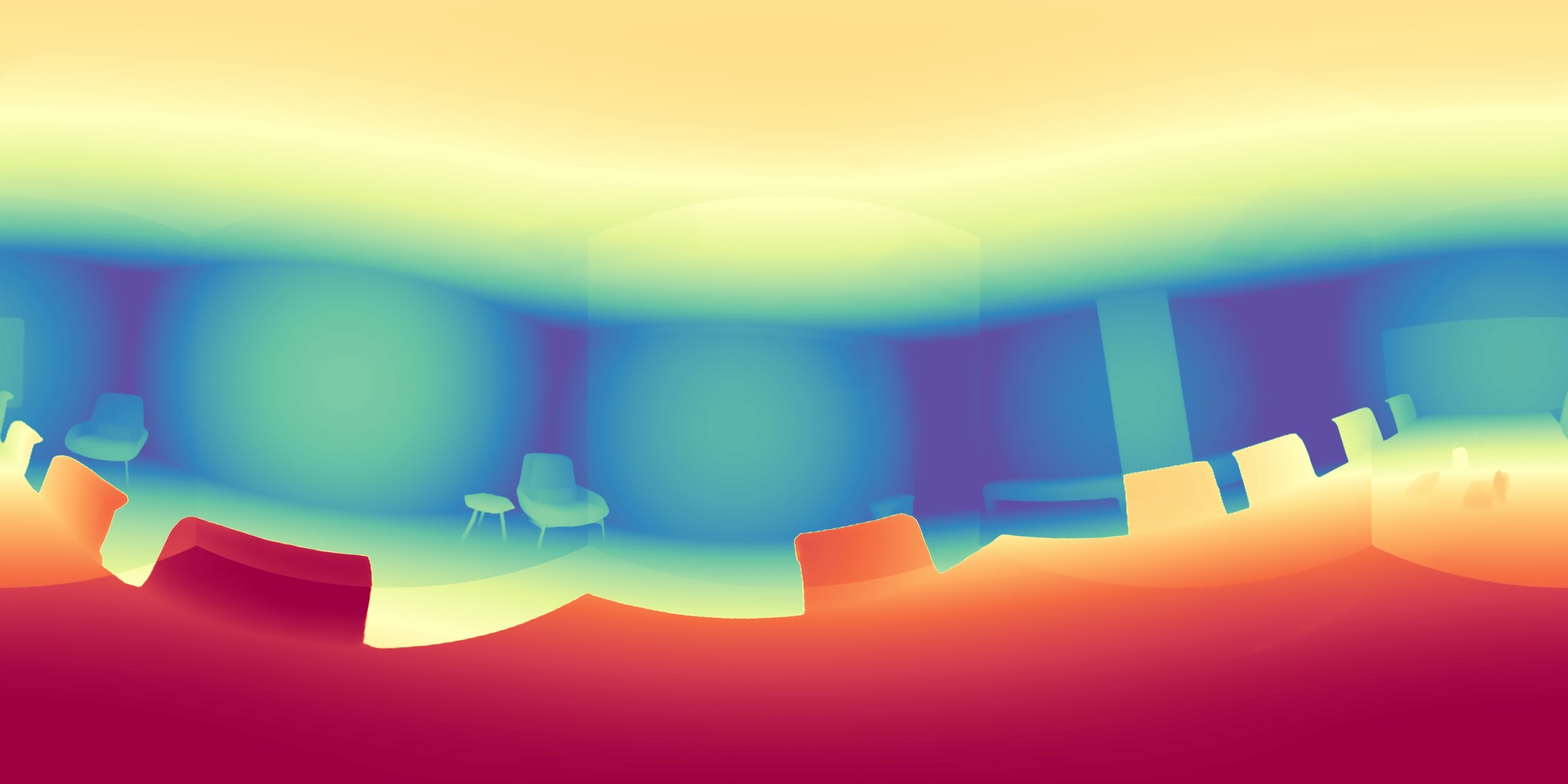}
        \end{minipage}%
        \begin{minipage}{0.198\linewidth}
            \centering 
            \includegraphics[width=\linewidth]{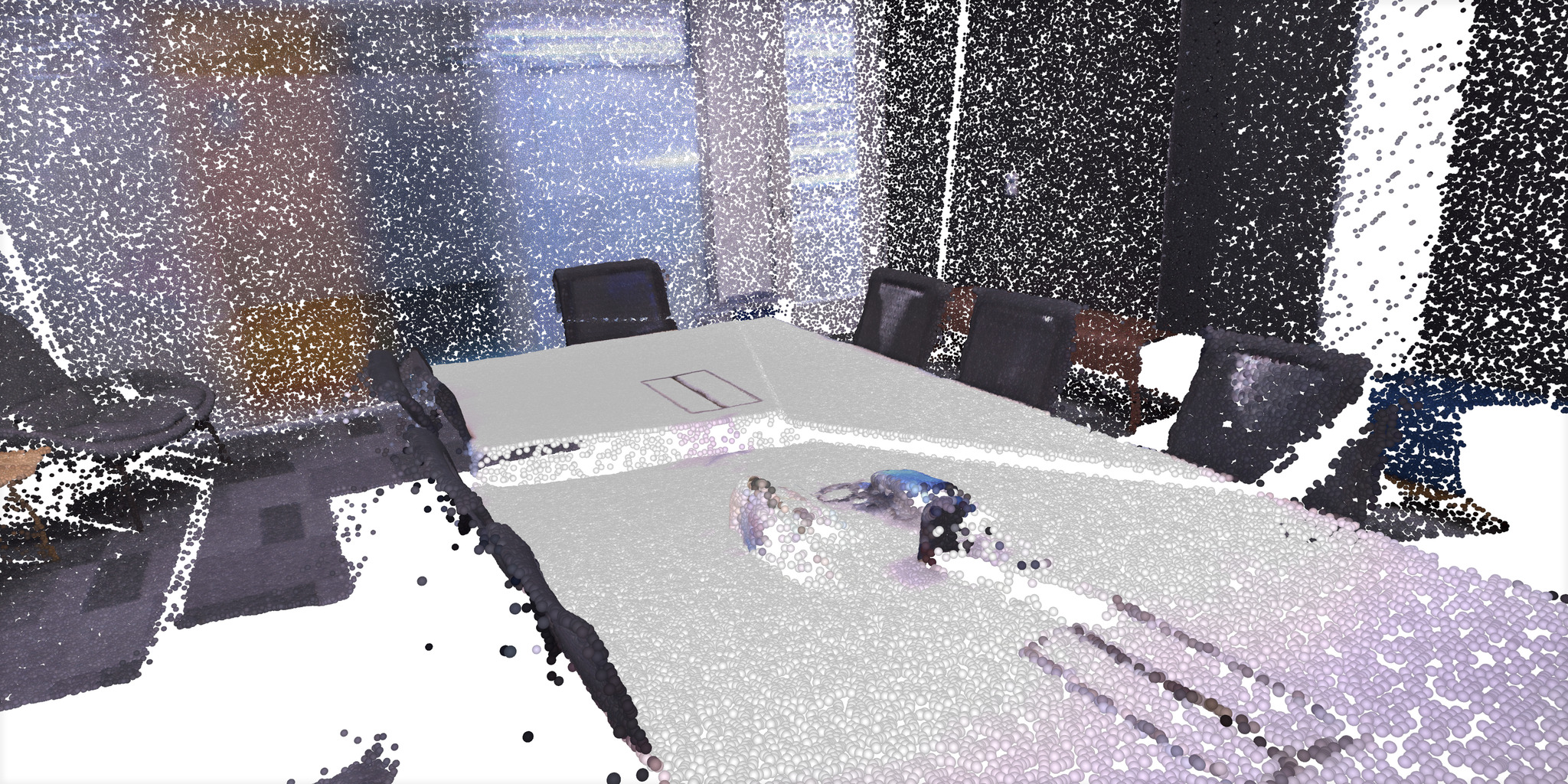}
        \end{minipage}%
        \hfill
        \begin{minipage}{0.198\linewidth}
            \centering 
            \includegraphics[width=\linewidth]{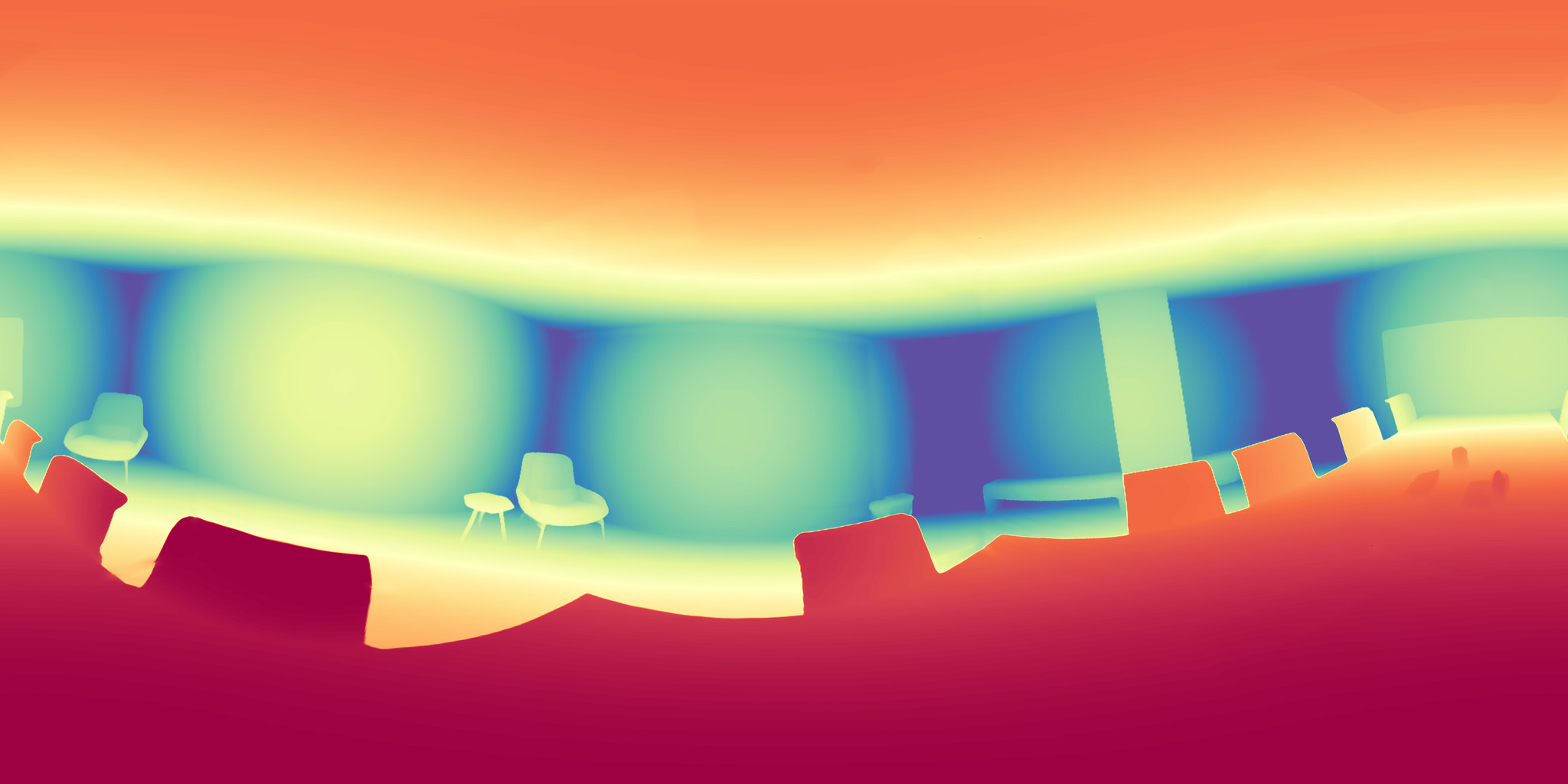}
        \end{minipage}%
        \begin{minipage}{0.198\linewidth}
            \centering 
            \includegraphics[width=\linewidth]{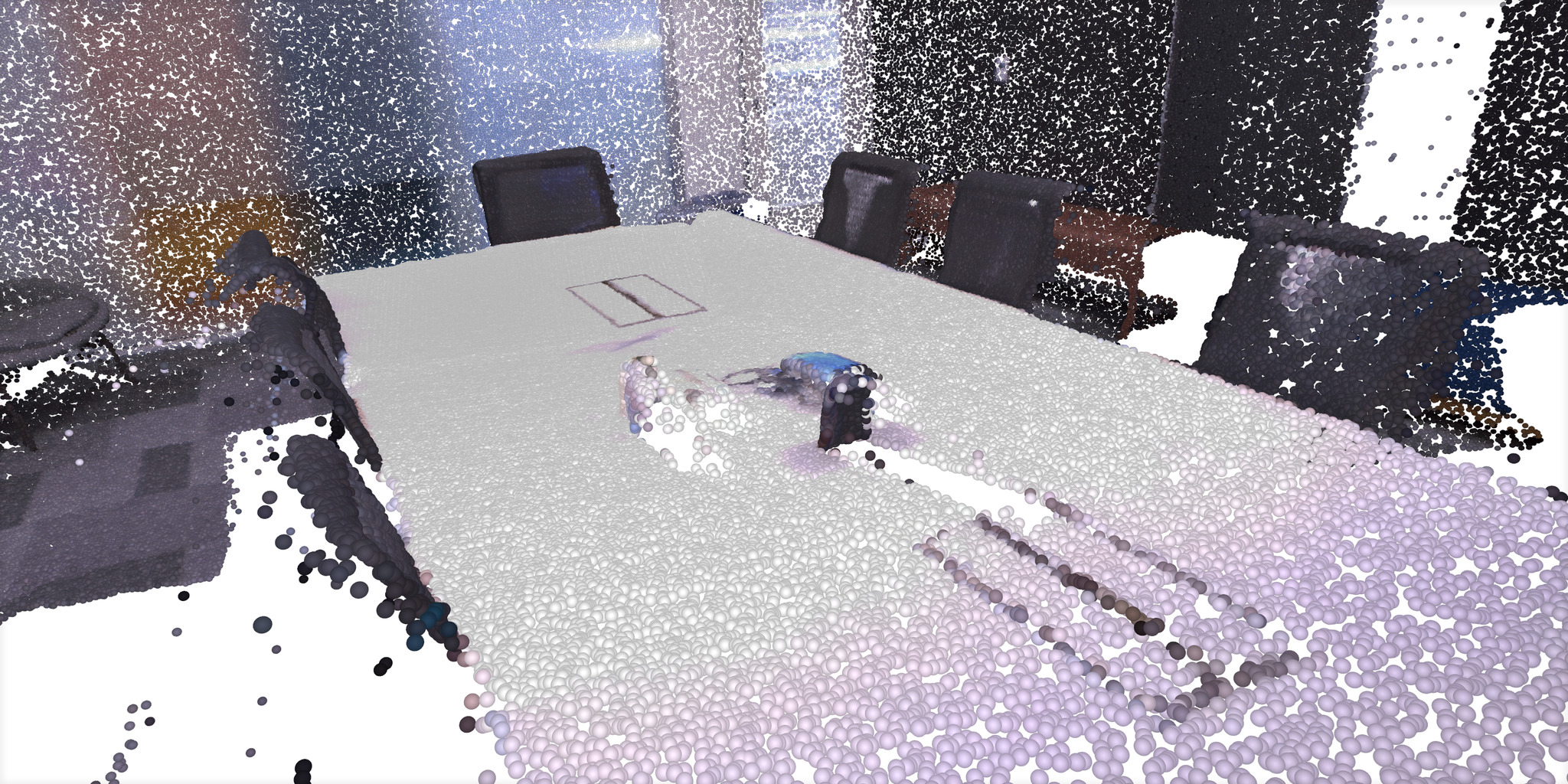}
        \end{minipage}%
        \hfill
    \caption{Comparison to vanilla DA3.}
    \label{fig:da3-comparison}
\end{figure}
\begin{figure}[!t]
    \centering
    \includegraphics[width=\linewidth]{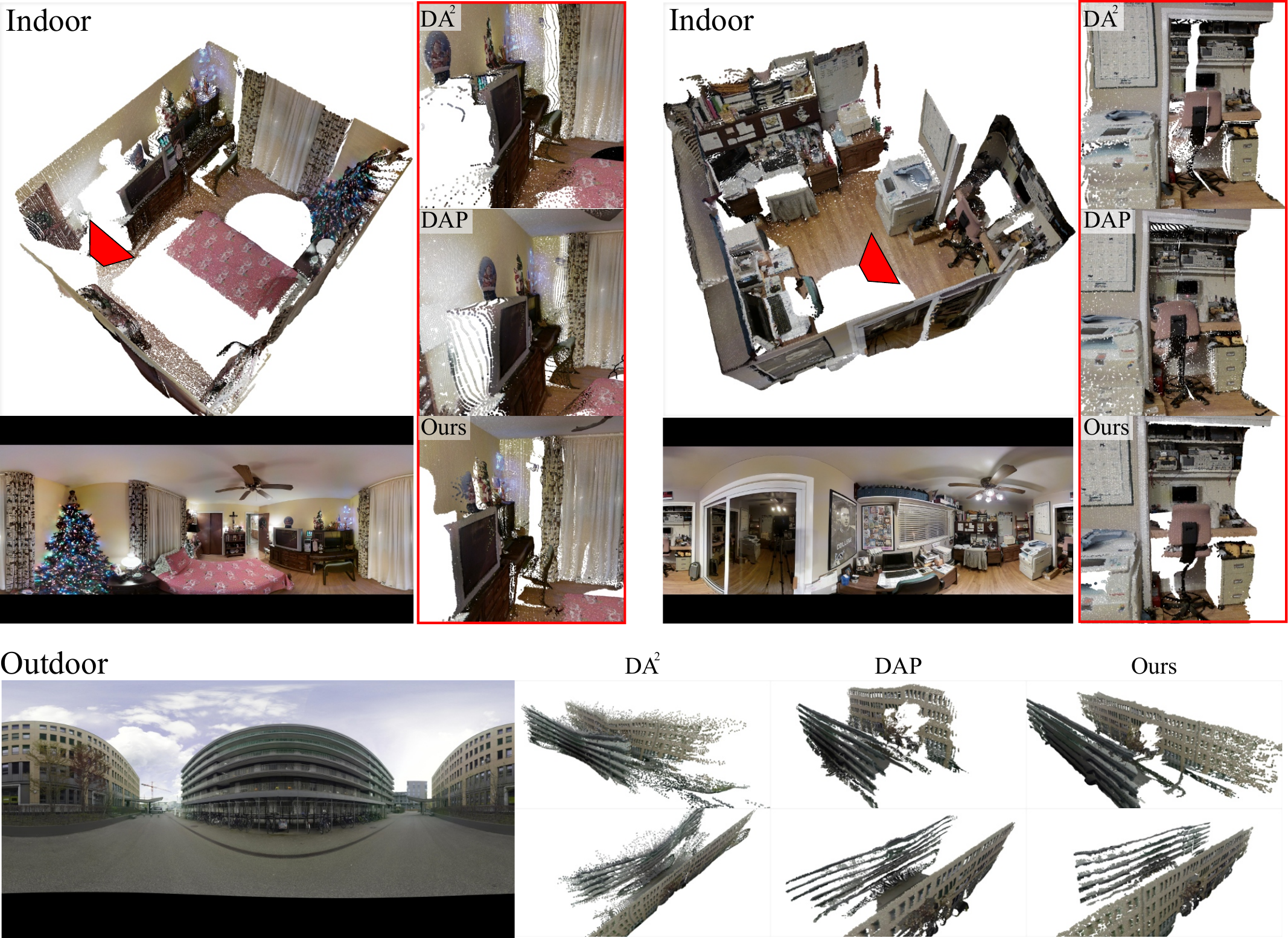}
    \caption{\textbf{Qualitative point-cloud comparison.} Indoor scenes (top) and an outdoor scene (bottom) are rendered as point clouds alongside the corresponding panoramic input images for competitors and our method. For the indoor examples, we show our point cloud reconstruction with zoomed-in novel-view rendering comparison to the main competitors, highlighted by red boxes.}
    \label{fig:pcl-comparison}
\end{figure}
\begin{figure}[!t]
    \centering
    \includegraphics[width=0.35\linewidth]{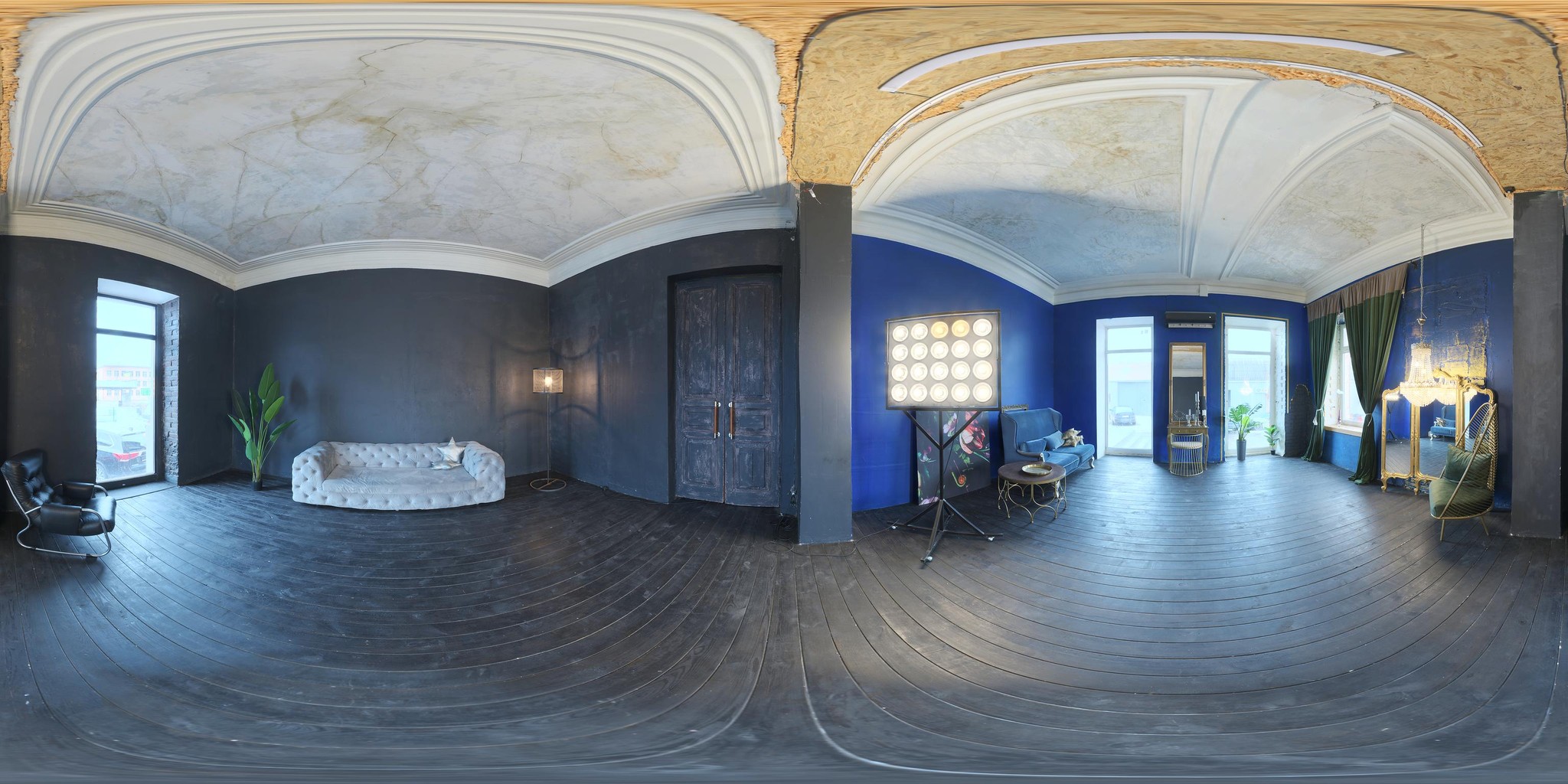}\hspace{5pt}
    \includegraphics[width=0.35\linewidth]{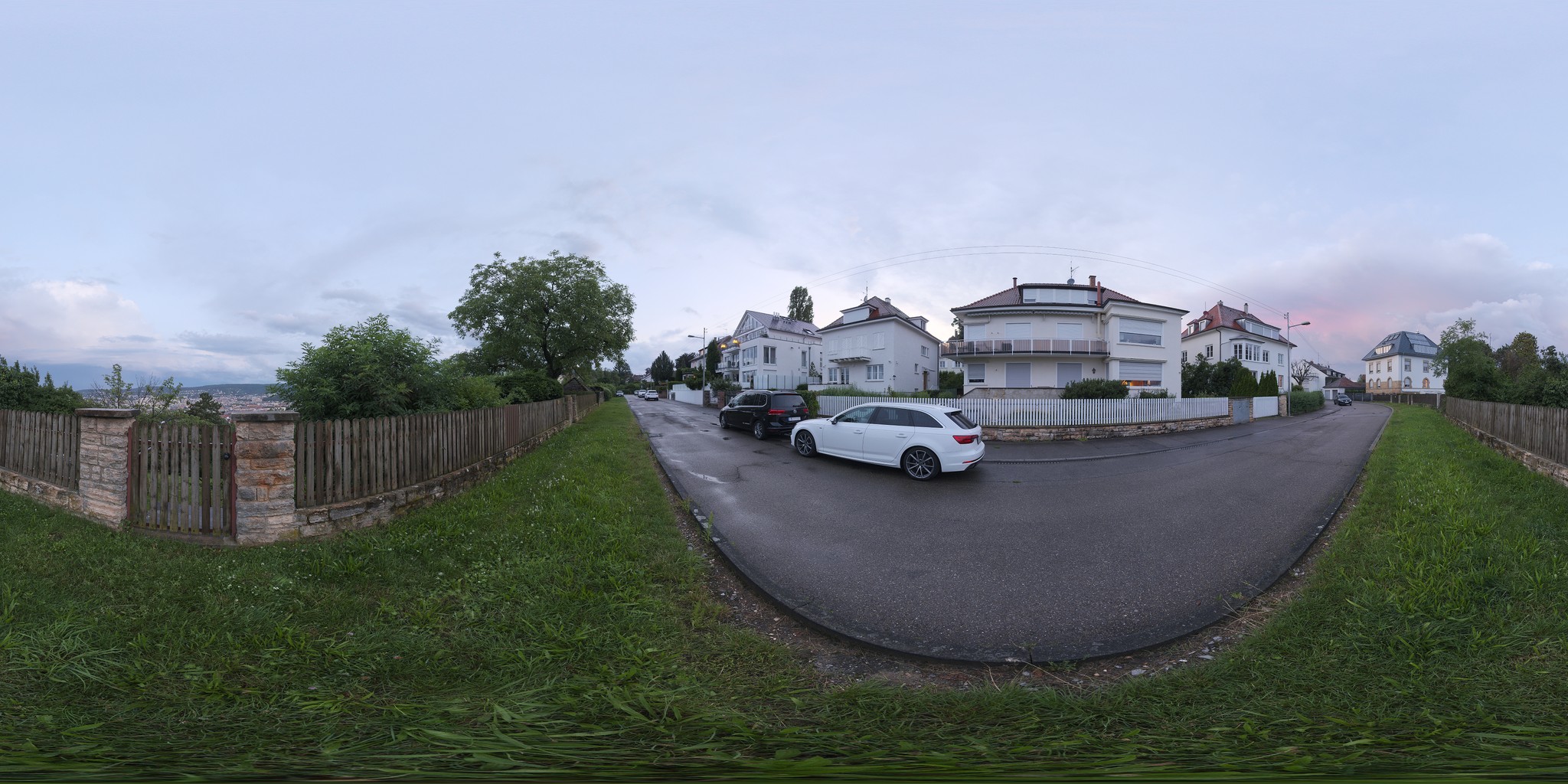}\\[-2pt] 
    \includegraphics[width=0.35\linewidth]{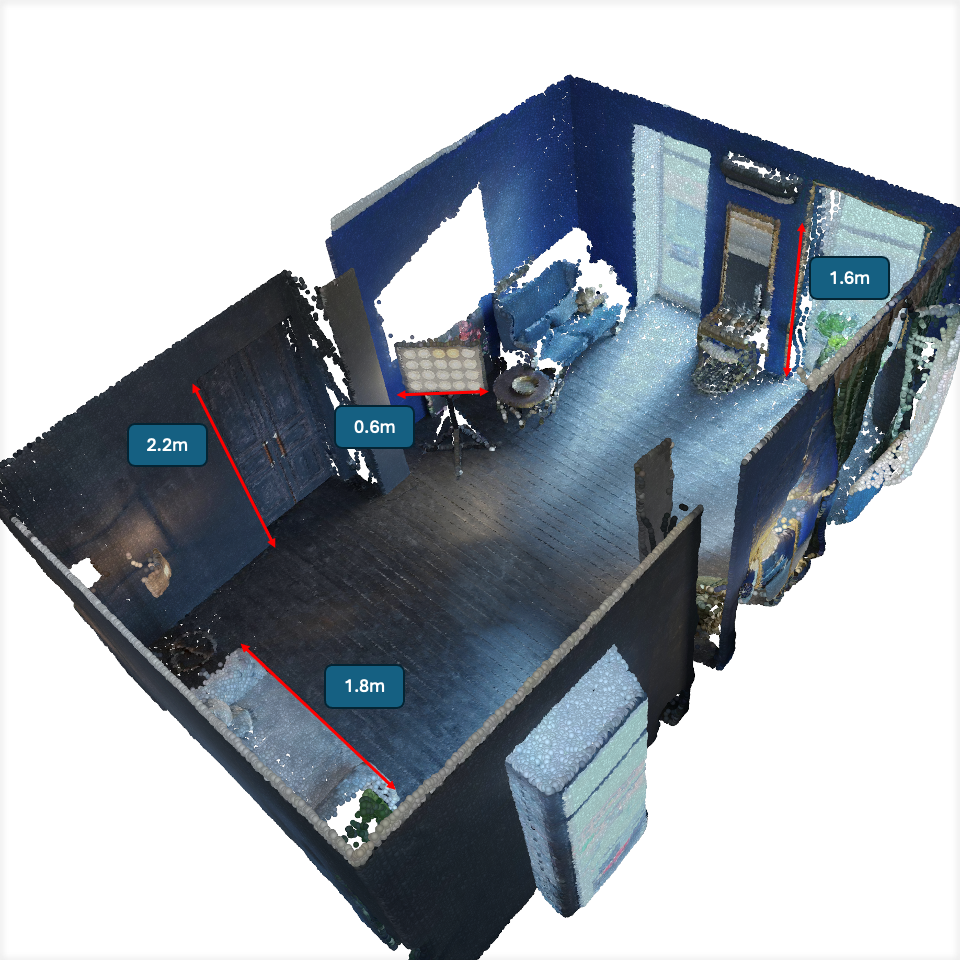}\hspace{5pt}
    \includegraphics[width=0.35\linewidth]{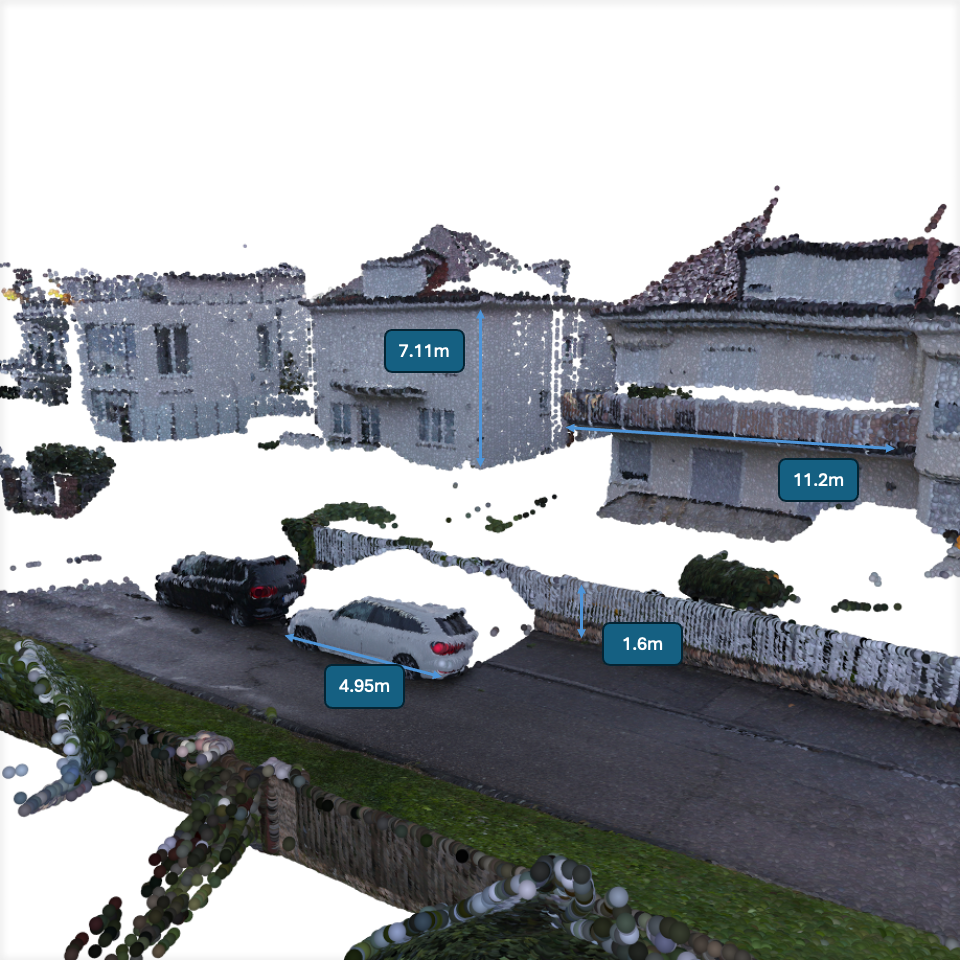}
    \caption{Examples of measured distances in our predicted point cloud. The measures are given in meters.}
    \label{fig:metric-pc}
\end{figure}

\begin{figure}[!t]
\resizebox{\textwidth}{!}{%
\begin{tikzpicture}[inner sep=0,rectspy={lens={scale=3.25}, width=3.9cm, height=2cm}]

\node[label={[inner sep=0pt,anchor=mid,rounded corners,yshift=12.1em]below:{\large Input}}]
    at (4*0,0) {\includegraphics[width=3.9cm]{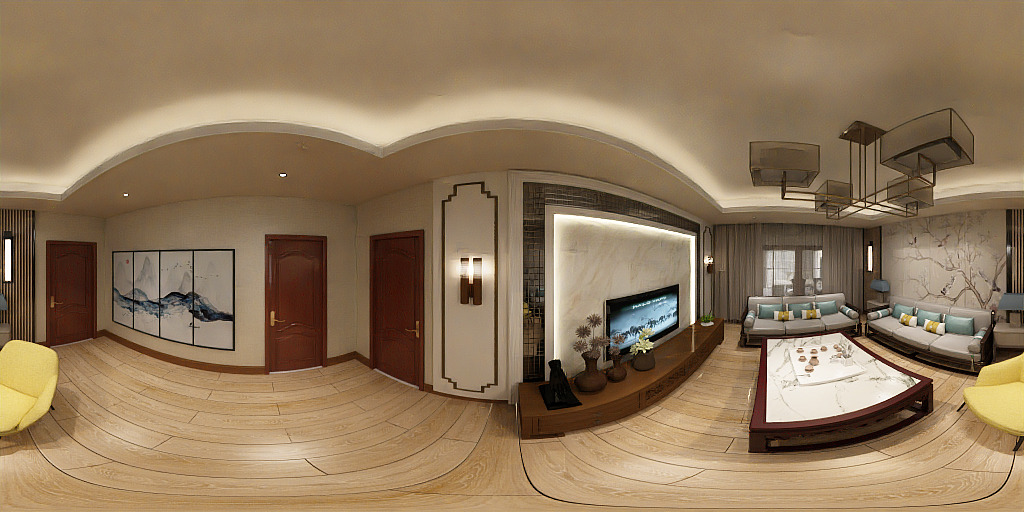}};
\node[label={[inner sep=0pt,anchor=mid,rounded corners,yshift=12.1em]below:{\large MTL}}]
    at (4*1,0) {\includegraphics[width=3.9cm]{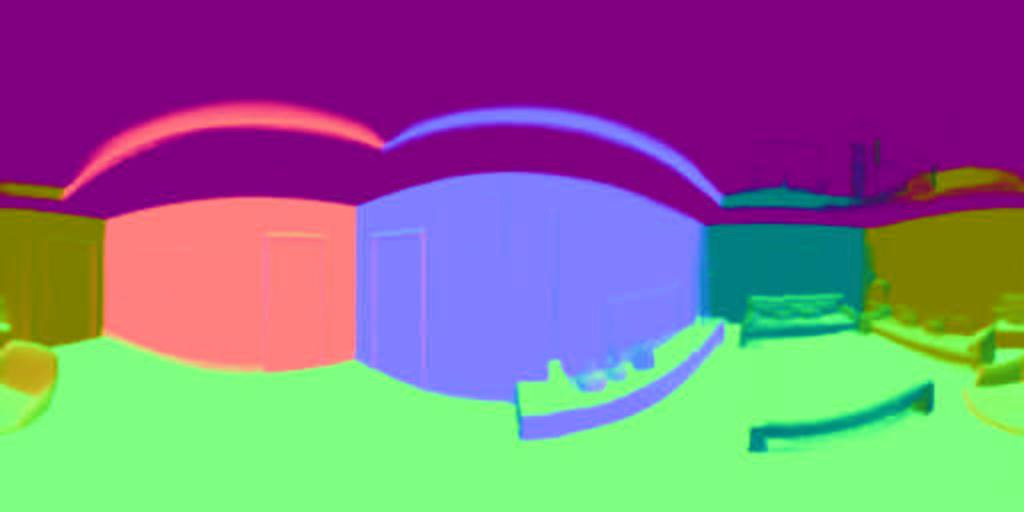}};
\node[label={[inner sep=0pt,anchor=mid,rounded corners,yshift=12.1em]below:{\large Ours}}]
    at (4*2,0) {\includegraphics[width=3.9cm]{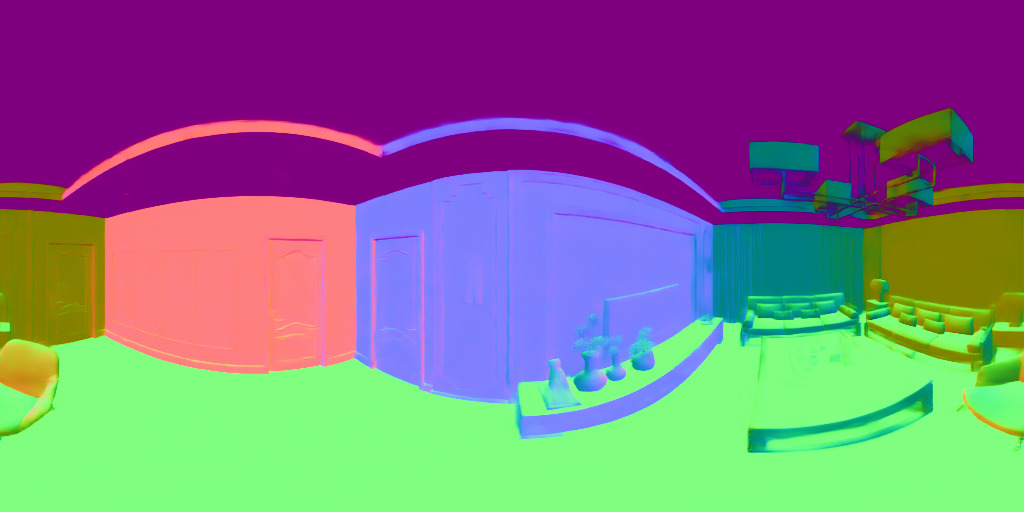}};
\node[label={[inner sep=0pt,anchor=mid,rounded corners,yshift=12.1em]below:{\large Ground Truth}}]
    at (4*3,0) {\includegraphics[width=3.9cm]{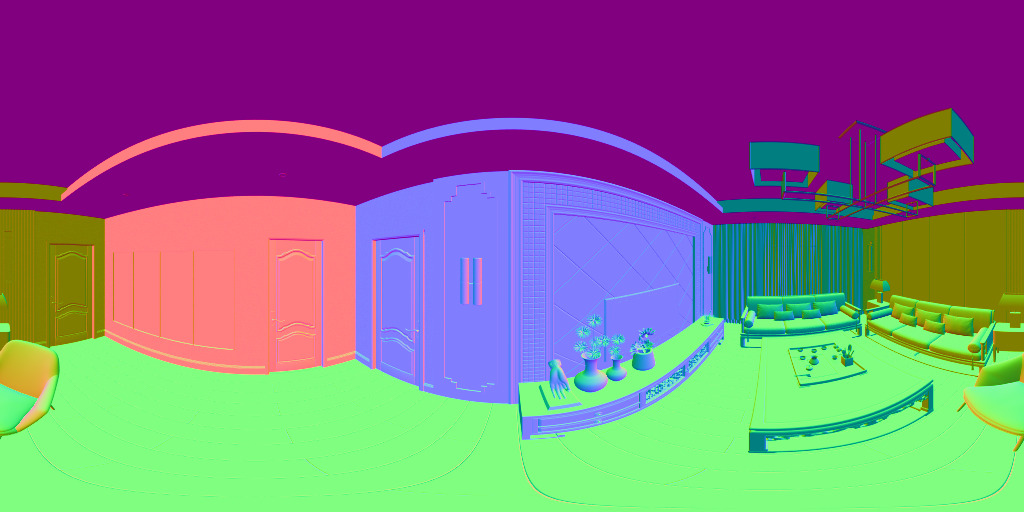}};


\node[label={[inner sep=0pt,anchor=mid,rounded corners,yshift=12.1em]below:{}}]
    at (4*0,-4.25) {\includegraphics[width=3.9cm]{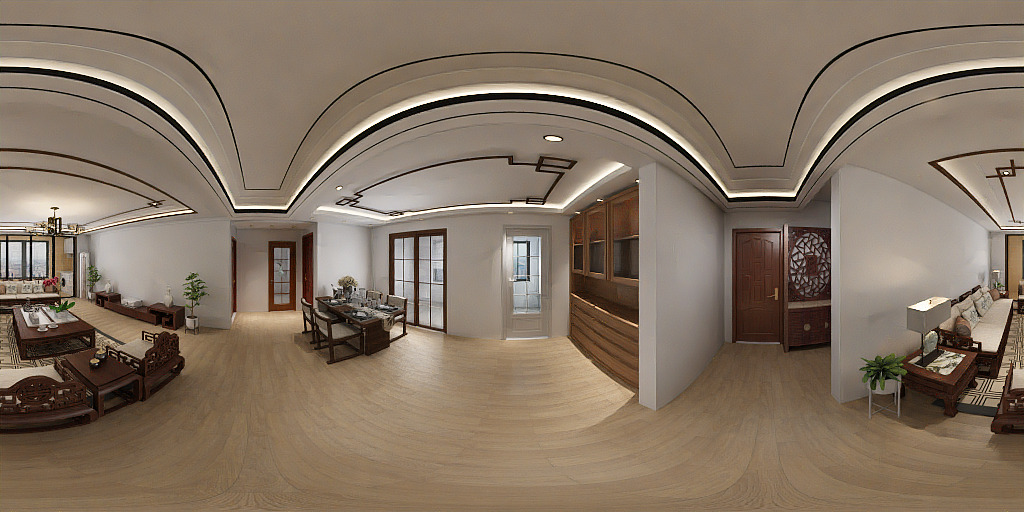}};
\node[label={[inner sep=0pt,anchor=mid,rounded corners,yshift=12.1em]below:{}}]
    at (4*1,-4.25) {\includegraphics[width=3.9cm]{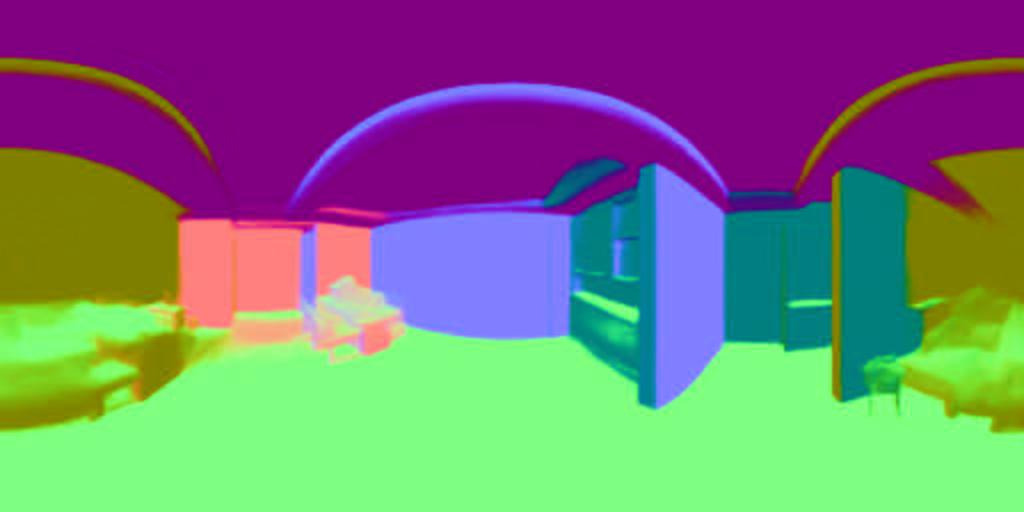}};
\node[label={[inner sep=0pt,anchor=mid,rounded corners,yshift=12.1em]below:{}}]
    at (4*2,-4.25) {\includegraphics[width=3.9cm]{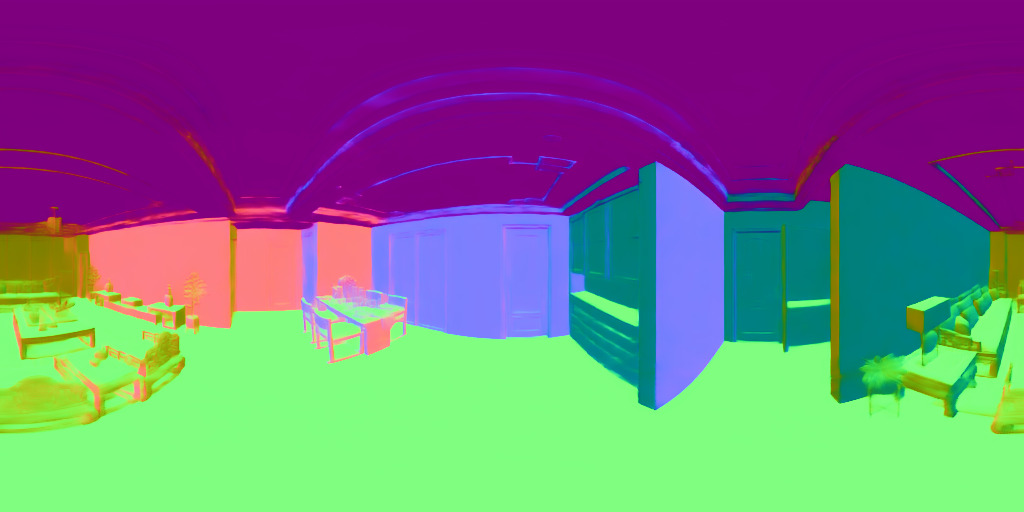}};
\node[label={[inner sep=0pt,anchor=mid,rounded corners,yshift=12.1em]below:{}}]
    at (4*3,-4.25) {\includegraphics[width=3.9cm]{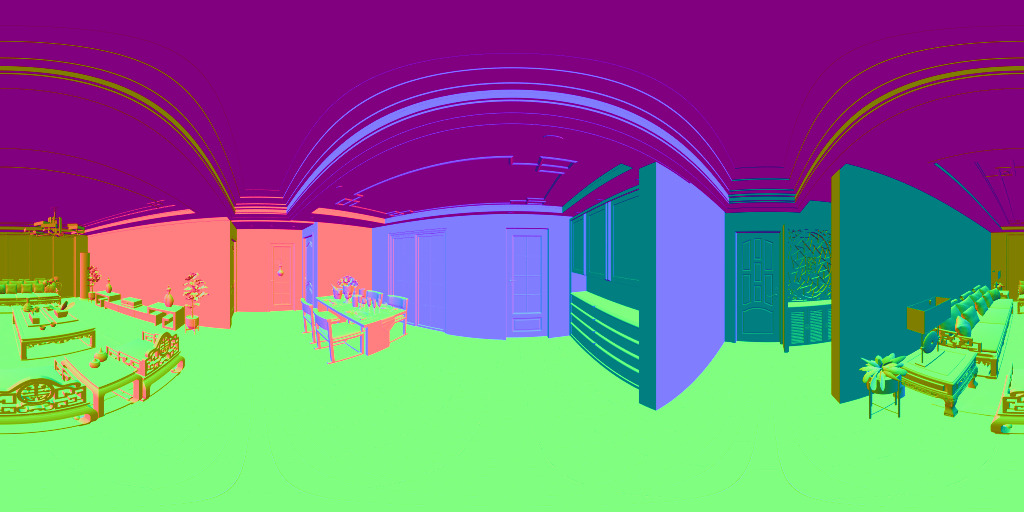}};


\node[label={[inner sep=0pt,anchor=mid,rounded corners,yshift=12.1em]below:{}}]
    at (4*0,-8.5) {\includegraphics[width=3.9cm]{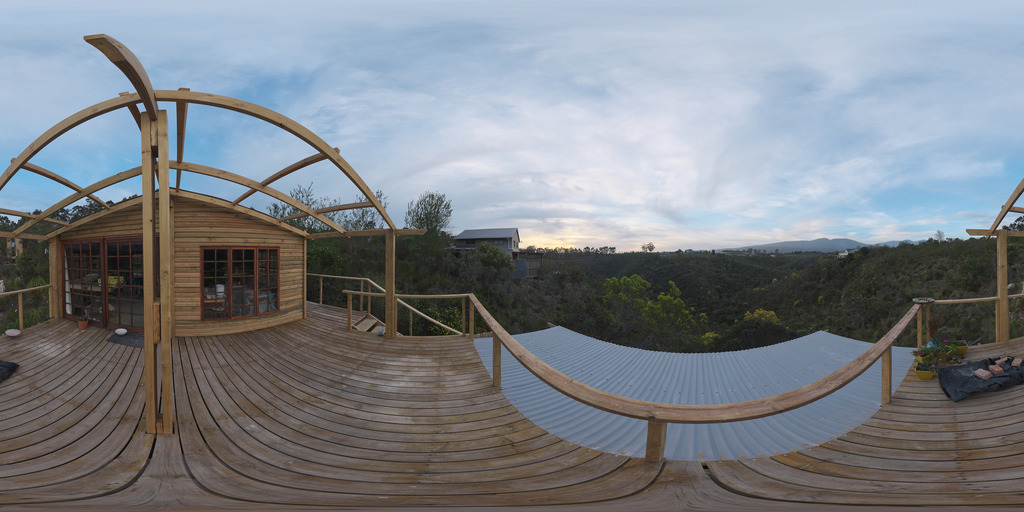}};
\node[label={[inner sep=0pt,anchor=mid,rounded corners,yshift=12.1em]below:{}}]
    at (4*1,-8.5) {\includegraphics[width=3.9cm]{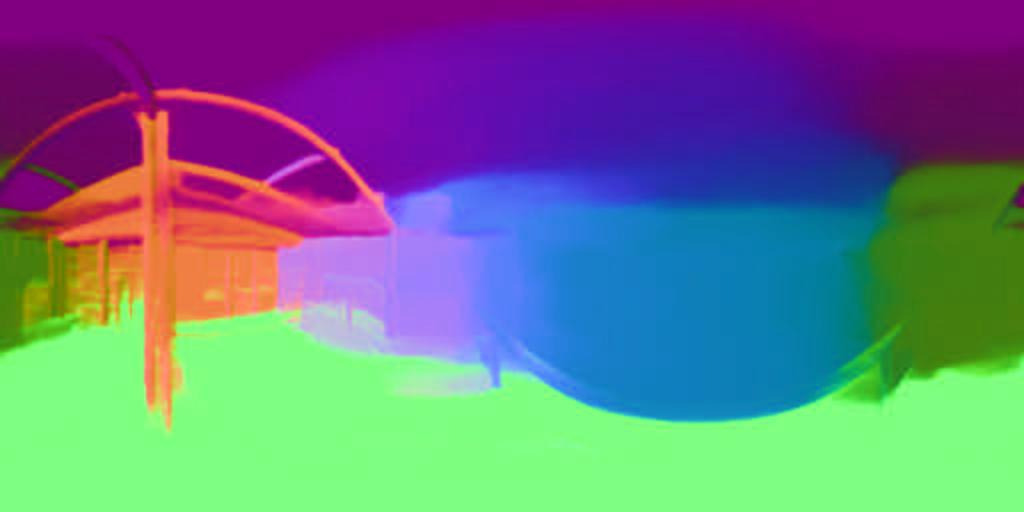}};
\node[label={[inner sep=0pt,anchor=mid,rounded corners,yshift=12.1em]below:{}}]
    at (4*2,-8.5) {\includegraphics[width=3.9cm]{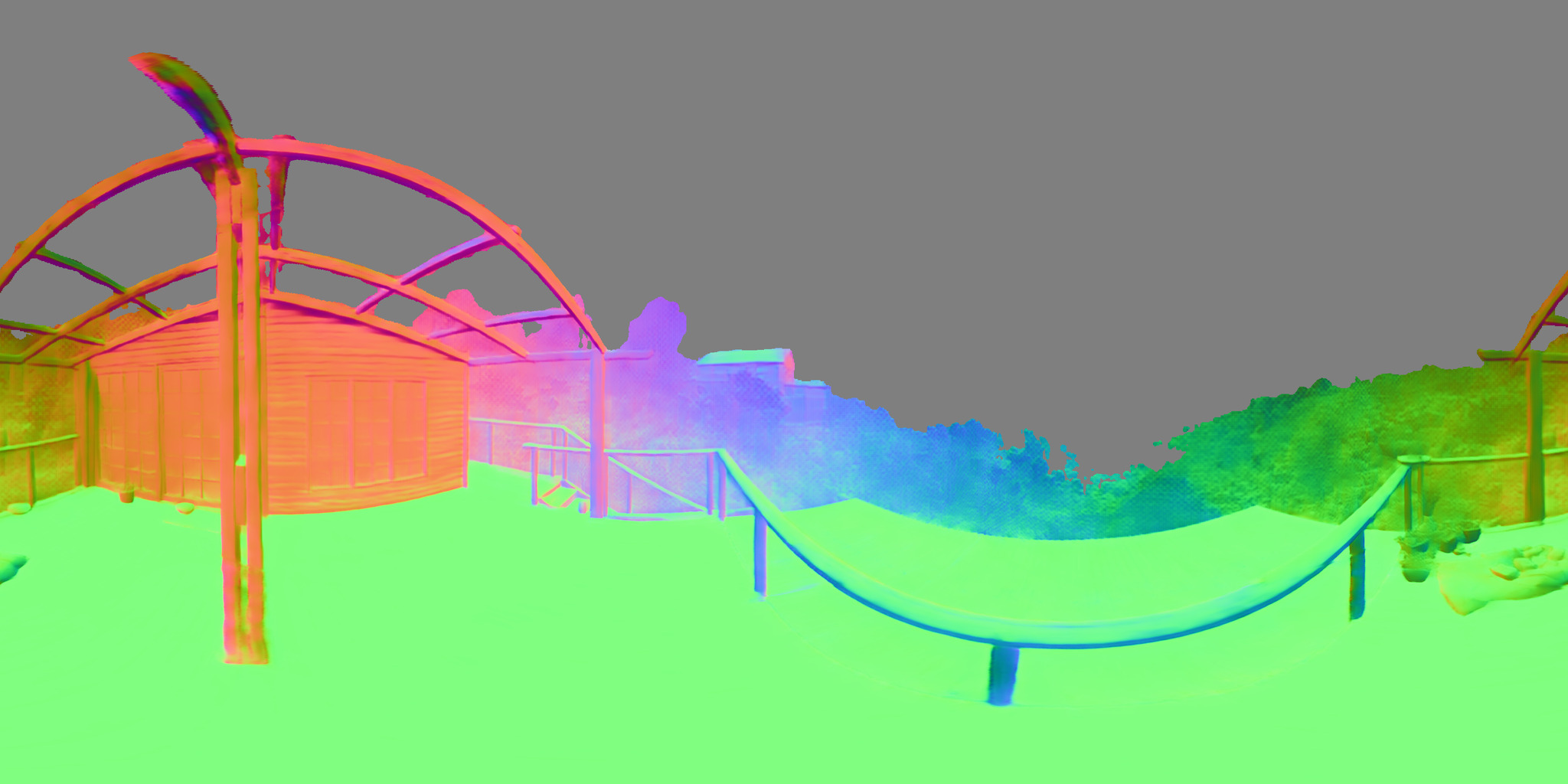}};

\spy on (0.1,-0.3) in node at (0,2);
\spy on (4.1,-0.3) in node at (4,2);
\spy on (8.1,-0.3) in node at (8,2);
\spy on (12.1,-0.3) in node at (12,2);

\spy on (-0.4,-4.5) in node at (0,-2.25);
\spy on (3.5,-4.5) in node at (4,-2.25);
\spy on (7.5,-4.5) in node at (8,-2.25);
\spy on (11.5,-4.5) in node at (12,-2.25);

\spy on (-0.5,-8.65) in node at (0,-6.5);
\spy on (3.6,-8.65) in node at (4,-6.5);
\spy on (7.6,-8.65) in node at (8,-6.5);

\end{tikzpicture}
}
\vspace{-.5cm}
\caption{\textbf{Qualitative comparison of panoramic surface normals estimation.} Visual results from \name{} and MTL (best available baseline method), shown alongside the RGB input and ground-truth depth on panoramas from the Structured3D dataset. (Best viewed zoomed in.)}
\label{fig:comparison_normals}
\end{figure}


\end{document}